\documentclass[10pt,twocolumn,letterpaper]{article}
\usepackage{cvpr}              

\usepackage{cuted}   
\usepackage{capt-of} 
\usepackage{stfloats}
\usepackage{amssymb}
\usepackage{pifont}
\newcommand{\cmark}{\textcolor[rgb]{0,0.6,0}{\ding{51}}}
\newcommand{\xmark}{\textcolor[rgb]{0.8,0,0}{\ding{55}}}
\usepackage{graphicx}
\usepackage{array}
\usepackage{rotating}
\usepackage{multirow}
\usepackage{booktabs}
\makeatletter
\renewcommand{\maketitlesupplementary}{%
  \begin{center}
    {\LARGE Supplementary Material \par}
    {\Large \@title \par}
    \vspace{0.5em}
    {\large \@author \par}
  \end{center}
}
\makeatother

\definecolor{cvprblue}{rgb}{0.21,0.49,0.74}
\usepackage[pagebackref,breaklinks,colorlinks,allcolors=cvprblue]{hyperref}

\def\paperID{9585} 
\def\confName{CVPR}
\def\confYear{2026}

\usepackage[dvipsnames]{xcolor}

\newcommand{\data}{\textsc{OpenTouch}}
\usepackage{tcolorbox}
\usepackage{listings}
\usepackage{enumitem}
\usepackage{listings}
\lstdefinelanguage{json}{
  basicstyle=\ttfamily\small,
  showstringspaces=false,
  morestring=[b]",
  morecomment=[l]{//},
  morecomment=[s]{/*}{*/},
}
\title{\data: Bringing Full-Hand Touch to Real-World Interaction
\vspace{-1em}
}

\author{%
\parbox{\textwidth}{\centering
Yuxin Ray Song\textsuperscript{1,*}\quad
Jinzhou Li\textsuperscript{2,*}\quad
Rao Fu\textsuperscript{3,*}\quad
Devin Murphy\textsuperscript{4}\quad
Kaichen Zhou\textsuperscript{1,5}\quad \\
Rishi Shiv\textsuperscript{1}\quad
Yaqi Li\textsuperscript{1}\quad
Haoyu Xiong\textsuperscript{1}\quad
Crystal E.\ Owens\textsuperscript{1}\quad
Yilun Du\textsuperscript{5}\quad \\
Yiyue Luo\textsuperscript{4}\quad
Xianyi Cheng\textsuperscript{2}\quad
Antonio Torralba\textsuperscript{1}\quad
Wojciech Matusik\textsuperscript{1}\quad
Paul Pu Liang\textsuperscript{1}\\[0.4em]
\textsuperscript{1}MIT\quad
\textsuperscript{2}Duke University\quad
\textsuperscript{3}Brown University\quad
\textsuperscript{4}University of Washington\quad
\textsuperscript{5}Harvard University\\[0.3em]
\small \textsuperscript{*}Equal contribution\\
\normalsize \url{https://opentouch-tactile.github.io/} \\
}
}

\begin{document}
\maketitle

\begin{strip}
\vspace{-5em}
\centering
\includegraphics[width=\textwidth,keepaspectratio]{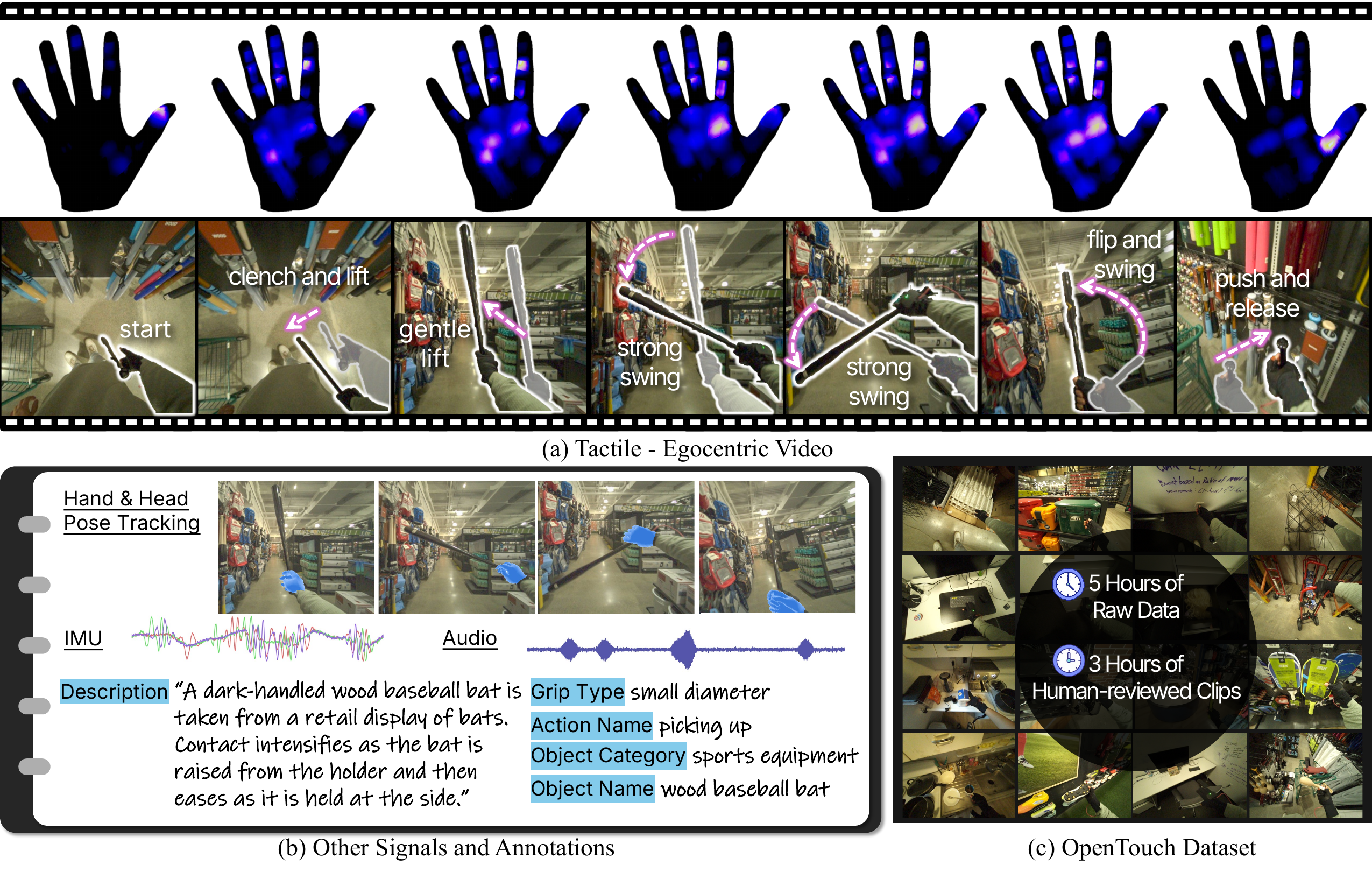}
\vspace{-2em}
\captionof{figure}{
\textbf{\data\ is the first in-the-wild, full-hand tactile dataset with synchronized egocentric video, force-aware full-hand touch, and hand-pose trajectories.} \data{} provides: (a) tactile signals reveal rich, time-varying contact and support forces that are indistinguishable from vision alone, even under nearly identical grasp poses. 
(b) hardware-based, robust hand-pose tracking, and extensive text annotations.
(c) \textbf{5 hours} of recordings, including \textbf{3 hours} of \textit{densely annotated}, contact-rich interactions. See the supplementary video for the sensitivity and fidelity of the tactile dynamics.}
\label{fig:teaser}
\end{strip}

\begin{abstract}
The human hand is our primary interface to the physical world, yet egocentric perception rarely knows when, where, or how forcefully it makes contact.
Robust wearable tactile sensors are scarce, and no existing in-the-wild datasets align first-person video with full-hand touch. 
To bridge the gap between visual perception and physical interaction, we present \data, the first in-the-wild egocentric full-hand tactile dataset, containing 5.1 hours of synchronized video-touch-pose data and 2,900 curated clips with detailed text annotations.
Using \data, we introduce retrieval and classification benchmarks that probe how touch grounds perception and action.
We show that tactile signals provide a compact yet powerful cue for grasp understanding, strengthen cross-modal alignment, and can be reliably retrieved from in-the-wild video queries.
By releasing this annotated vision-touch-pose dataset and benchmark, we aim to advance multimodal egocentric perception, embodied learning, and contact-rich robotic manipulation.

\end{abstract}

\begin{table*}[!t]
\footnotesize
\centering
\caption{\textbf{Comparison of egocentric contact and tactile datasets,} 
showing how our in-the-wild \data\ dataset complements and extends prior work in hand pose, contact sensing, and real-force tactile coverage.}
\vspace{-1em}

{\setlength{\tabcolsep}{3pt}
\begin{tabular}{@{}lccccccccc@{}}
\toprule
\textbf{Dataset} & \textbf{In-the-wild} & \textbf{Hand Pose} & \textbf{Contact} & \textbf{Real-force} & \textbf{\# Object Types} & \textbf{\# Environment} & \textbf{Text Description} \\
\midrule
GRAB \cite{grab} & \xmark & MoCap & Analytical & \xmark & 51 & 1 & \xmark \\
ContactDB \cite{contactDB} & \xmark & \xmark & Thermal & \xmark & 50 & 1 & \xmark \\
ContactPose \cite{contactpose} & \xmark & Estimation & Thermal & \xmark & 25 & 1 & \xmark \\
ARCTIC \cite{arctic} & \xmark & MoCap & Analytical & \xmark & 11 & 1 & \xmark \\
OakInk \cite{yang2022oakink} & \xmark & MoCap & Analytical & \xmark & 32 & 1 & \cmark \\
DyTact \cite{cong2025dytact} & \xmark & Estimation & Wet Paint & \xmark & - & 1 & \xmark \\
DexYCB \cite{dexycb} & \xmark & Estimation & Analytical & \xmark & 20 & 1 & \xmark \\
STAG \cite{sundaram2019learning} & \xmark & \xmark & Pressure & \cmark & 26 & 1 & \xmark \\
21 ADL \cite{tactileadl} & \xmark & \xmark & Pressure & \cmark & 16 & 6 & \xmark \\
ActionSense \cite{delpreto2022actionsense} & \xmark & Glove & Pressure & \cmark & 21 & 1 & \xmark \\
EgoPressure \cite{egopressure} & \xmark & Estimation & Pressure & \cmark & - & 1 & \xmark \\
HOI4D \cite{hoi4d} & \cmark & Estimation & \xmark & \xmark & 800 & 610 & \cmark \\
\midrule
\textbf{\data\ (Ours)} & \cmark & \textbf{Glove} & \textbf{Pressure} & \cmark & \textbf{$\sim$ 800} & \textbf{14} & \cmark \\
\bottomrule
\end{tabular}
}

\vspace{-0.5em}
\label{tab:tactile_datasets}
\end{table*}

\vspace{-4mm}
\section{Introduction}

Humans do not experience the world through vision alone. 
Everyday actions like picking up a mug, closing a zipper, or fastening a medical brace rely on a tightly coupled loop between visual perception, touch, and proprioception. 
Our sense of touch is not merely a backup channel, but a primary source of information about contact, force, stability, material properties, and subtle motions that vision alone cannot resolve \cite{bilaloglu2016effect, ung2025acute}. 
Yet this modality is often ignored: most large-scale egocentric datasets and models for manipulation emphasize what the world \emph{looks} like \cite{ego4d, egoexo4d, banerjee2025hot3d, egodex, govid5m} rather than what it \emph{feels} like.

Capturing full-hand touch \textit{``in-the-wild''} is technically challenging.  
A practical system must simultaneously record vision, touch, and hand motion while remaining non-intrusive, durable, and robust.
It must also provide accurate, low-latency temporal synchronization and denoise highly non-linear tactile signals.
Systems based on parallel-jaw grippers with cameras or fingertip sensors~\cite{umi, egomi, umi-touch} often restrict interactions to simplified grasps and tool use, missing the high-DOF dexterity of the human hand.
Motion-capture gloves and exoskeletons~\cite{dexcap, dexumi} provide detailed kinematics but either omit tactile sensing, or operate in controlled settings with limited environmental diversity.
As a result, we lack a comprehensive in-the-wild dataset that jointly captures full-hand tactile sensing, egocentric vision, and detailed hand motion during natural manipulation.

\begin{figure*}[t]
    \centering
    \begin{subfigure}[b]{0.80\linewidth}
        \centering
        \includegraphics[width=\linewidth]{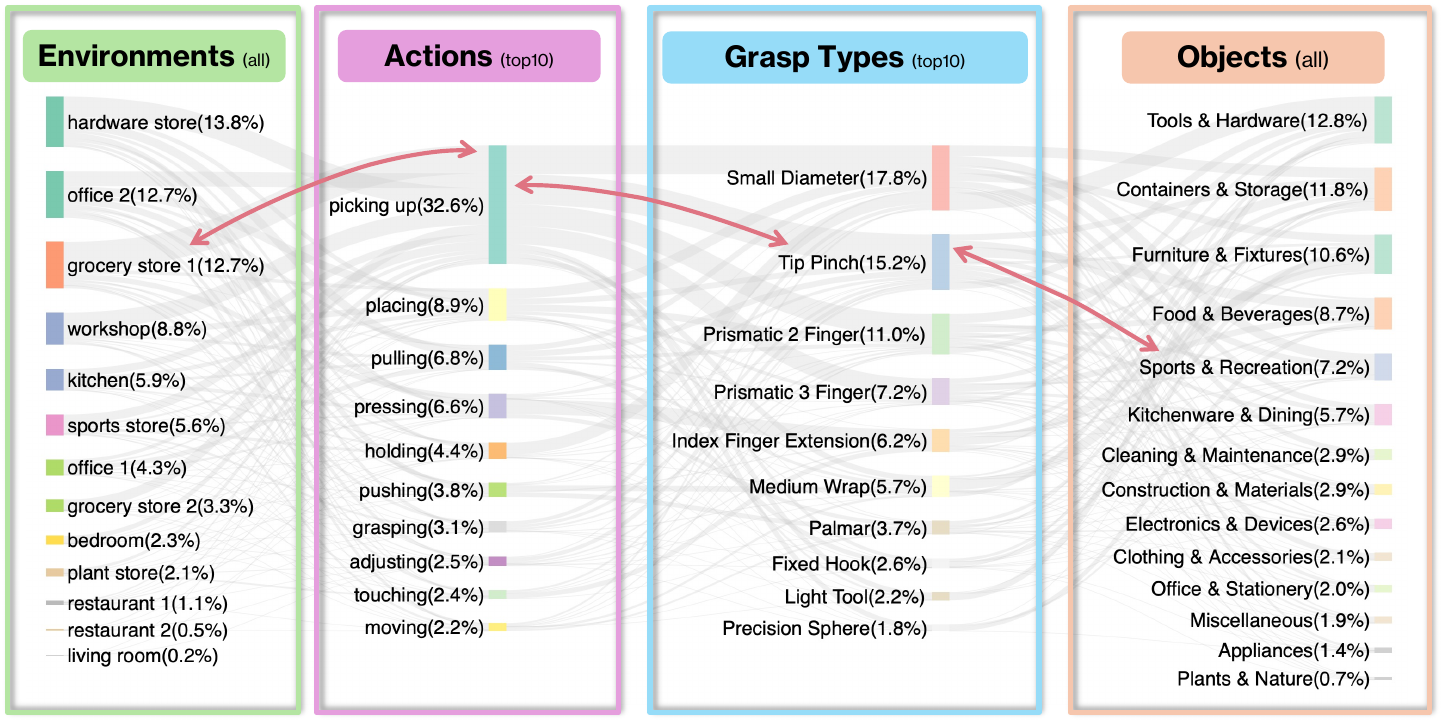}
        \caption{}
        \label{fig:sample-left}
    \end{subfigure}
    \begin{subfigure}[b]{0.18\linewidth}
        \centering
        \includegraphics[width=\linewidth]{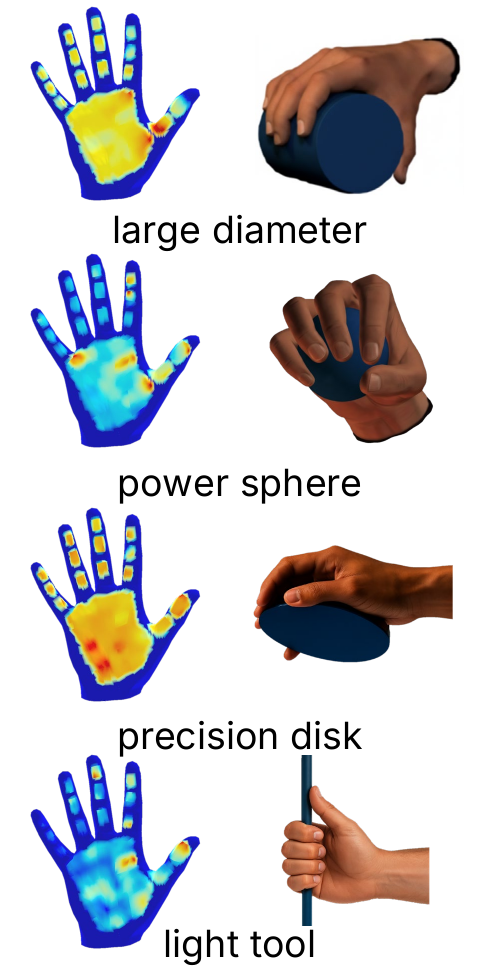}
        \caption{}
        \label{fig:sample-right}
    \end{subfigure}
    \vspace{-1em}
    \caption{
    (a) Sankey diagram visualizing the distribution of dataset labels, including environment, action, grasp type, and object category. See the full list of actions and grasp types in the Supp. Mat.
    (b) Accumulated tactile maps across dataset for different grasp types. The spatial pressure patterns strongly correlate with the underlying grasp configuration, demonstrating the accuracy and quality of our tactile data and grasp type annotation. See the Supp. Mat. for complete tactile–grasp mappings covering all grasp taxonomies.}
    \label{fig:stats}
\end{figure*}

In this paper, we move toward understanding human touch at scale from an egocentric perspective by jointly advancing sensing hardware, in-the-wild data collection, and benchmark design.
We introduce \data, the first egocentric, full-hand tactile dataset captured in the wild.
We build a low-cost wearable sensing sensors, a custom FPC-based tactile glove paired with a hand-tracking glove and smart glassess, and ask participants to freely manipulate objects across 14 everyday environments, covering over 8{,}000 objects from 14 categories.
The resulting dataset contains 5.1 hours of synchronized vision-touch-pose recordings and 2{,}900 human-reviewed clips (3 hours).
\Cref{fig:teaser} visualizes example data and the corresponding text annotations.
Each clip is labeled with object name, object category, environment, action, grasp type, and a brief description (\Cref{fig:sample-left}).
Beyond RGB video, \data\ also includes temporally aligned head motion, eye tracking, and audio.
Tab.~\ref{tab:tactile_datasets} compares \data with other full-hand visual-contact datasets.
To our knowledge, \data\ is the first dataset to jointly provide in-the-wild egocentric RGB, full-hand tactile maps, and 3D hand pose across diverse, uncontrolled environments.

\begin{figure*}[t]
    \centering
    \includegraphics[width=1.00\linewidth]{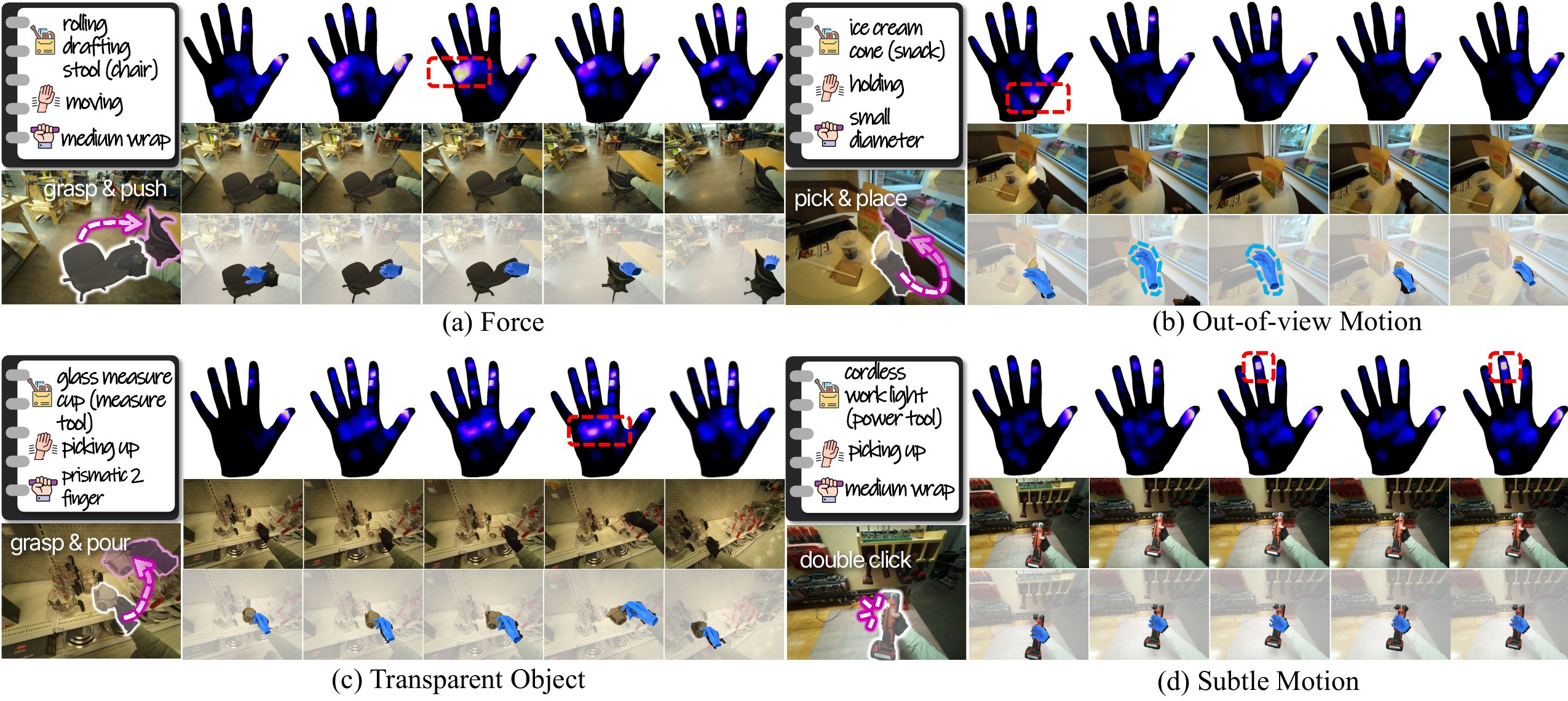}
\caption{
\textbf{Example data from \data\ demonstrates that hardware-based tactile sensing and pose tracking reveal critical force, contact, and motion cues that vision alone cannot capture.}
(a) Although the first three frames show nearly identical hand poses, the tactile signals reveal that in the third frame the hand applies sufficient force to move the chair.
(b) In the first frame, tactile readings clearly indicate contact with the table, ambiguous from RGB alone. In the next two frames, the hand moves out of view, making vision-based pose estimation unreliable; \data~provides accurate hardware-tracked poses throughout.
(c) Tactile sensing exposes clear interaction patterns with transparent object that remain difficult to infer from visual tracking alone.
(d) The tactile map captures a subtle middle-finger double-click on a button, a fine-grained motion that even pose tracking may miss.
See the supplementary video for the high-fidelity tactile signals and subtle dynamic patterns. }
    \label{fig:sample}
\end{figure*}

\data\ provides high-quality tactile and annotations: aggregated tactile maps match grasp-specific force distributions (\cref{fig:sample-right}), and the synchronized vision-touch-pose signals capture rich contact dynamics even for transparent or reflective objects (\cref{fig:sample}).

Built on \data, we introduce retrieval and classification benchmarks to study how touch grounds perception and action.
Experiments reveal three trends.
First, \textbf{multimodal outperform unimodal}: combining video, pose, and tactile (e.g., video+pose$\rightarrow$tactile or video+tactile$\rightarrow$pose) yields markedly higher retrieval and grasp recognition accuracy, as video provides global scene context, pose encodes kinematics, and tactile captures local contact and force.
Second, \textbf{tactile is compact yet powerful}: a lightweight tactile encoder outperforms a deep ResNet backbone, tactile alone is highly informative for grasp type.
Third, \textbf{time and structure matter}: longer windows and moderate signal discretization significantly improve cross-modal alignment.

By capturing contact-rich manipulation at scale in diverse, uncontrolled settings,
\data\ can also act as a tactile database: we demonstrate that in-the-wild video (e.g., Ego4D~\cite{ego4d}) can retrieve plausible tactile sequences, \textbf{enabling large-scale egocentric video to be augmented with contact and force cues}.

We will release the full dataset, hardware designs, and training code to support the community.
Our main contributions are:
\begin{enumerate}[leftmargin=*]
    \item \textbf{A practical, low-cost sensing setup} that captures synchronized egocentric RGB video, full-hand tactile maps, and hand pose in the wild, along with supporting software for calibration, denoising, and temporal alignment.
    \item \textbf{A large-scale in-the-wild tri-modal dataset}, \data, with 5.1 hours of vision-touch-pose recordings and 2{,}900 human-reviewed clips, annotated with object identity, object category, environment, action type, grasp type, and natural-language descriptions.
    \item \textbf{Benchmarks for vision-touch-pose learning}, including cross-sensory retrieval and tactile grasp classification, together with systematic ablations on temporal context, tactile encoder capacity, and discretization strategies that show how to efficiently encode full-hand tactile signals and how touch most effectively complements vision and pose.
\end{enumerate}

\section{Related Work}
\vspace{-0.25em}
\textbf{Egocentric human video datasets.}
Notable contributions to egocentric human video datasets have been made in both the computer vision~\cite{ego4d, egoexo4d, sthsth, egohuman,  egobody, jawaid2025openego, emhi} and robotics communities~\cite{egomimic, egodex, egomi, robotube, su2025robosense, egovla, emma, li2025learning, maple}. Most existing datasets focus on capturing human-object grasps with head-mounted cameras~\cite{epickitchen, epickitchen100, hoi4d, egoexofitness, egoexolearn, egoobjects, egogesture}. However, when relying solely on egocentric vision, visual occlusion remains a significant limitation~\cite{via}. To address this, our dataset augments egocentric video with tactile sensing, which is crucial for detecting and understanding contacts that are often imperceptible through vision alone.

\vspace{1mm}
\noindent
\textbf{Tactile sensing.}
Prior work has introduced datasets of human-object contact, e.g., ContactDB~\cite{contactDB}, ContactPose~\cite{contactpose}, and DyTact~\cite{cong2025dytact}. 
Hand pressure can be captured either by instrumenting the environment~\cite{pham2017hand, pressurevision, egopressure}, such as or pressure pads~\cite{pressurevision, egopressure}.
While effective in constrained settings, such pads are not easily scalable and are fundamentally limited by their fixed geometry. We instead adopt a hand-worn form factor, which preserves dexterity and supports a wide range of objects, grasps, and environments. 
Our design is inspired by tactile glove systems~\cite{sundaram2019learning, murphy2025flexglove, BUSCHER2015glove, tactileadl} but improves accessibility and reproducibility. 
Exisitng sensors are either proprietary \cite{tekscan, pps}, difficult to fabricate \cite{sundaram2019learning}, or require specialized assembly pipelines \cite{luo2024adaptive}. 
Our tactile sensor, by contrast, leverages flexible printed circuits (FPCs) to automate layout and streamline production, enabling low-cost, repeatable fabrication and easy adaptation to new hand sizes and setups.

\vspace{1mm}
\noindent
\textbf{Multimodal learning.}
Cross-modal contrastive learning has been widely used to align modalities, including image-text~\cite{radford2021learning,jia2021scaling,liang2024foundations}, video-text~\cite{miech2019howto100m,gabeur2020multi, 10.1016/j.neucom.2022.07.028}, and sound~\cite{audioclip}. Large multisensory models such as ImageBind~\cite{girdhar2023imagebind} embed vision, audio, depth, and IMU signals in a shared space, while object-centric datasets like ObjectFolder~\cite{gao2022ObjectFolderV2, gao2023objectfolder} study unified representations of appearance, geometry, and sound. Touch–vision\cite{li2024hypertaxel,yang2024binding, yang2022touch} and touch–sound studies~\cite{li2024learning, li2019connecting, li2025multimodal} shows tactile cues complement vision. \data\ is the first \textit{in-the-wild, egocentric} dataset with synchronized vision-touch-pose signals, extending cross-modal retrieval to this setting and enabling systematic studies of how touch grounds perception and action in natural environments.


\section{\data\ Benchmark}

\vspace{-2mm}

\begin{figure*}[t]
  \centering
\includegraphics[width=0.98\textwidth,keepaspectratio]{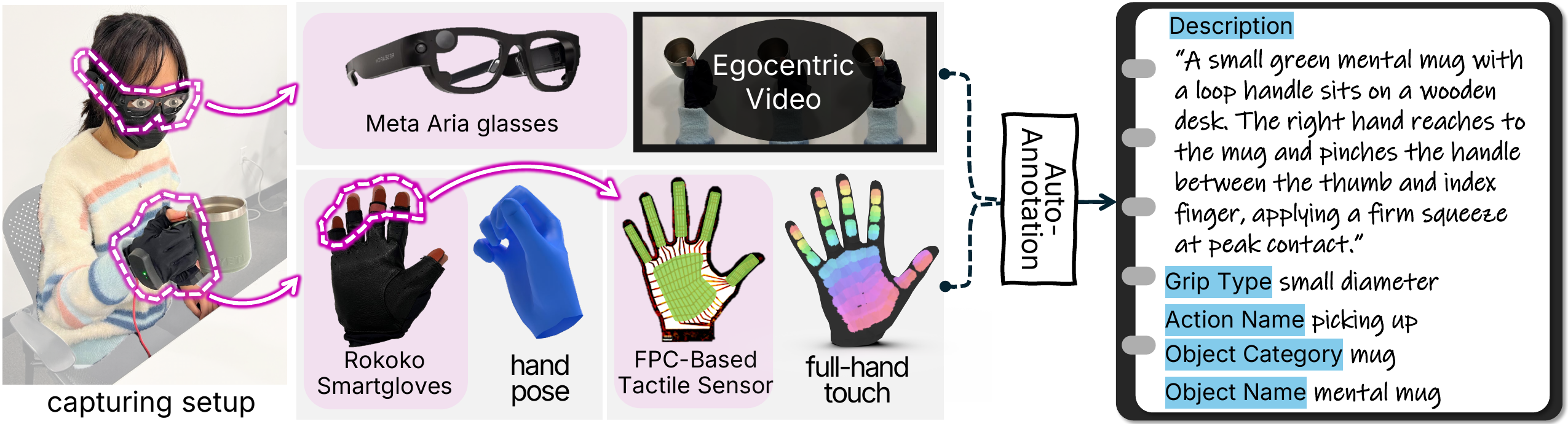}
    \caption{
    \textbf{Overview of the data-capture and annotation setup.} Meta Aria glasses, Rokoko Smartgloves, and the FPC-based tactile sensor are synchronized at 30 Hz with an average 2 ms latency, enabled by a zero-potential readout circuit and lightweight ESP-NOW wireless transmission. The system captures synchronized egocentric video, hand pose, and dense full-hand touch signals. High-level descriptions and detailed annotations are automatically generated from the egocentric video and the rendered tactile maps using a large language model.}
  \label{fig:glove}
\vspace{-2mm}
\end{figure*}

\Cref{fig:sample} shows example clips from \data, illustrating diverse in-the-wild interactions with high-quality synchronized vision--touch--pose signals that capture force, contact, and motion.
\Cref{fig:glove} summarizes our hardware setup and dataset annotation pipeline.

\subsection{Hardware Setup}
\label{sec:hardware}


\vspace{1mm}
\noindent
\noindent\textbf{Customized tactile sensing glove.} Existing tactile gloves often trade off resolution, coverage, and robustness, especially when relying on hard-to-reproduce conductive textiles \cite{BUSCHER2015glove, luo2024tactile}. We instead design a thin, low-cost, fully open-source hand-shaped sensor for large-scale, in-the-wild multimodal data collection. Our FPCs layout routes a 16×16 electrode grid around a commercial piezoresistive film, forming 169 taxels that uniformly cover the fingers and palmar surface. This combines PCB-level precision and reliability with the compliance needed for wearable sensing, enabling stable, high-resolution pressure mapping without bulky cabling or manual wiring \cite{murphy2025flexglove}.

\vspace{1mm}
\noindent\textbf{Hand-tracking glove.} Hand pose is captured with Rokoko Smartglove \cite{rokoko}, a professional motion-capture system using fused IMU and EMF sensing. Each glove provides seven 6-DOF sensors and streams at 30 Hz. The reported rotational dynamic accuracy is $\pm1^\circ$; positional accuracy follows from the kinematic chain. Before recording, subjects perform a system-defined calibration pose (standing upright with elbows bent at $90^\circ$) to define a consistent zero-point.

\vspace{1mm}
\noindent\textbf{Egocentric video capture.}
We record egocentric vision using Meta’s Project Aria glasses \cite{project_aria_research_kit}, which integrate two mono SLAM cameras, an RGB point-of-view camera with a $110^{\circ}$ field of view, eye-tracking cameras, IMUs, and microphones (Profile 28). For our benchmarks, we focus on the RGB video streaming at 1408×1408 pixels and 30 Hz.

\vspace{1mm}
\noindent\textbf{Time sync and calibration.}
We synchronize vision, tactile, and hand-pose data using a visual cue displayed in the terminal and captured by the RGB camera. 



\subsection{Collection and Annotation}

\data\ was constructed to capture natural, contact-rich human manipulation with synchronized vision, motion, and tactile sensing in diverse real-world environments. 

\vspace{1mm}
\noindent
\textbf{Collection protocol and environments.}
Data was collected following a standardized yet naturalistic protocol across 14 everyday environments. In each location, participants were instructed to freely explore and manipulate every available object, such as grasping tools of varied geometries in a workshop. Each recording session lasted 5 to 25 minutes, yielding short clips averaging 57 frames.

\noindent\textbf{Objects and grasp taxonomy.}
Unlike datasets built around predefined object or verb lists~\cite{gao2021ObjectFolder, fu2025gigahands}, we collected \data\ across 14 everyday environments using the objects already present, yielding more natural and representative interactions than lab settings.
In each environment, participants manipulated all available objects using the right hand. 
We instrument only the right dominant hand to simplify hardware and standardize annotations, , while retaining left-hand generalization via mirroring and pose relabeling.
Activities were performed unscripted, producing a diverse set of power, precision, pinch, lateral, and palmar grasps consistent with the GRASP taxonomy~\cite{feix2015grasp}.


\vspace{1mm}
\noindent\textbf{Annotations.} 
Each clip is annotated with six labels: object name, object category, environment, action, grasp type, and a brief description.
Manual labeling of in-the-wild sequences is costly due to diverse objects and contexts.
To scale annotation, we used GPT-5 for automated labeling.
For each clip, the model received three time-ordered RGB-tactile pairs.
Because the most informative moment of interaction typically occurs near peak contact force, 
we sample frames by pressure dynamics: lowest pressure pre-peak (approach), peak pressure (manipulation), and lowest pressure post-peak (release).
These three frames capture the interaction without requiring the full video.
Using these samples, GPT-5 was prompted to identify the environment, object, object category, primary action, grasp type, and to generate a natural-language description. 
Grasp types are selected from the GRASP taxonomy~\cite{feix2015grasp}.
Human verification shows roughly 90\% labeling accuracy.
Failures mostly occur when the hand leaves the view at peak force, under difficult lighting or clutter, or due to limited context from three frames.




\subsection{Benchmark Tasks and Metrics}
\label{sec:benchmarks}

\data\ introduces two benchmark tasks of cross-sensory retrieval and tactile pattern classification.

\subsubsection{Cross-Sensory Retrieval}
\label{cross-retrieval-task}


Given one modality (e.g., an egocentric RGB frame), we test whether a model can retrieve the corresponding signals in another (e.g., tactile map or hand pose), requiring a shared representation of interaction dynamics.


\vspace{2mm}
\noindent \textbf{Task 1: Video-tactile sequence retrieval.} Given synchronized first-person video and tactile sequences, the goal is to learn a shared embedding space where semantically corresponding segments are close while non-corresponding segments are far apart, enabling retrieval across modalities. We evaluate two directions: \textit{Video} $\rightarrow$ \textit{Tactile} and \textit{Tactile} $\rightarrow$ \textit{Video}.
This task measures how well a model can align the dynamic visual appearance of manipulation with touch.

\vspace{1mm}
\noindent \textbf{Task 2: Pose-tactile sequence retrieval.} We investigate the coupling between hand motion and touch pattern. We first evaluate (i) \textit{Pose} $\leftrightarrow$ \textit{Tactile} retrieval, which examines whether the magnitude and spatial pattern of contact forces are consistent with the corresponding hand configuration. We then extend to (ii) \textit{Multimodal} $\rightarrow$ \textit{Unimodal} retrieval, where combined cues (e.g., \textit{Video + Tactile}) are used to retrieve the remaining modality. This task probes how well a model understands the relationship between hand configuration and contact pattern. 

\begin{figure*}[t]
    \centering

    \setlength{\tabcolsep}{0pt}
    \renewcommand{\arraystretch}{0.0}

    \newcommand{\rowlabel}[1]{\rotatebox[origin=c]{90}{#1}}

    \newcommand{\sqimgright}[1]{%
        \includegraphics[height=1.6cm,trim=600 120 80 280,clip]{#1}%
    }
    \newcommand{\sqimgcenter}[1]{%
        \includegraphics[height=1.6cm,trim=240 0 240 160,clip]{#1}%
    }

    \begin{tabular}{
        >{\centering\arraybackslash}m{0.35cm}  
        *{5}{>{\centering\arraybackslash}m{1.7cm}} 
        @{\hspace{1mm}}  @{\hspace{1mm}} 
        *{5}{>{\centering\arraybackslash}m{1.7cm}} 
    }
      & \rule{0pt}{1.8ex}frame \#1 & frame \#2 & frame \#3 & frame \#4 & frame \#5
      & \rule{0pt}{1.8ex}frame \#1 & frame \#2 & frame \#3 & frame \#4 & frame \#5 \\[0.8ex]


        \rowlabel{query} &
            \sqimgright{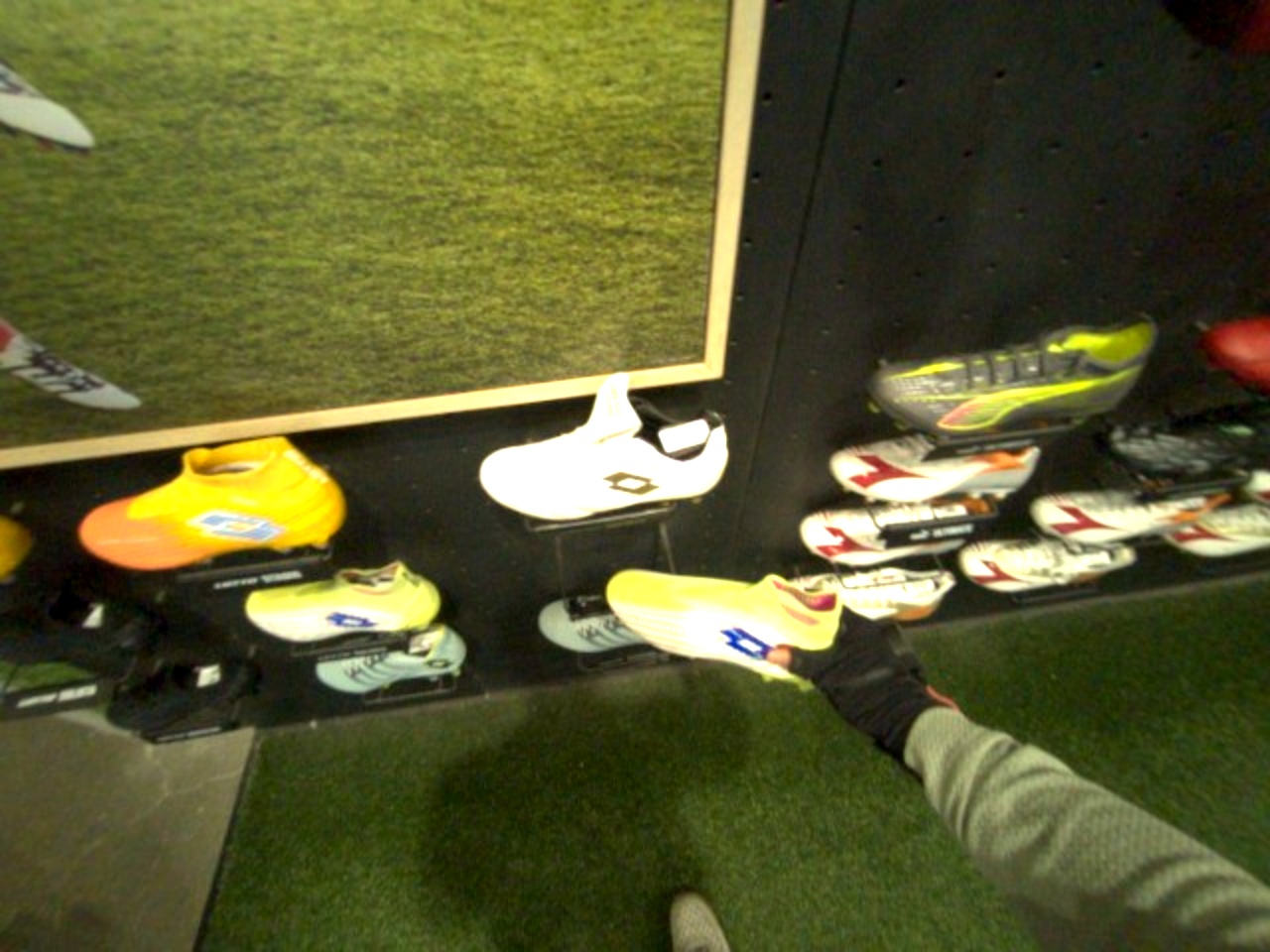} &
            \sqimgright{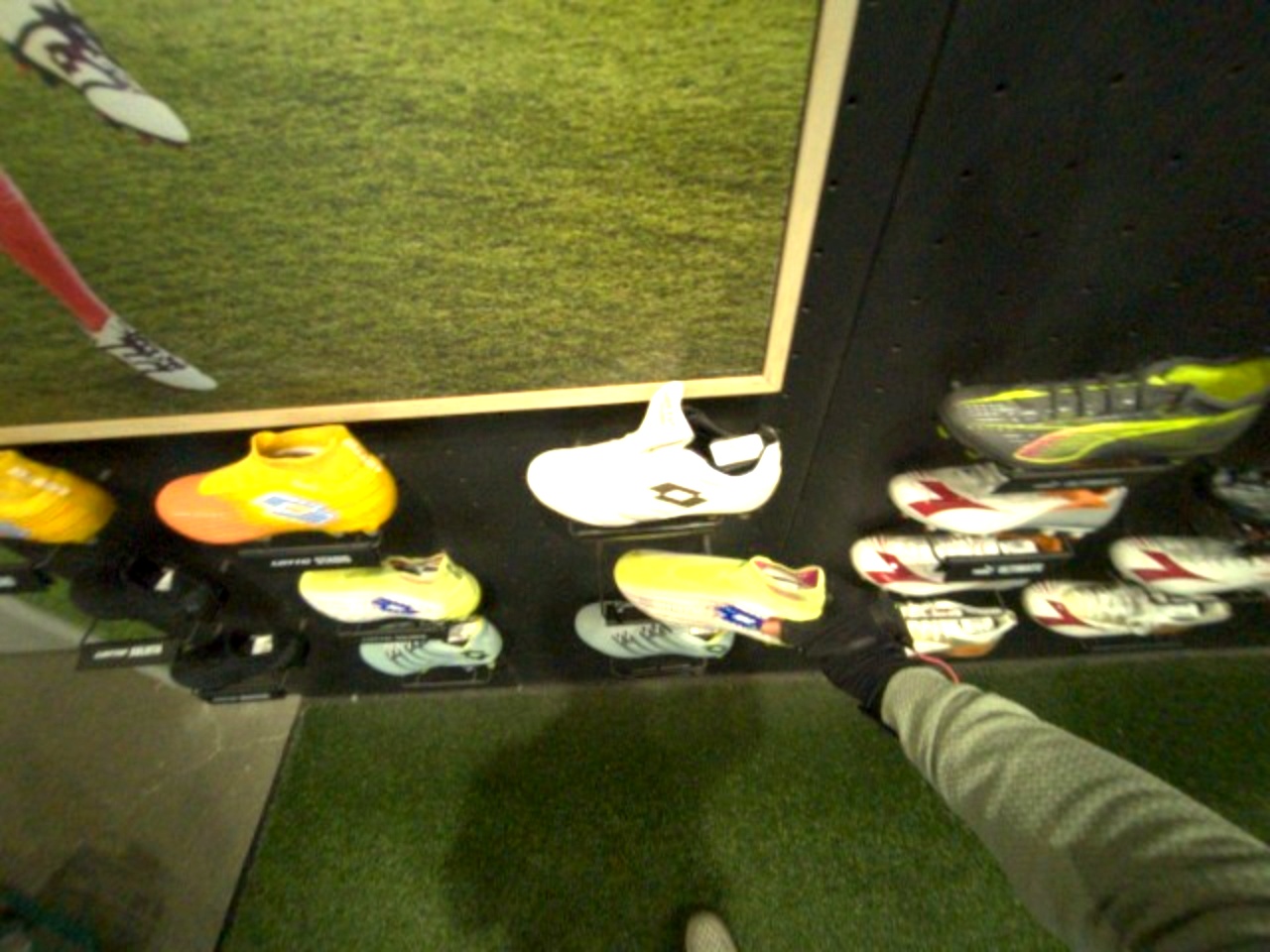} &
            \sqimgright{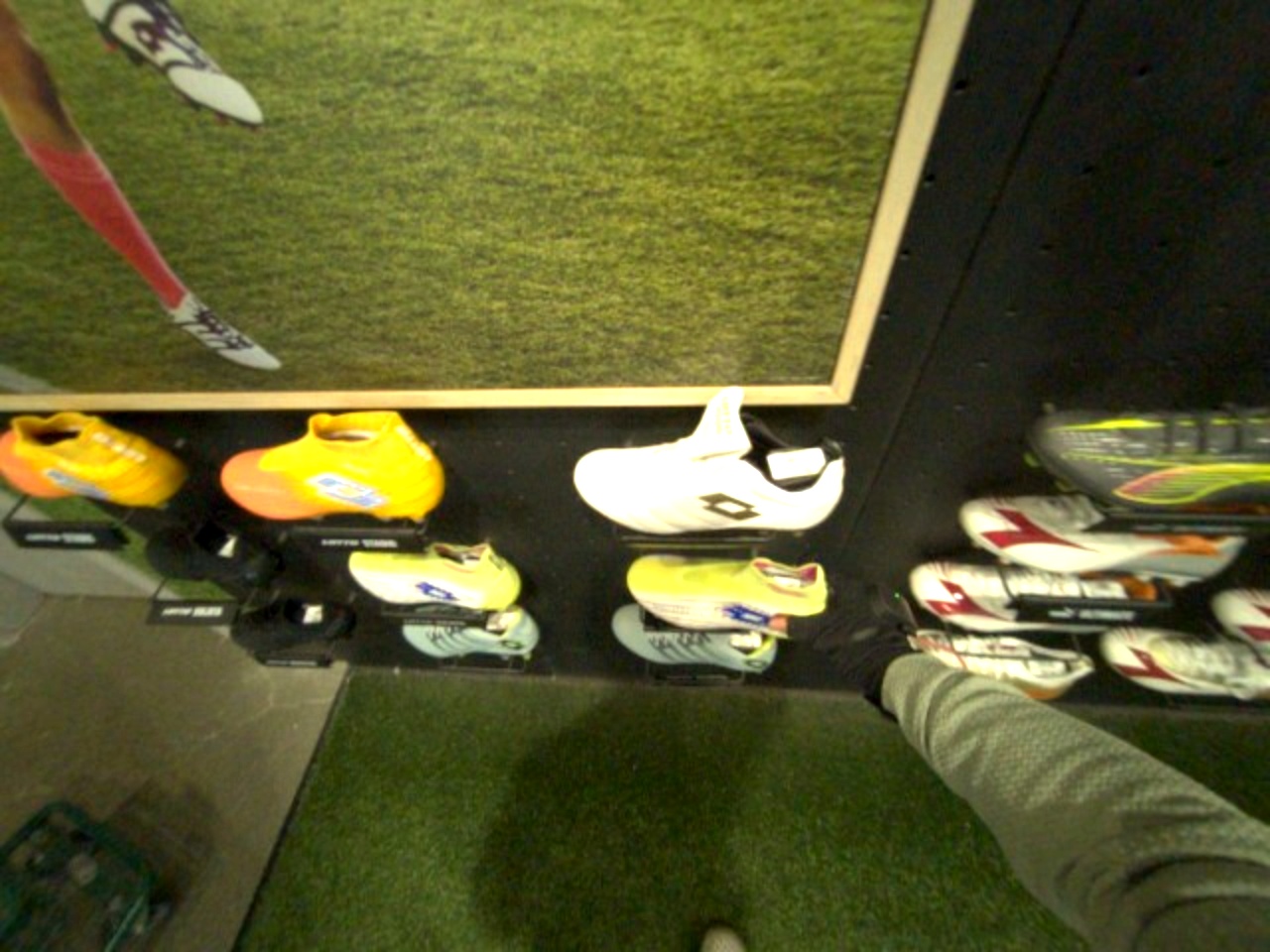} &
            \sqimgright{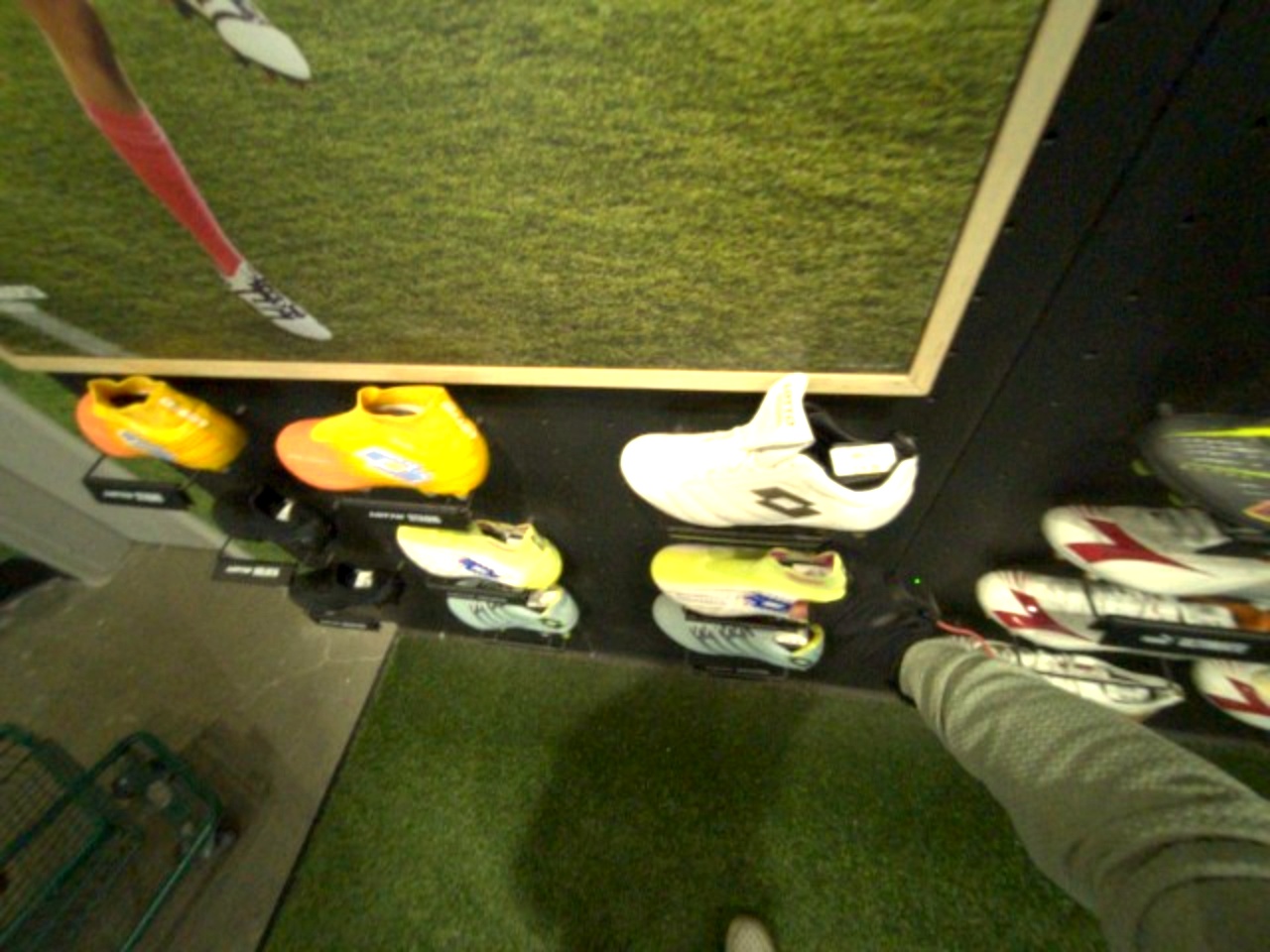} &
            \sqimgright{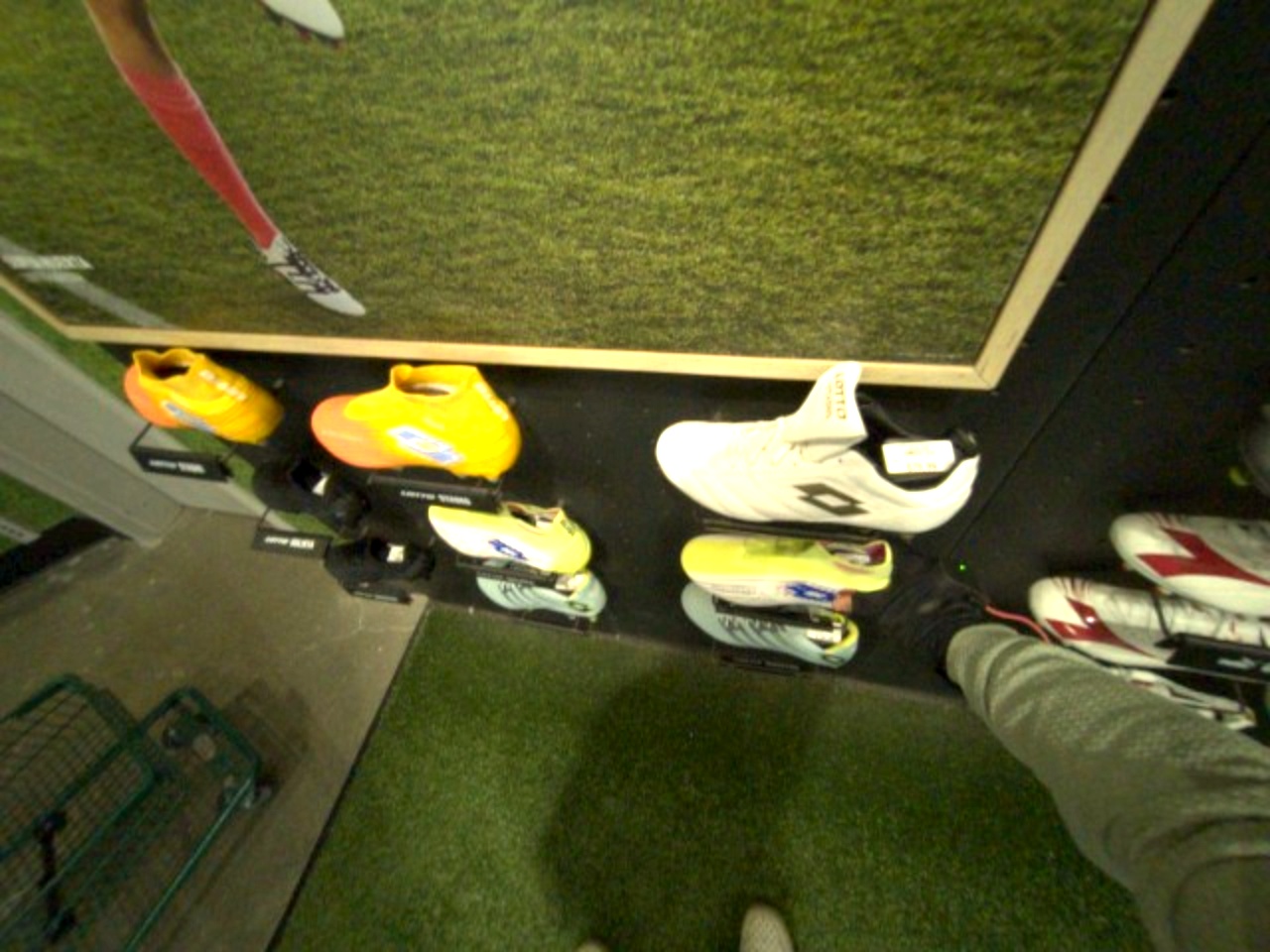} &
            \sqimgcenter{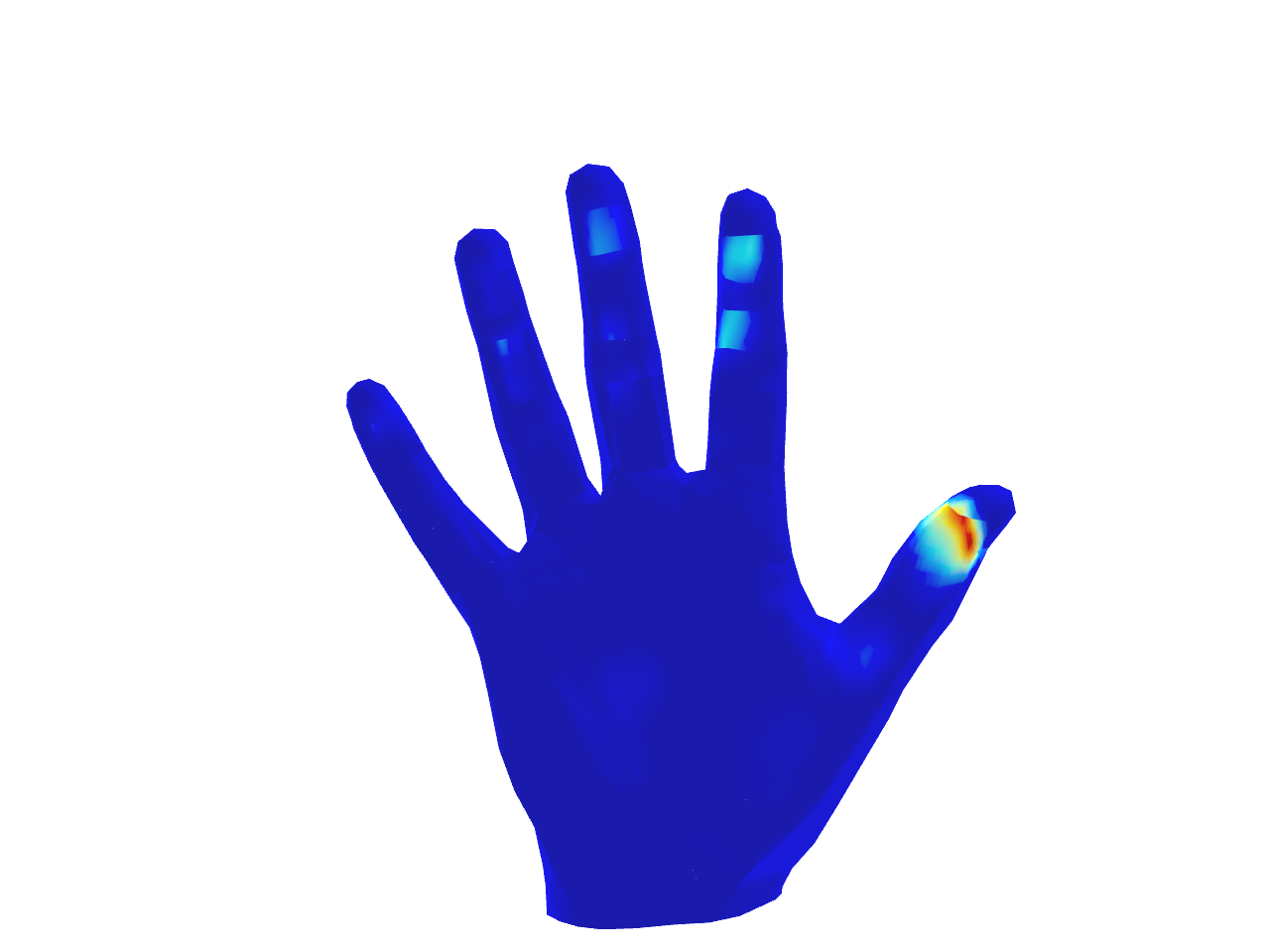} &
            \sqimgcenter{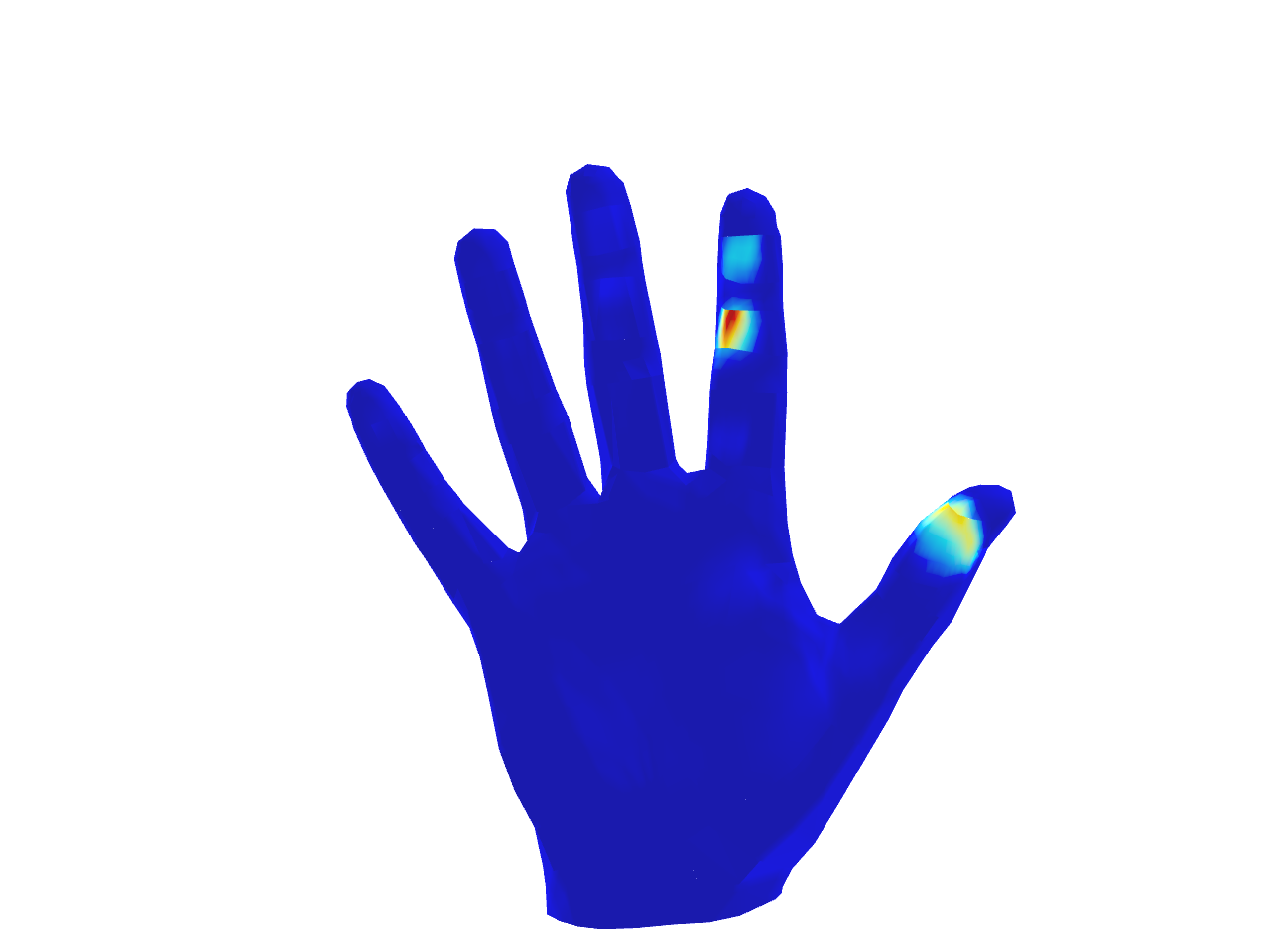} &
            \sqimgcenter{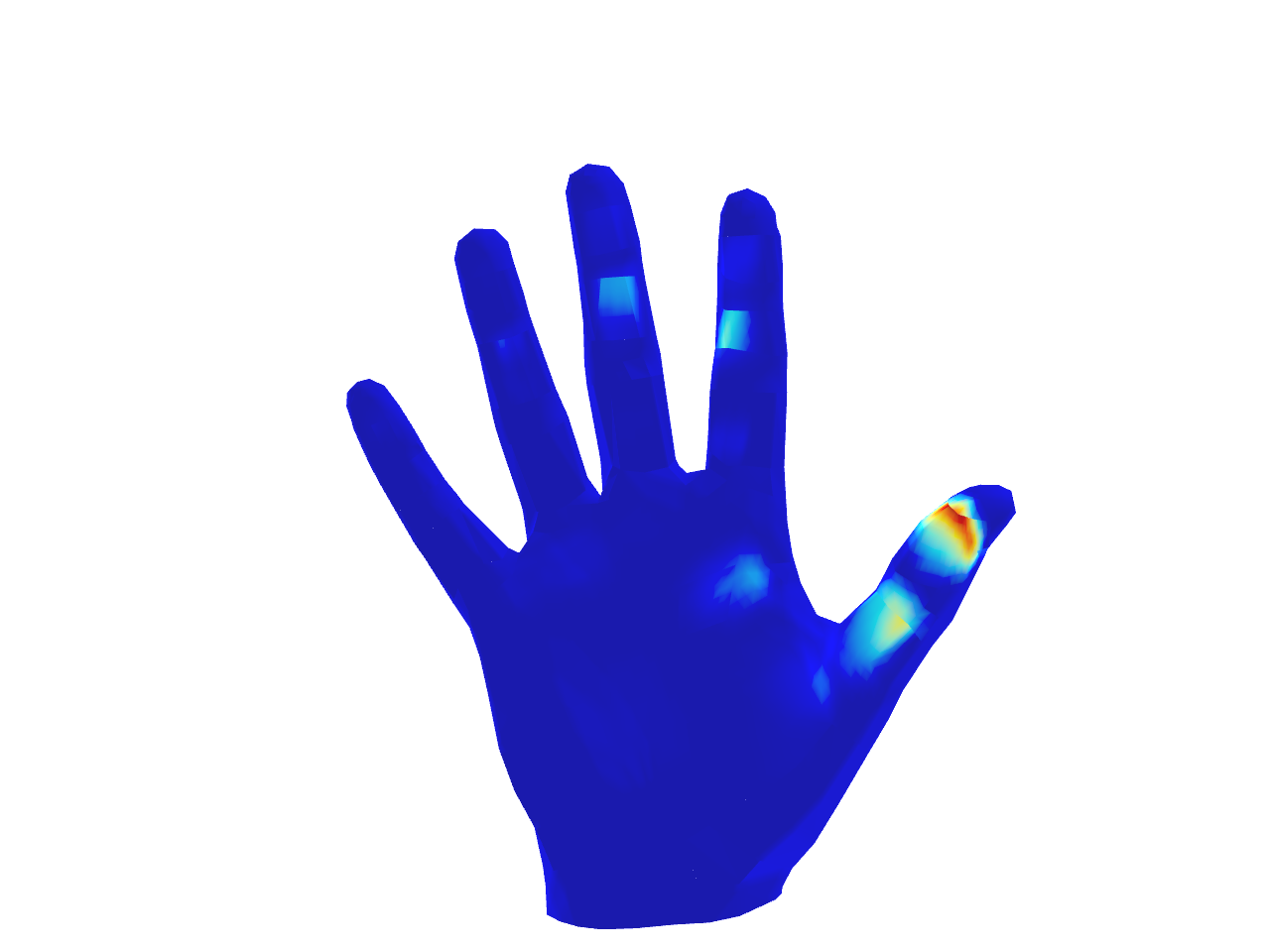} &
            \sqimgcenter{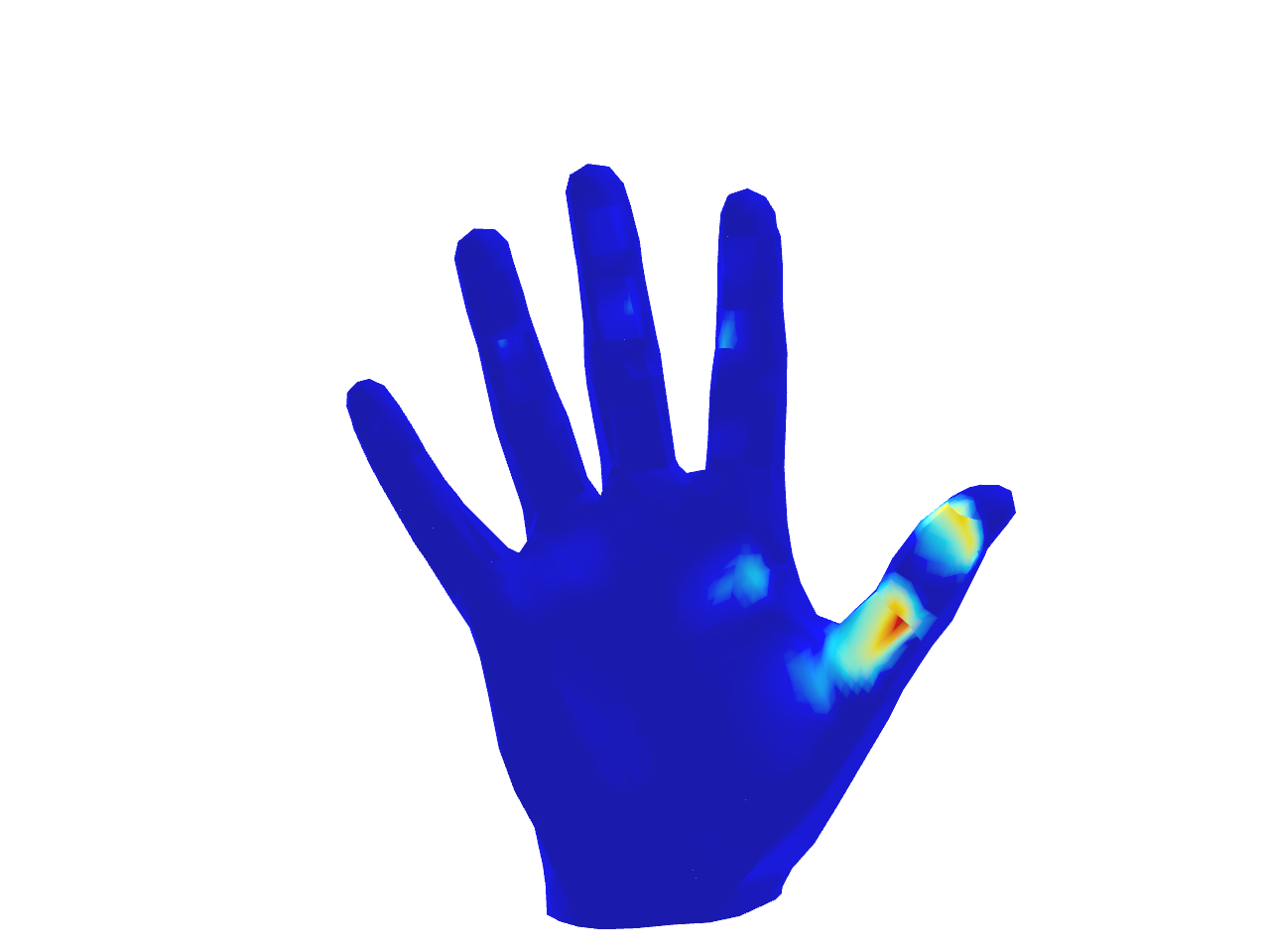} &
            \sqimgcenter{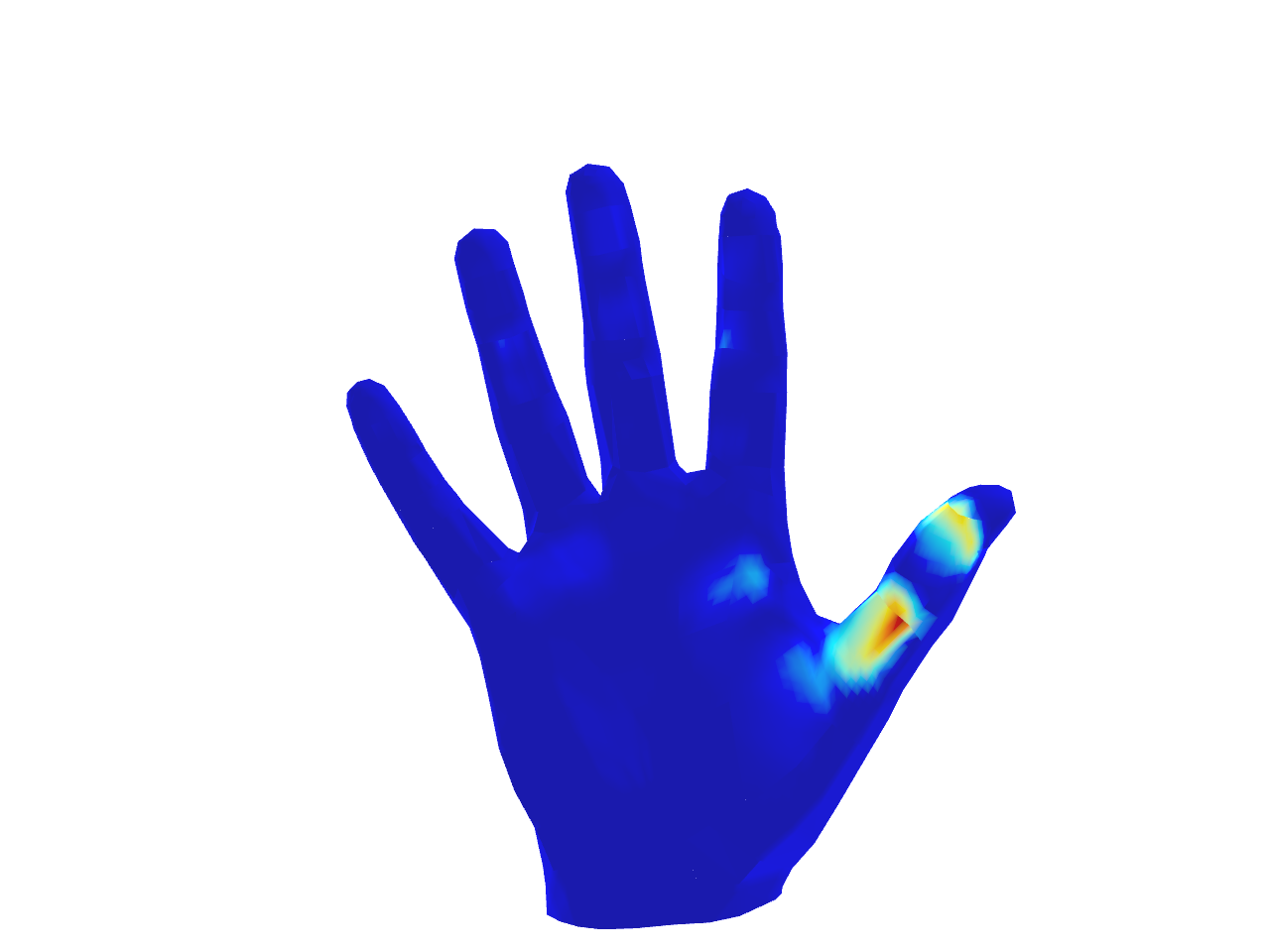} \\

        \rowlabel{gt} &
            \sqimgcenter{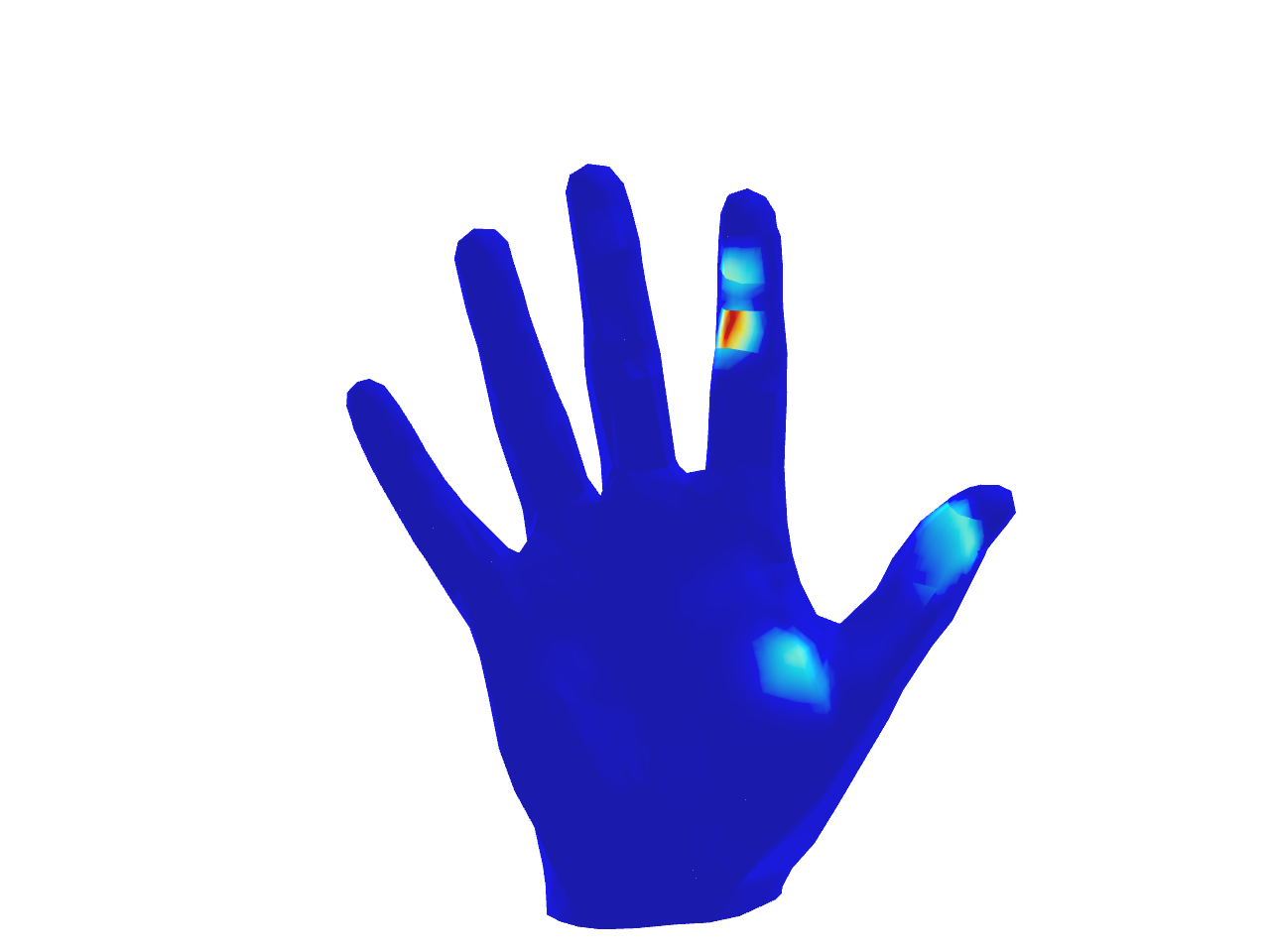} &
            \sqimgcenter{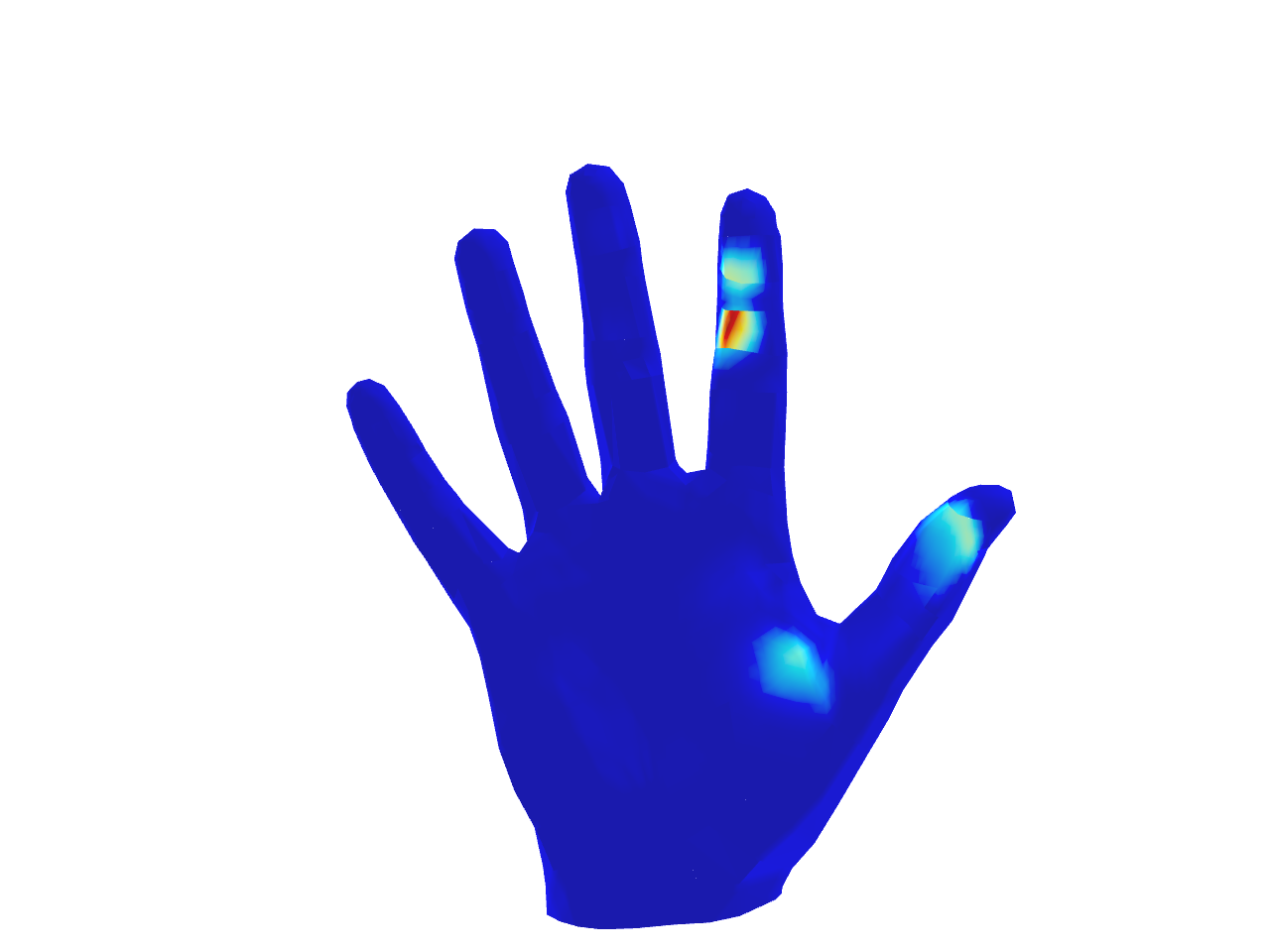} &
            \sqimgcenter{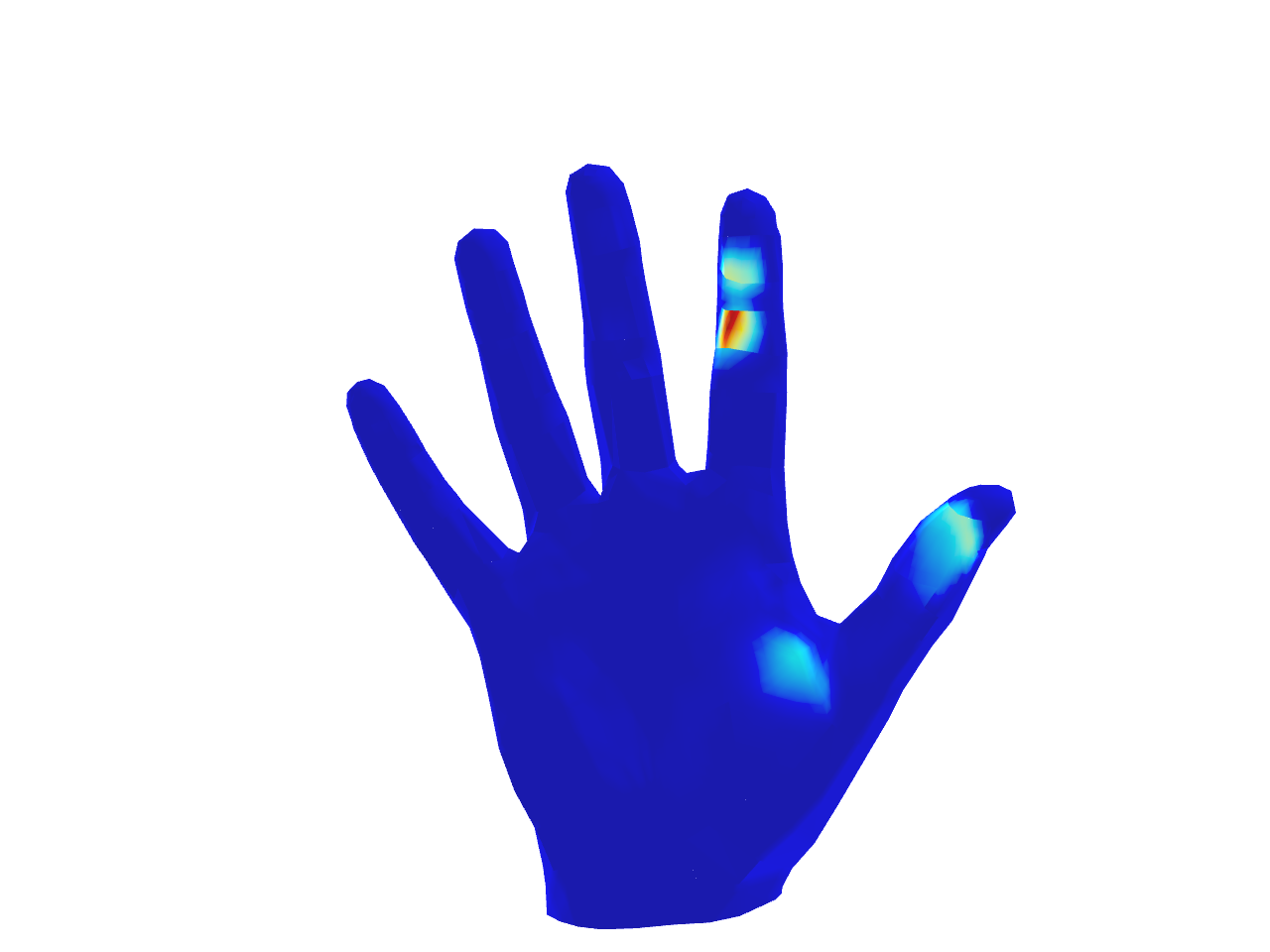} &
            \sqimgcenter{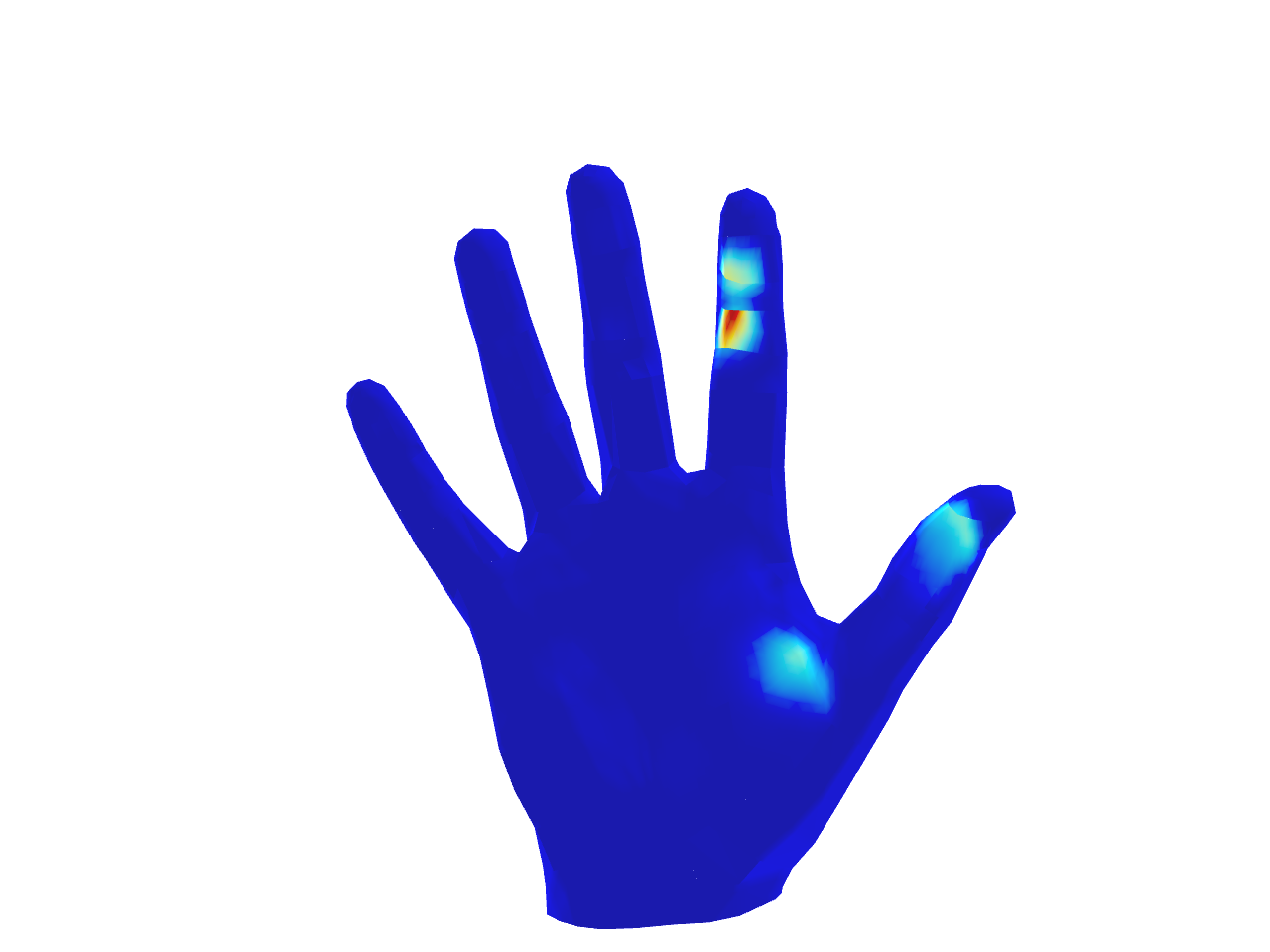} &
            \sqimgcenter{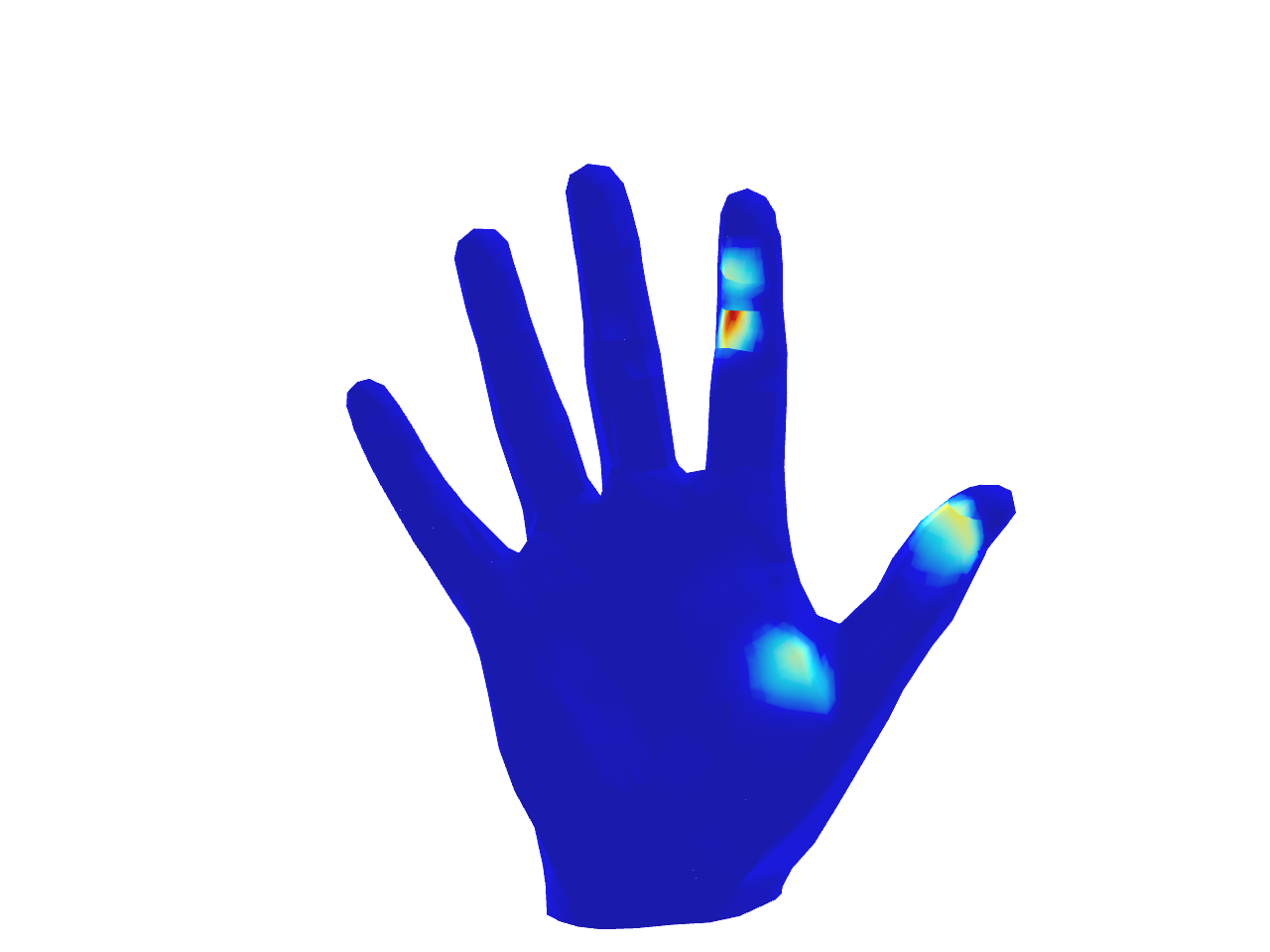} &
            \sqimgright{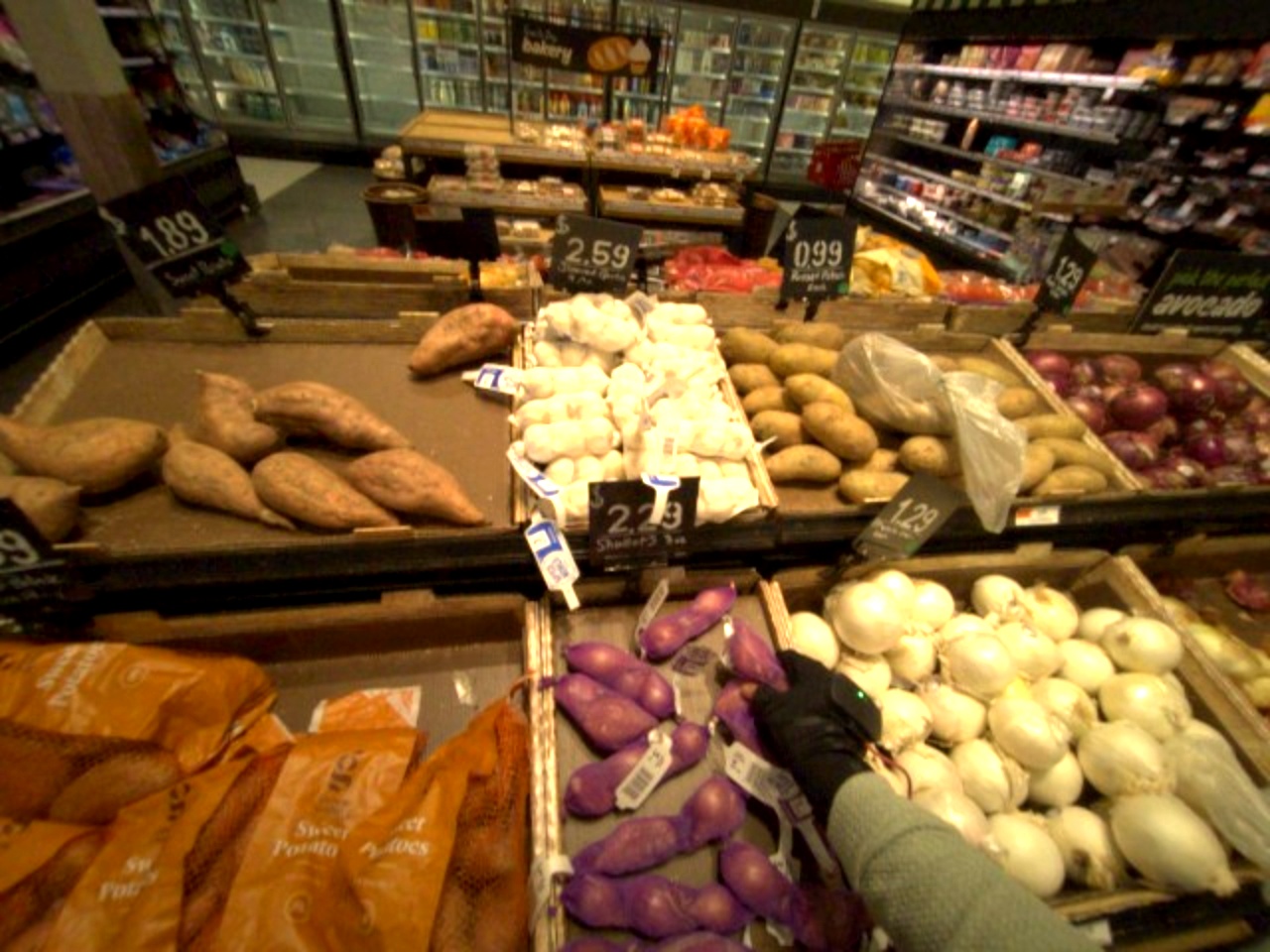} &
            \sqimgright{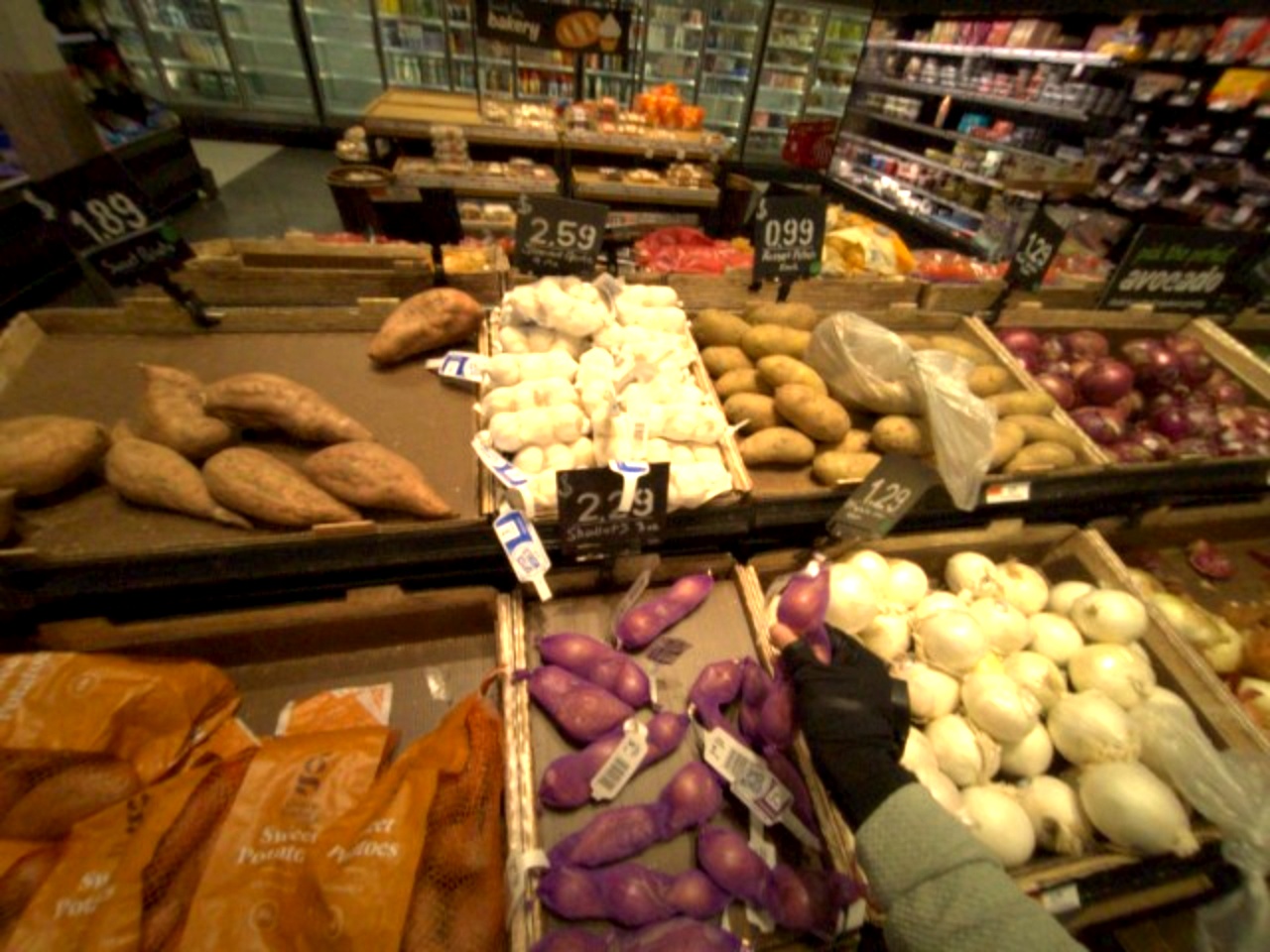} &
            \sqimgright{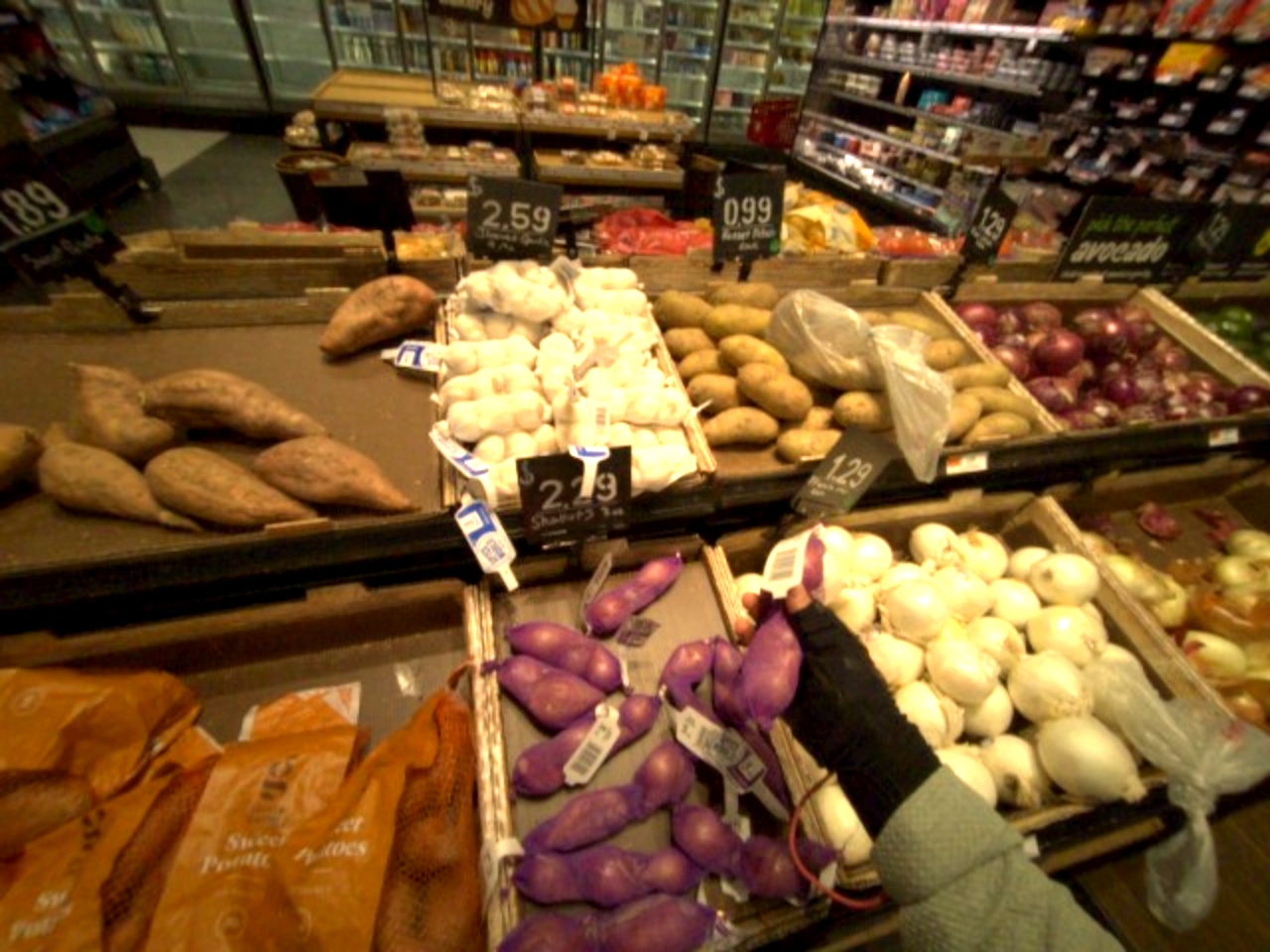} &
            \sqimgright{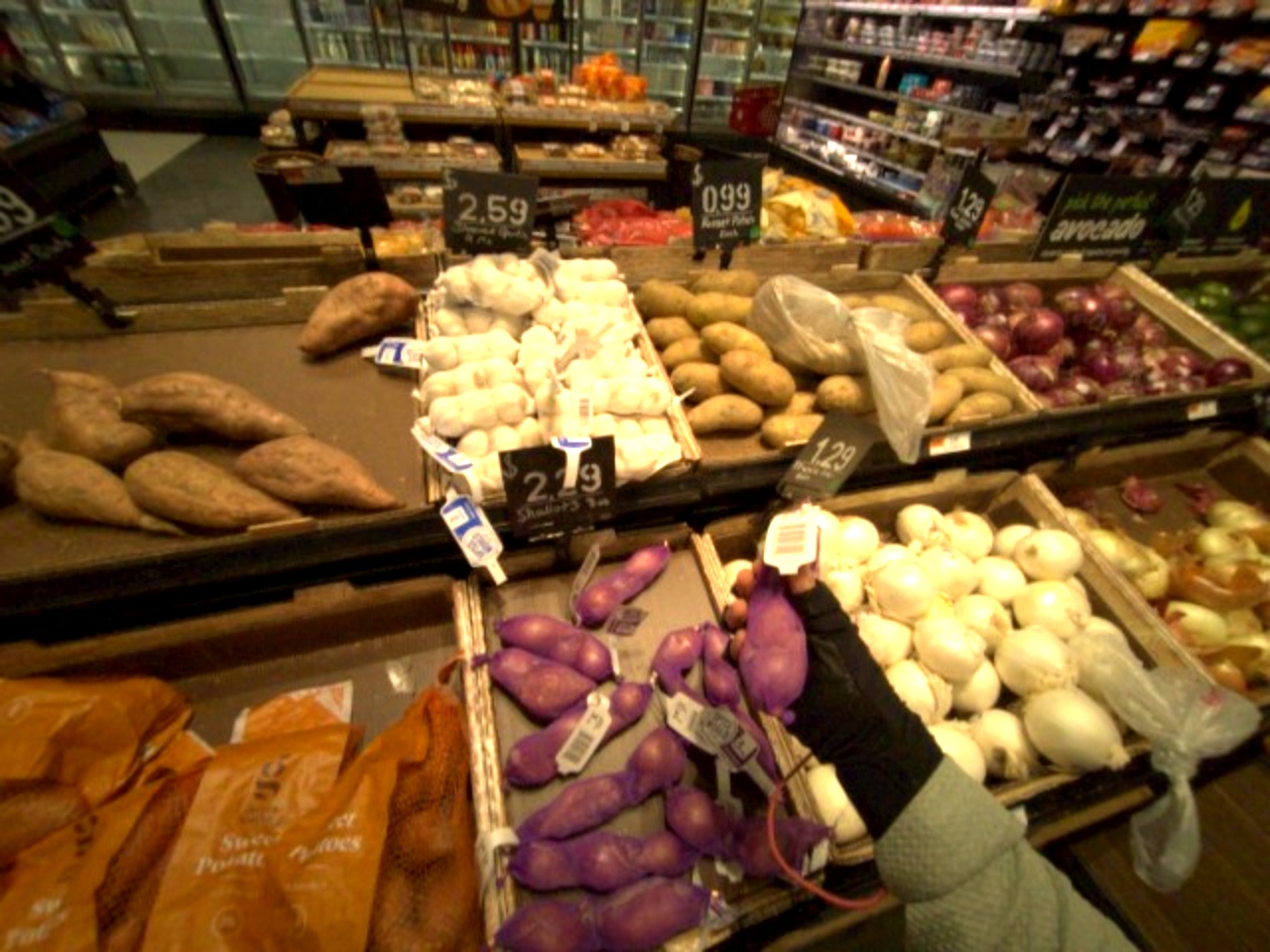} &
            \sqimgright{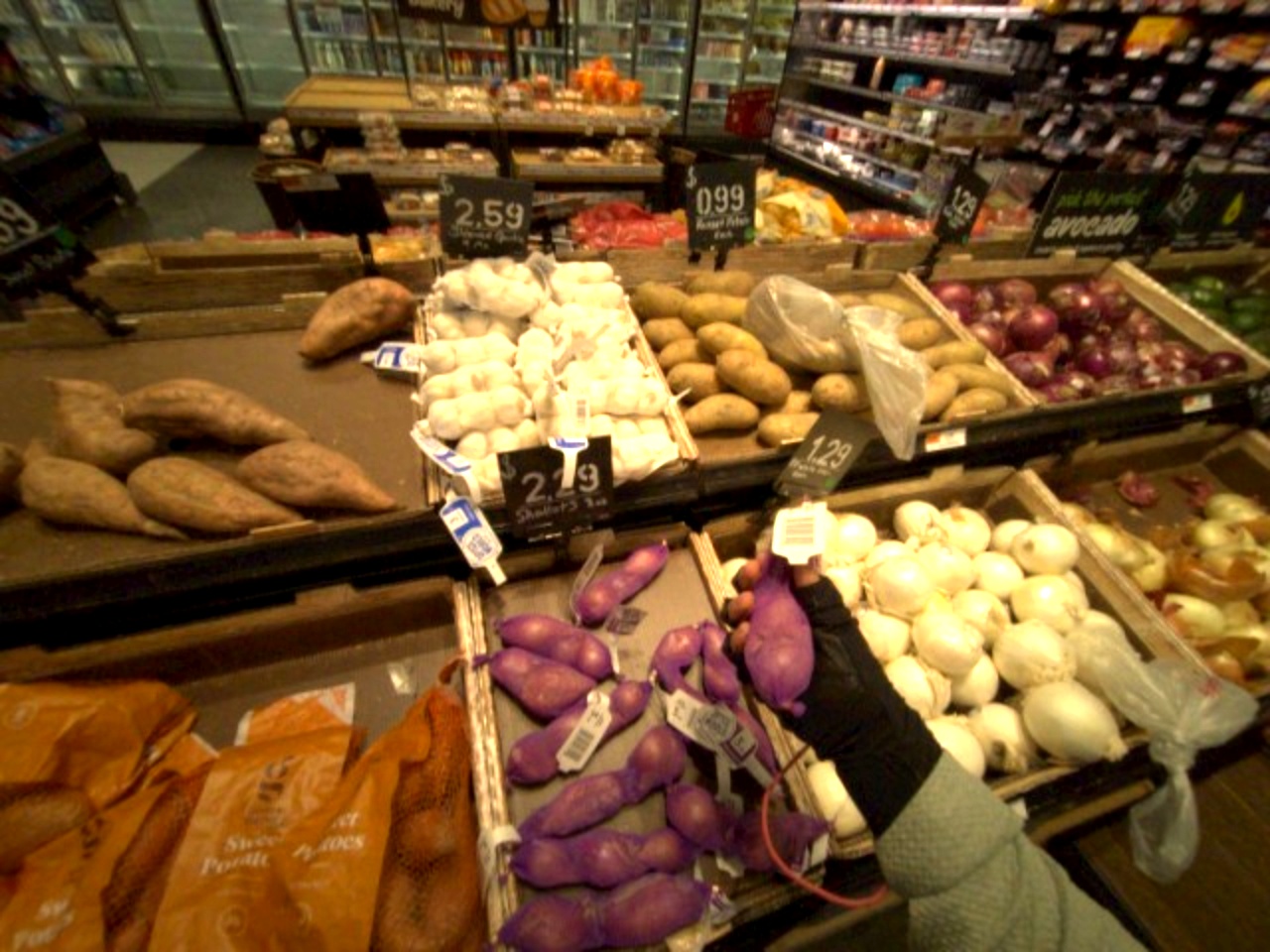} \\

        \rowlabel{result} &
            \sqimgcenter{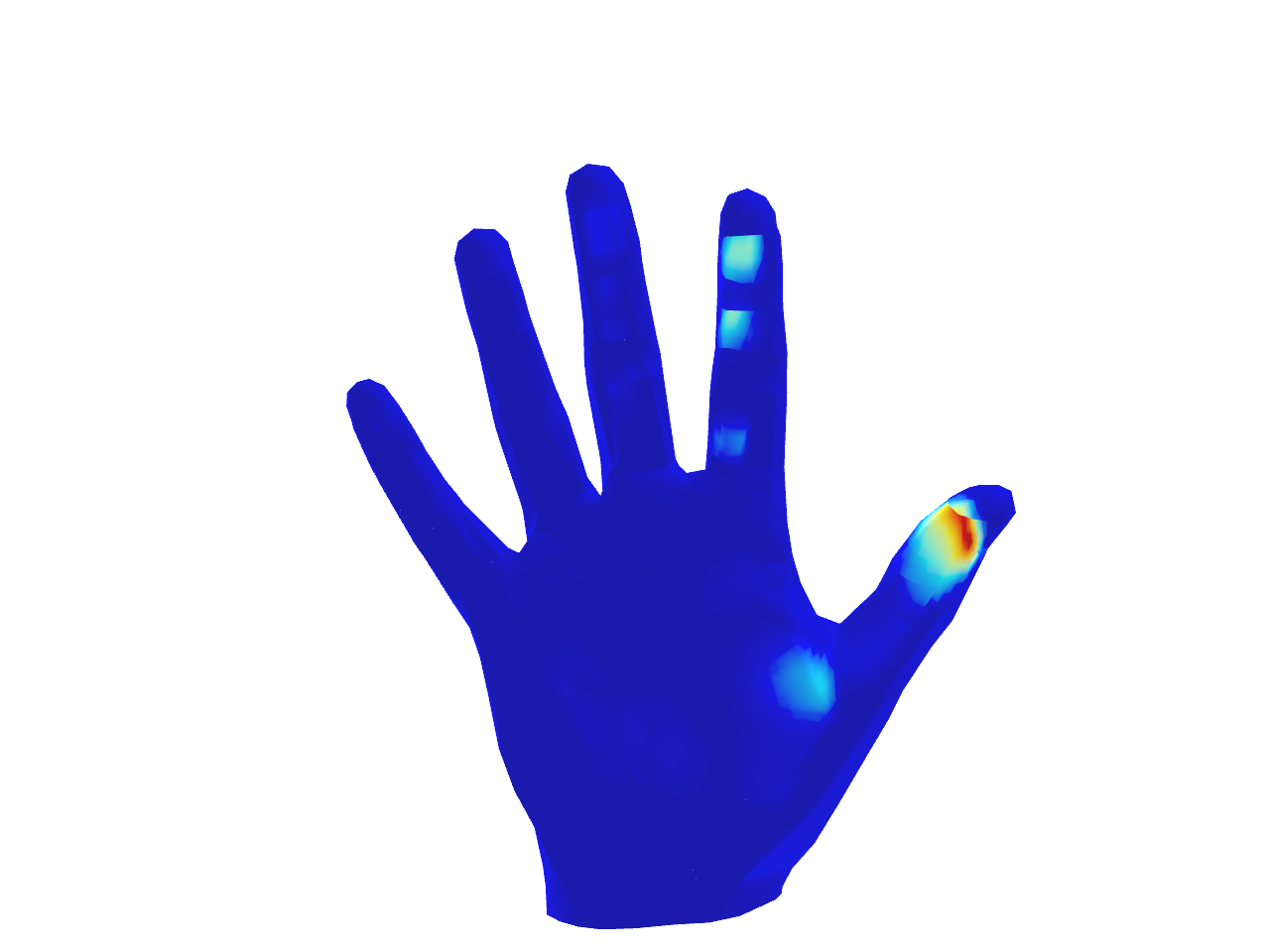} &
            \sqimgcenter{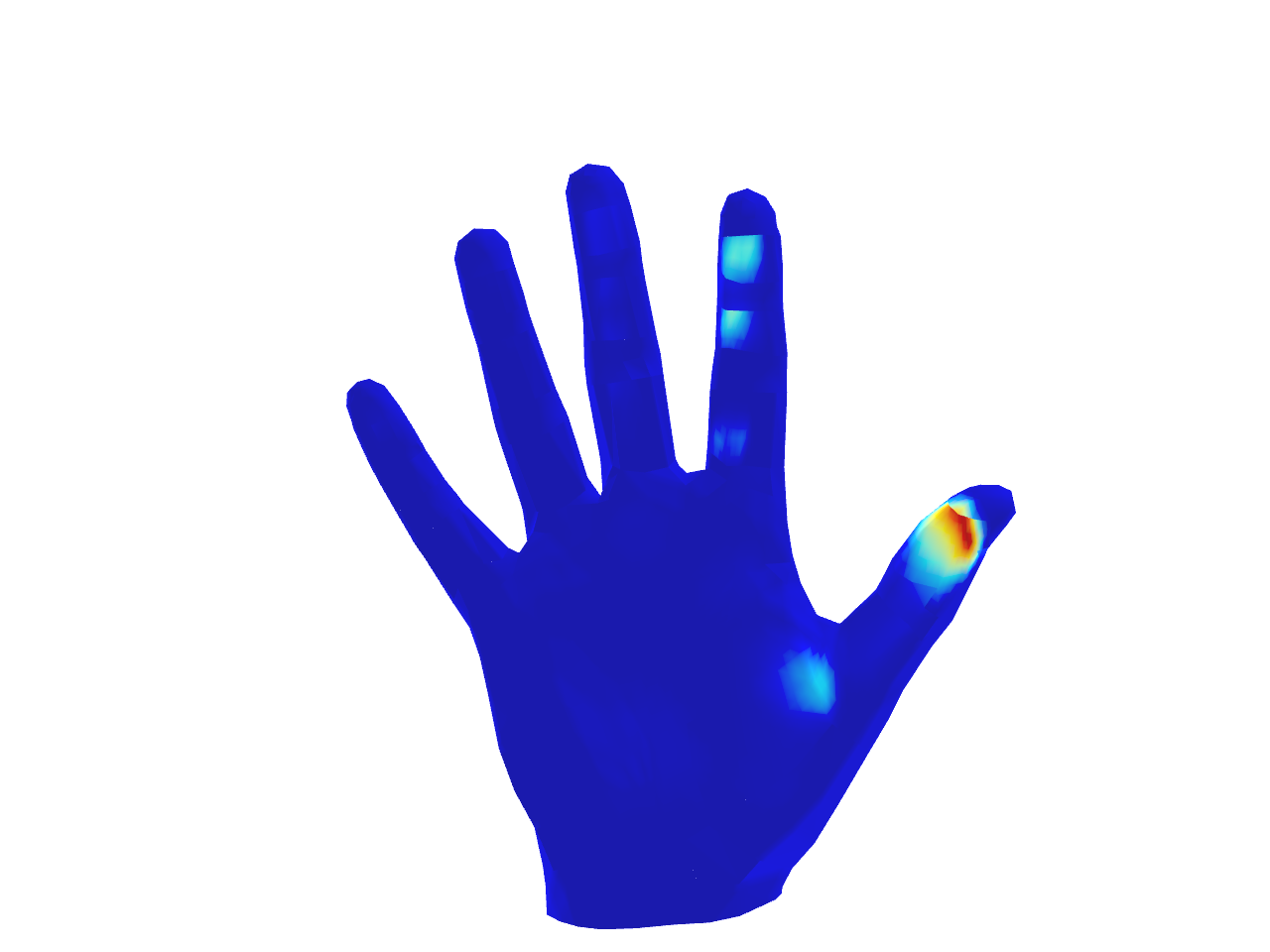} &
            \sqimgcenter{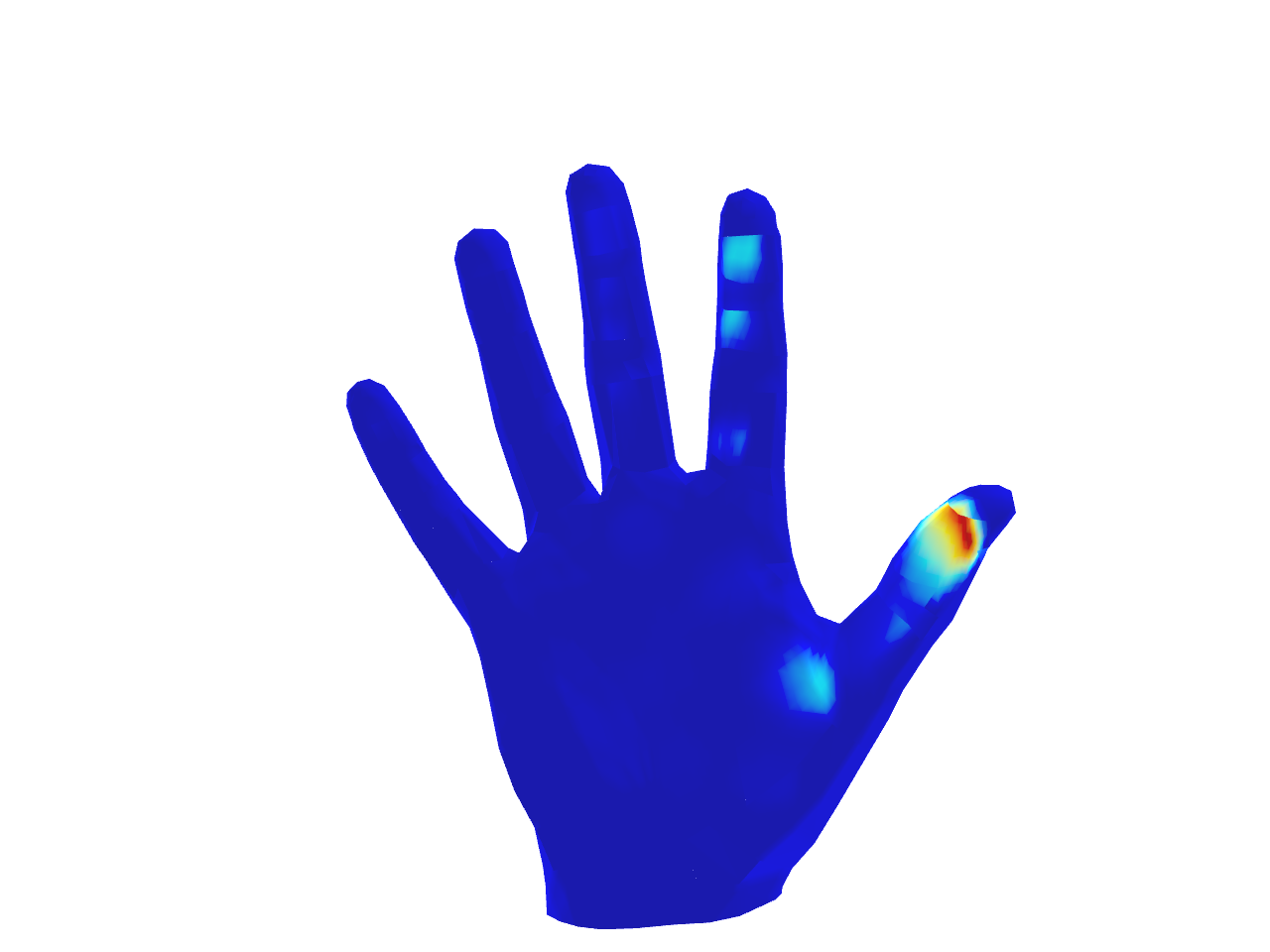} &
            \sqimgcenter{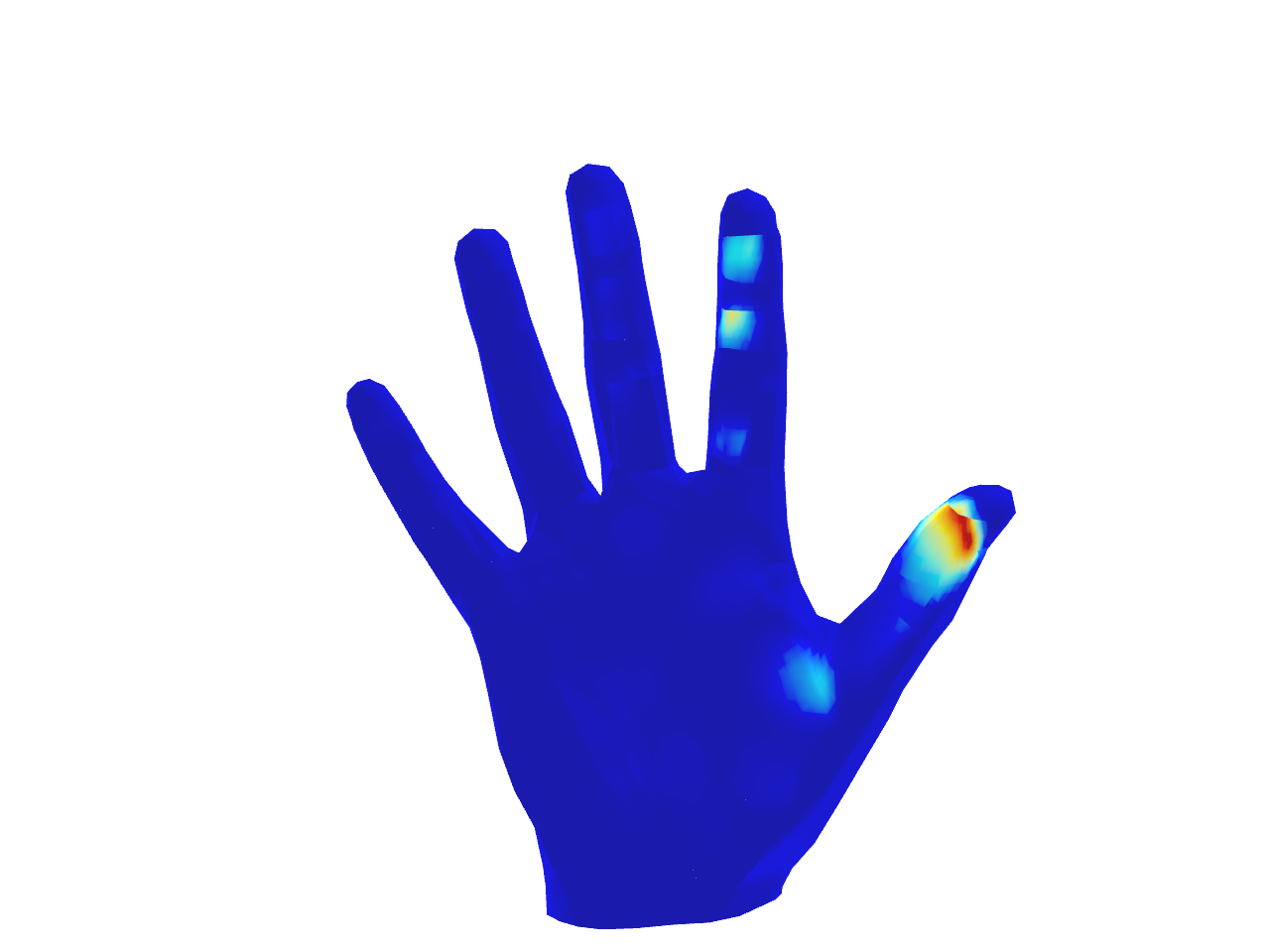} &
            \sqimgcenter{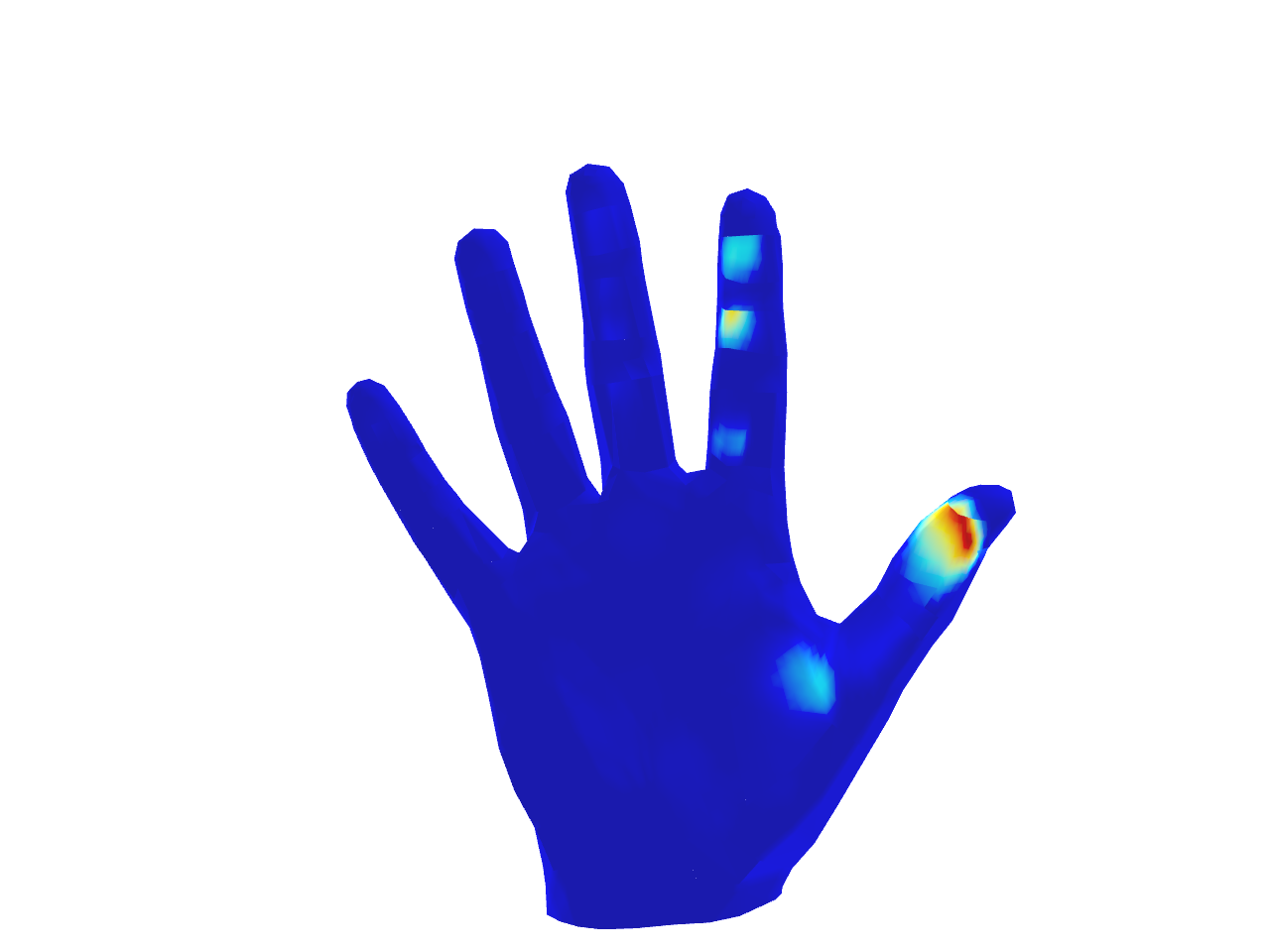} &
            \sqimgright{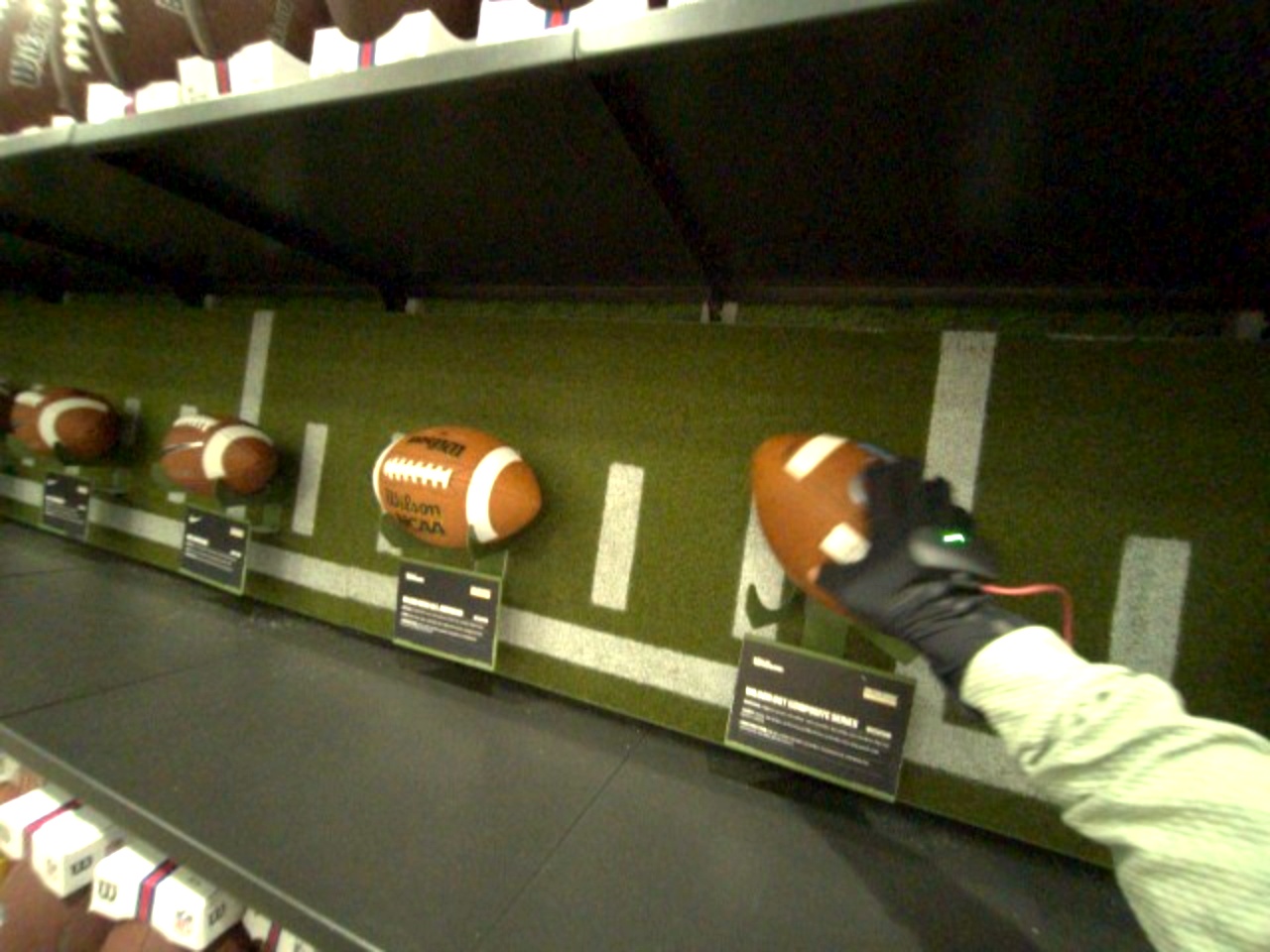} &
            \sqimgright{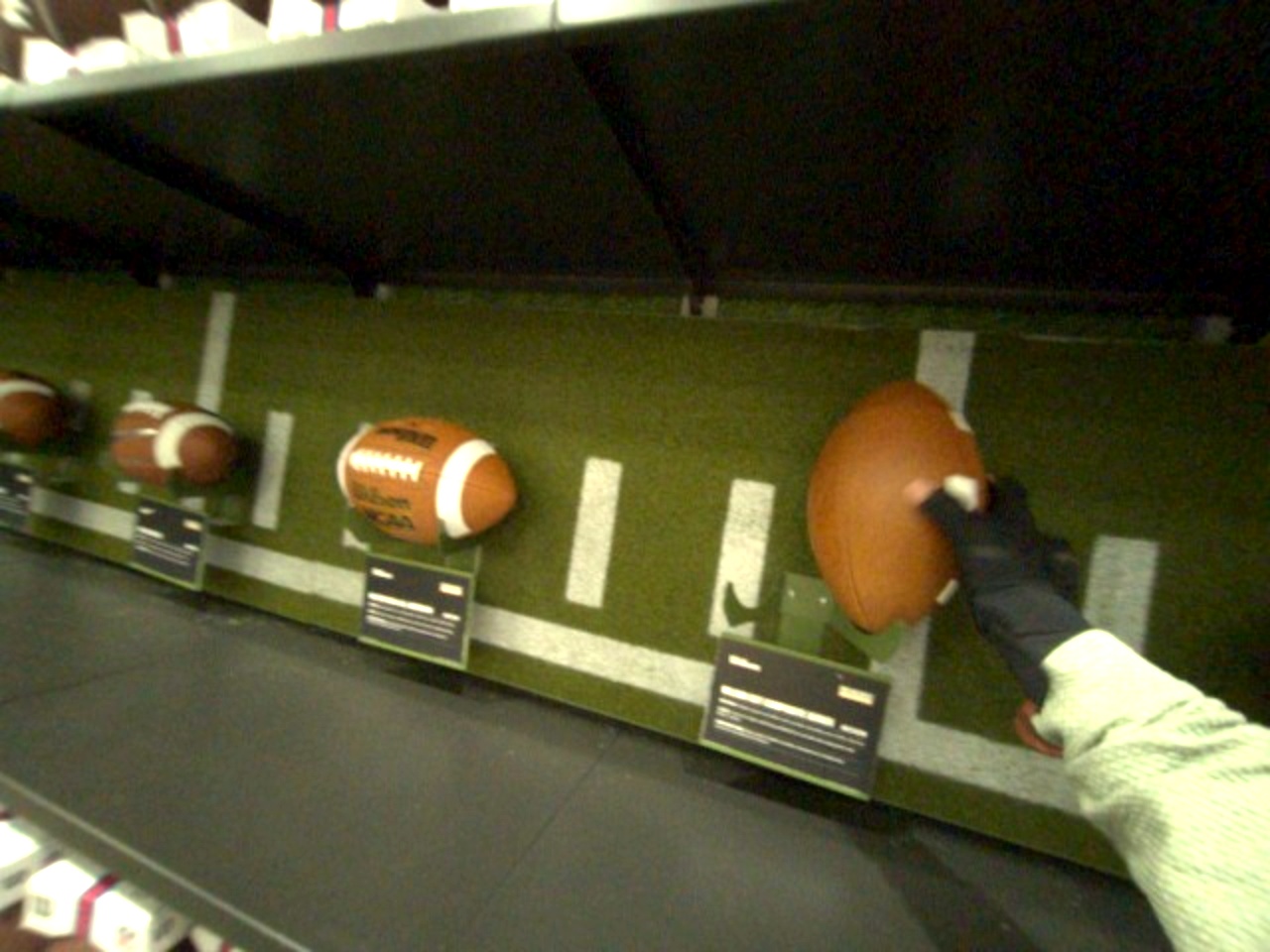} &
            \sqimgright{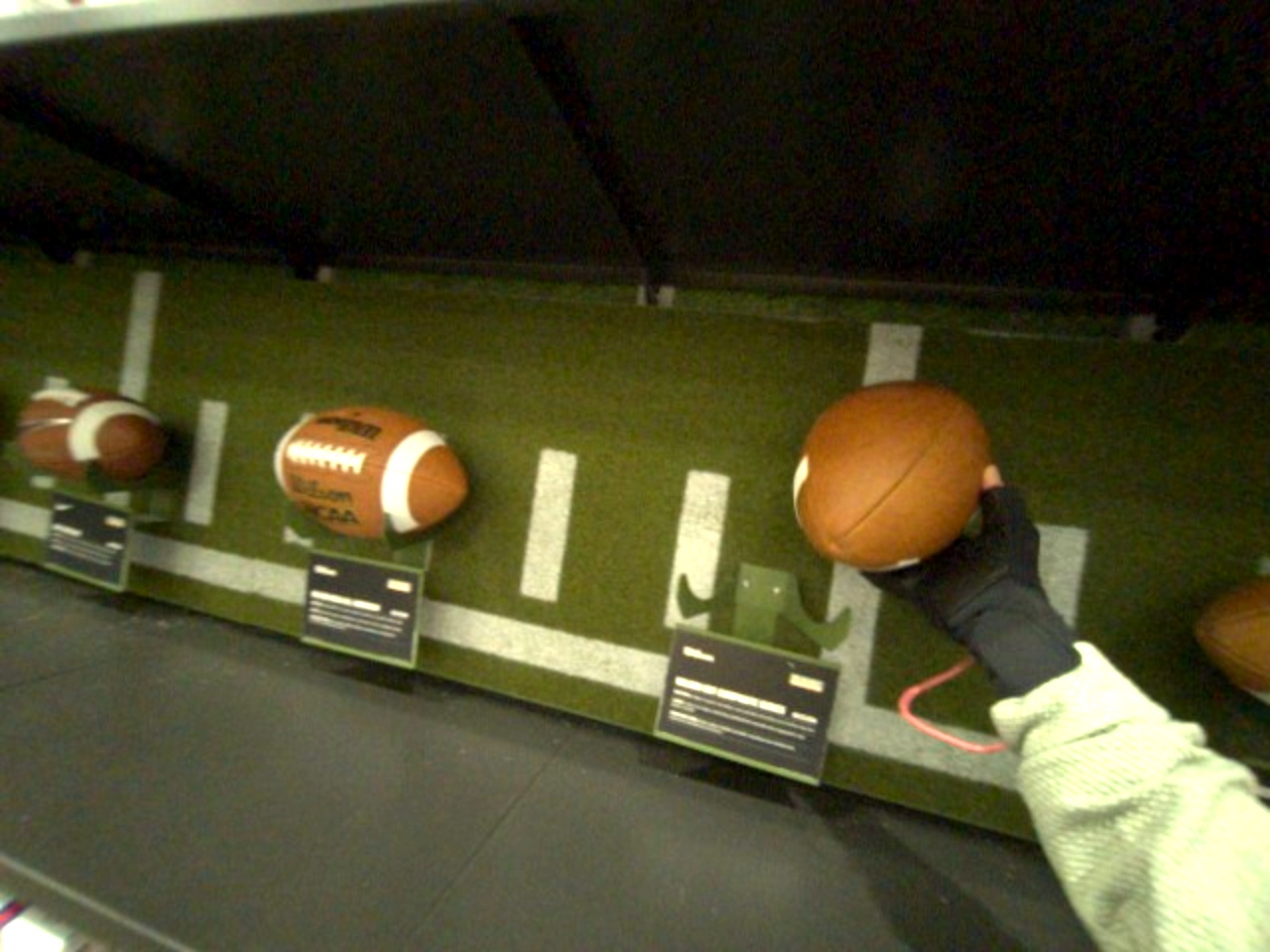} &
            \sqimgright{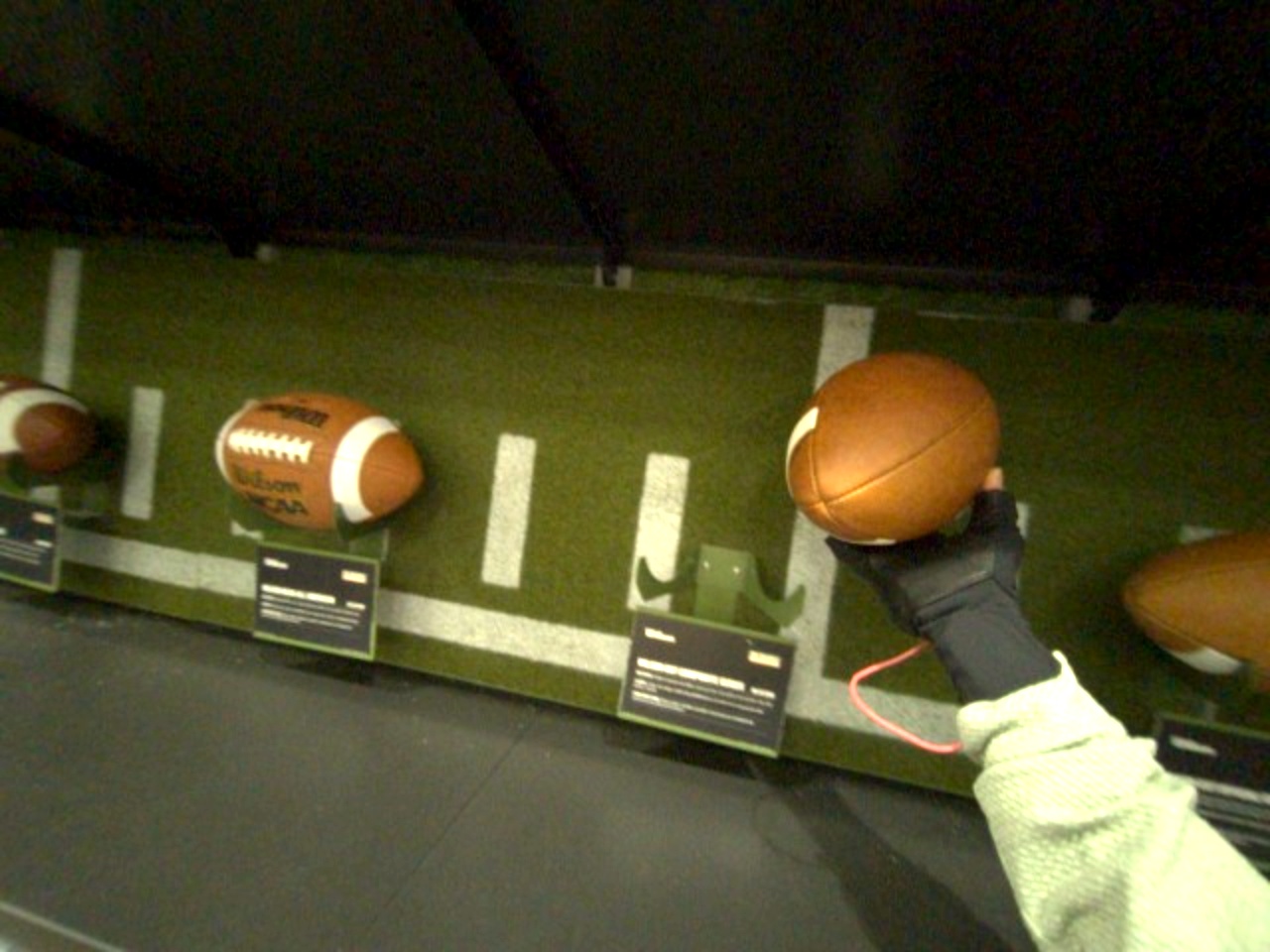} &
            \sqimgright{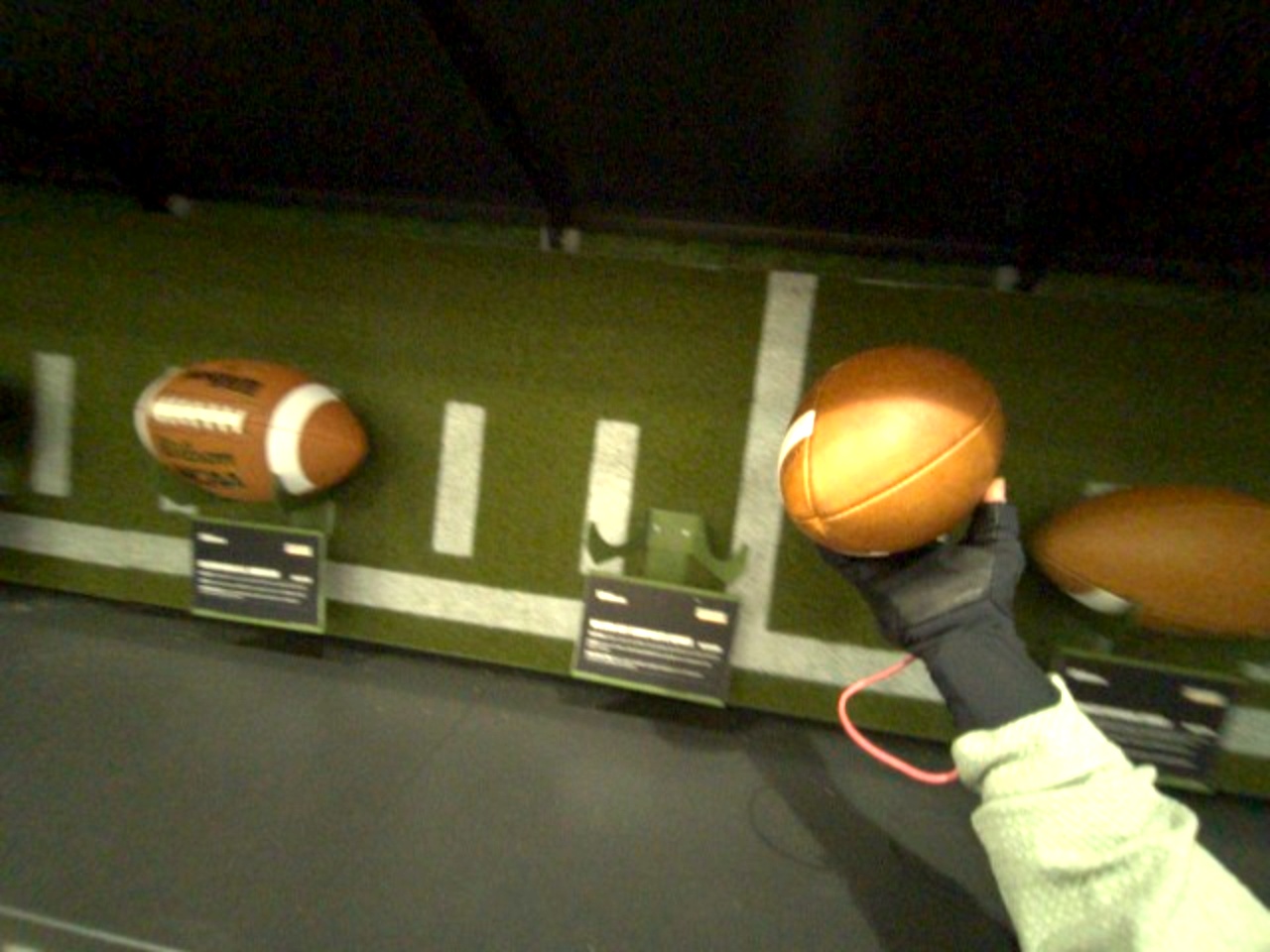} \\

        \\[1.2ex] 
        \rowlabel{query} &
            \sqimgright{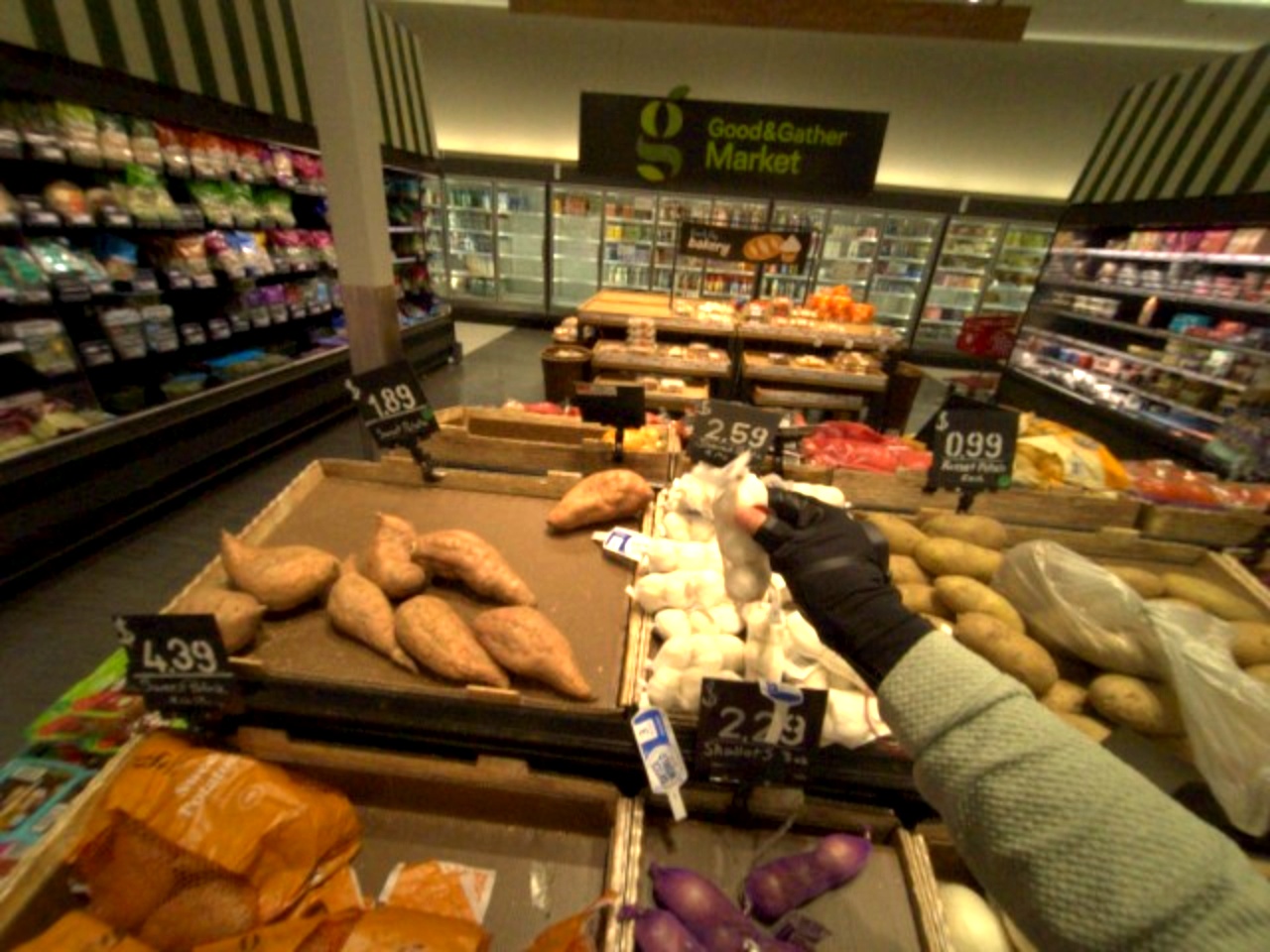} &
            \sqimgright{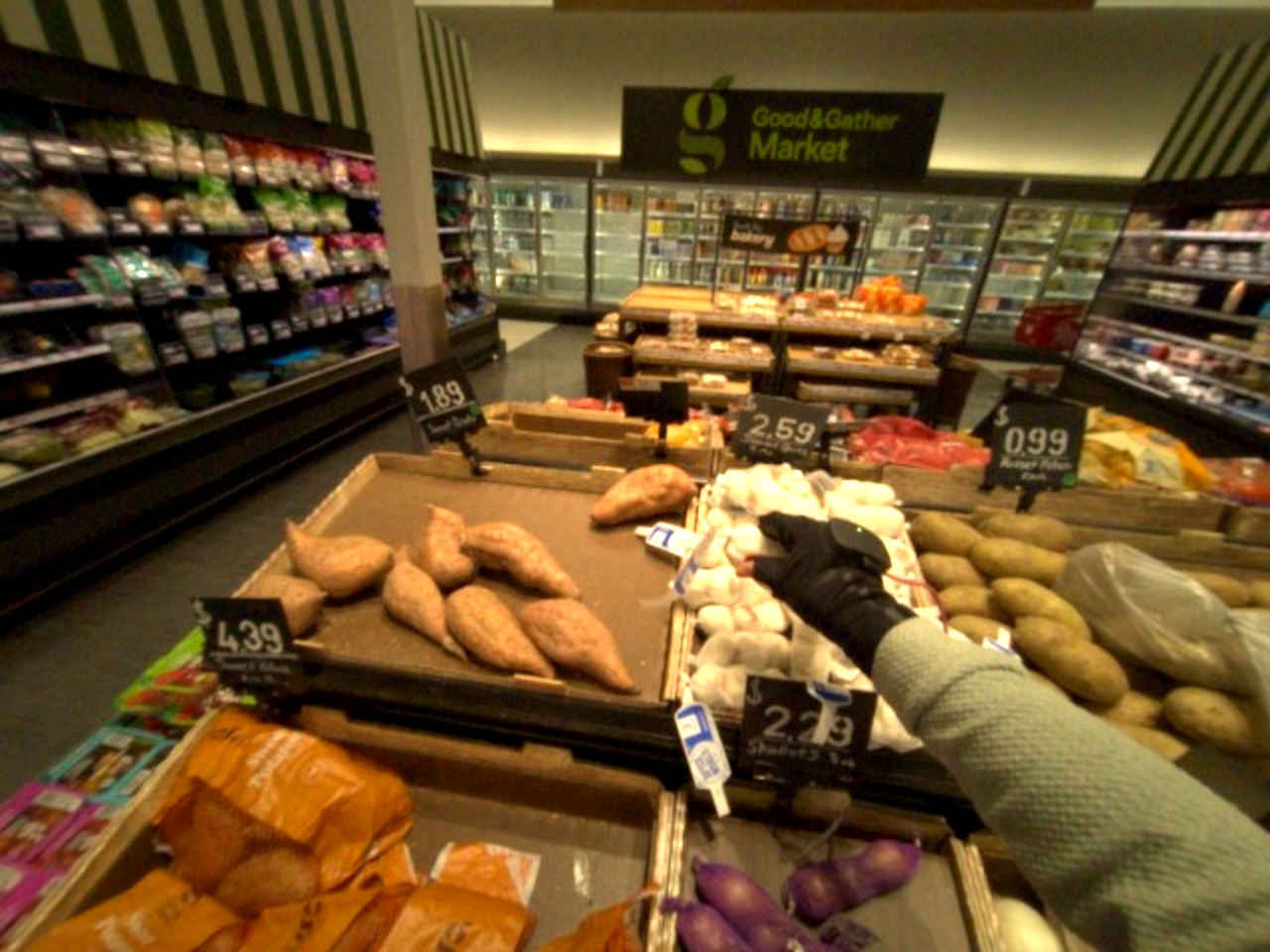} &
            \sqimgright{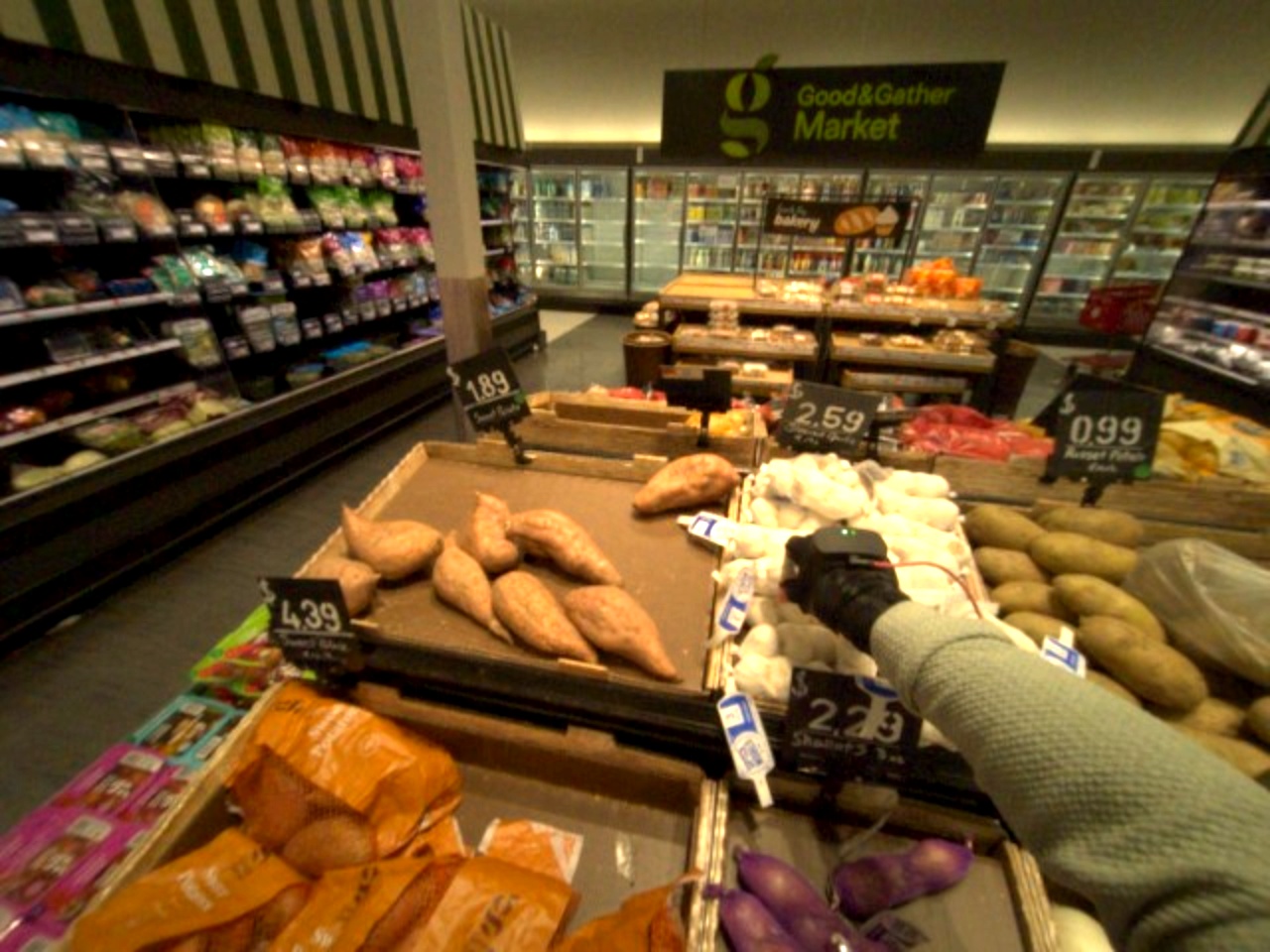} &
            \sqimgright{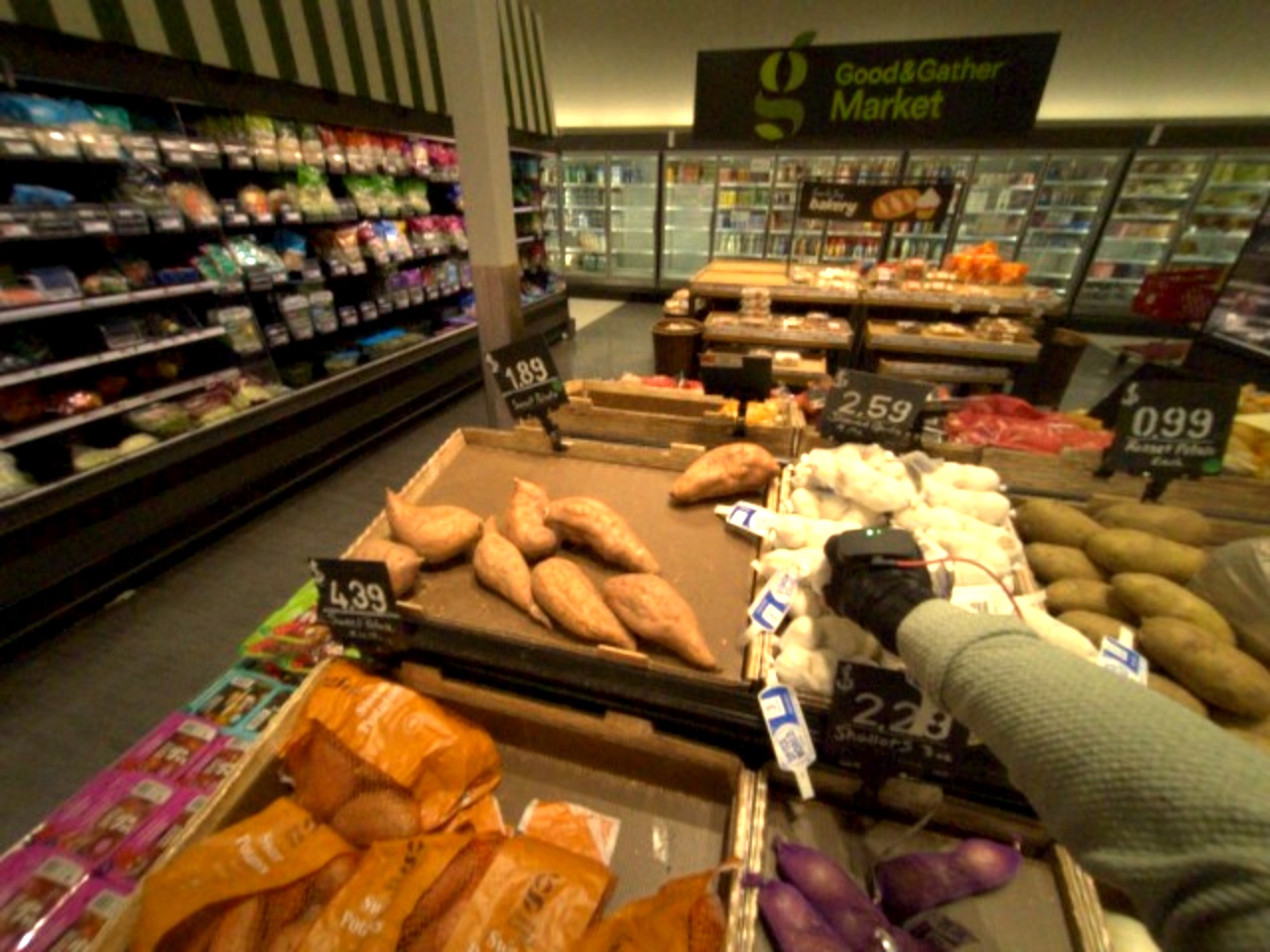} &
            \sqimgright{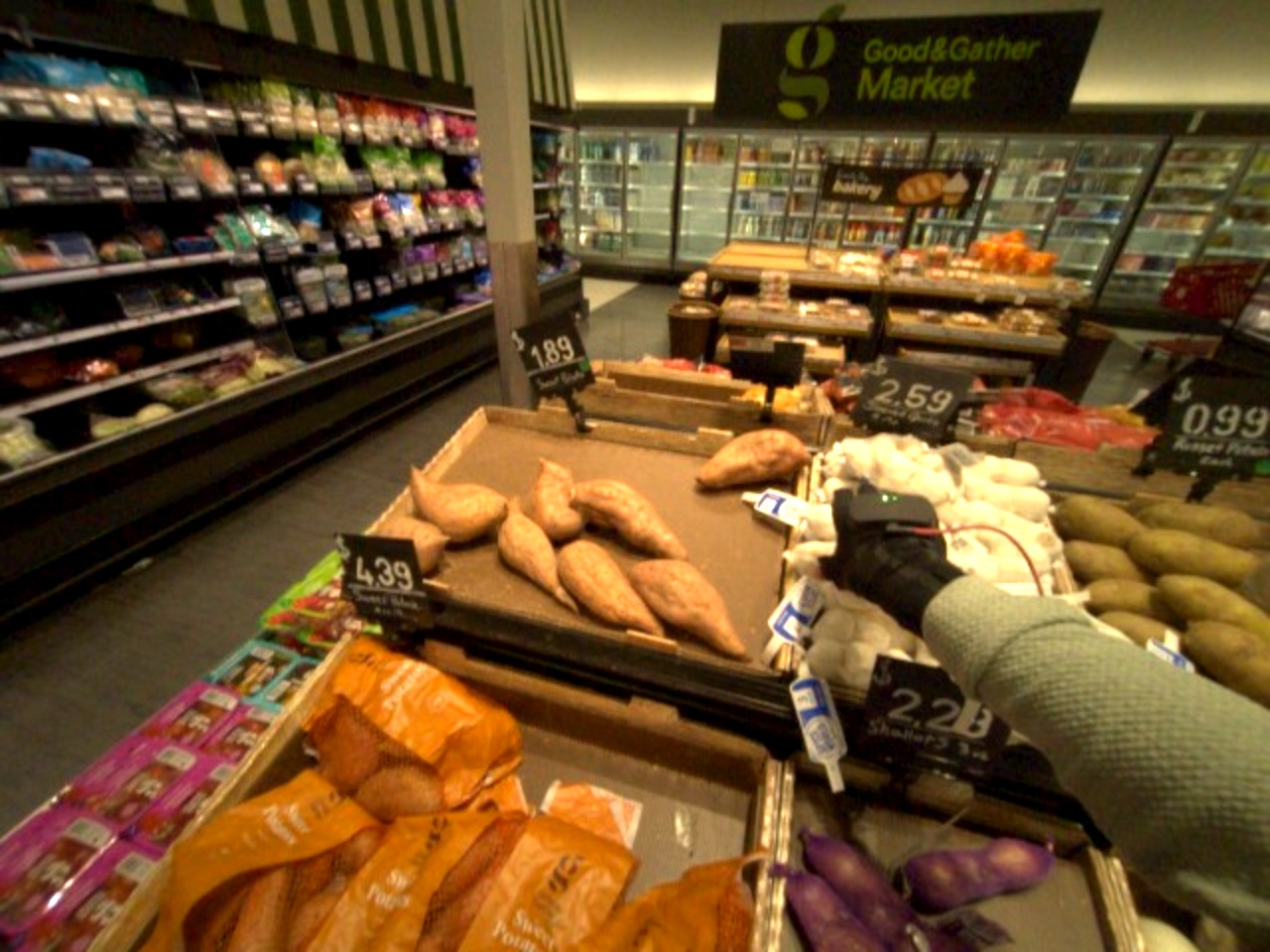} &
            \sqimgcenter{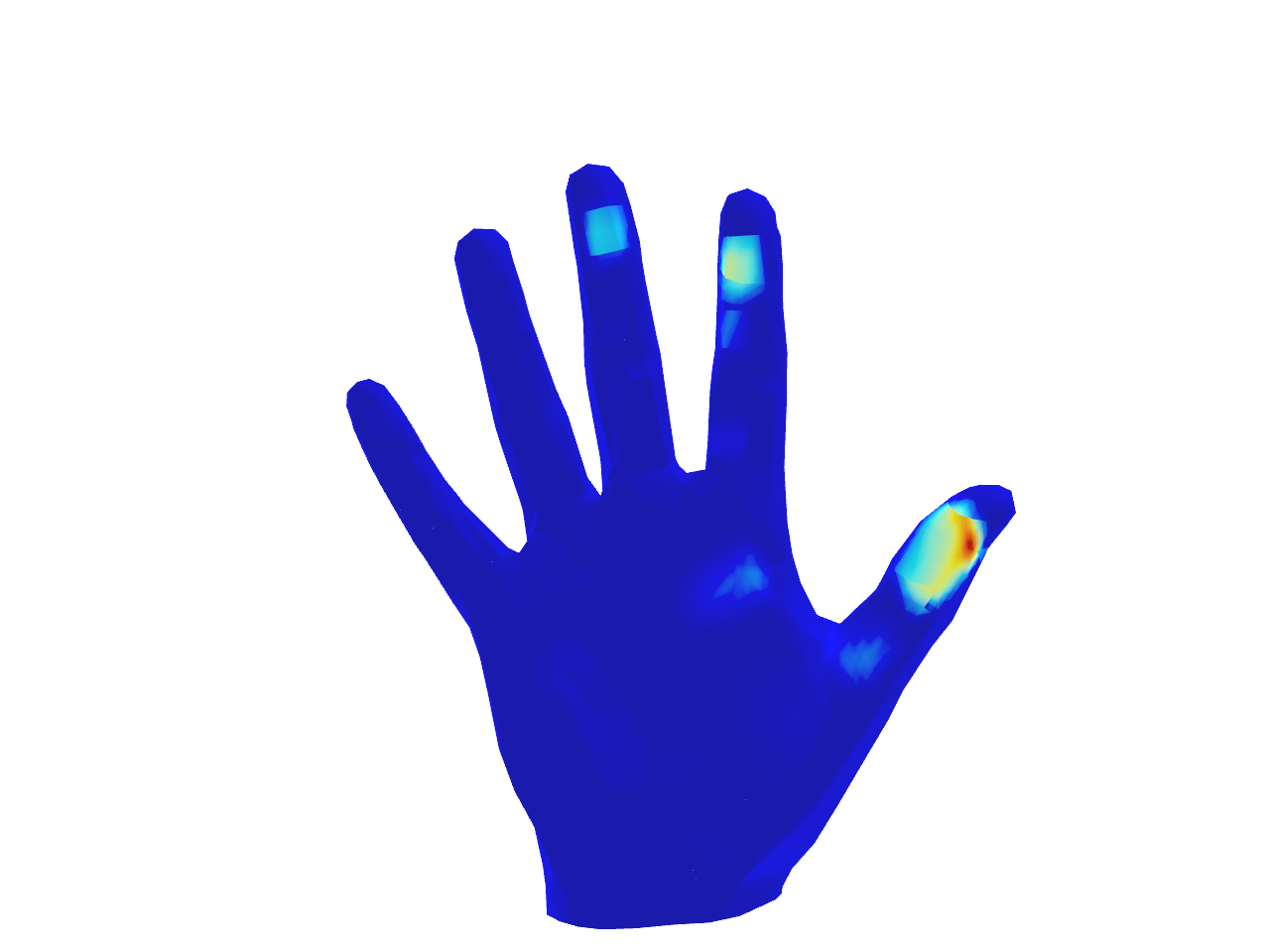} &
            \sqimgcenter{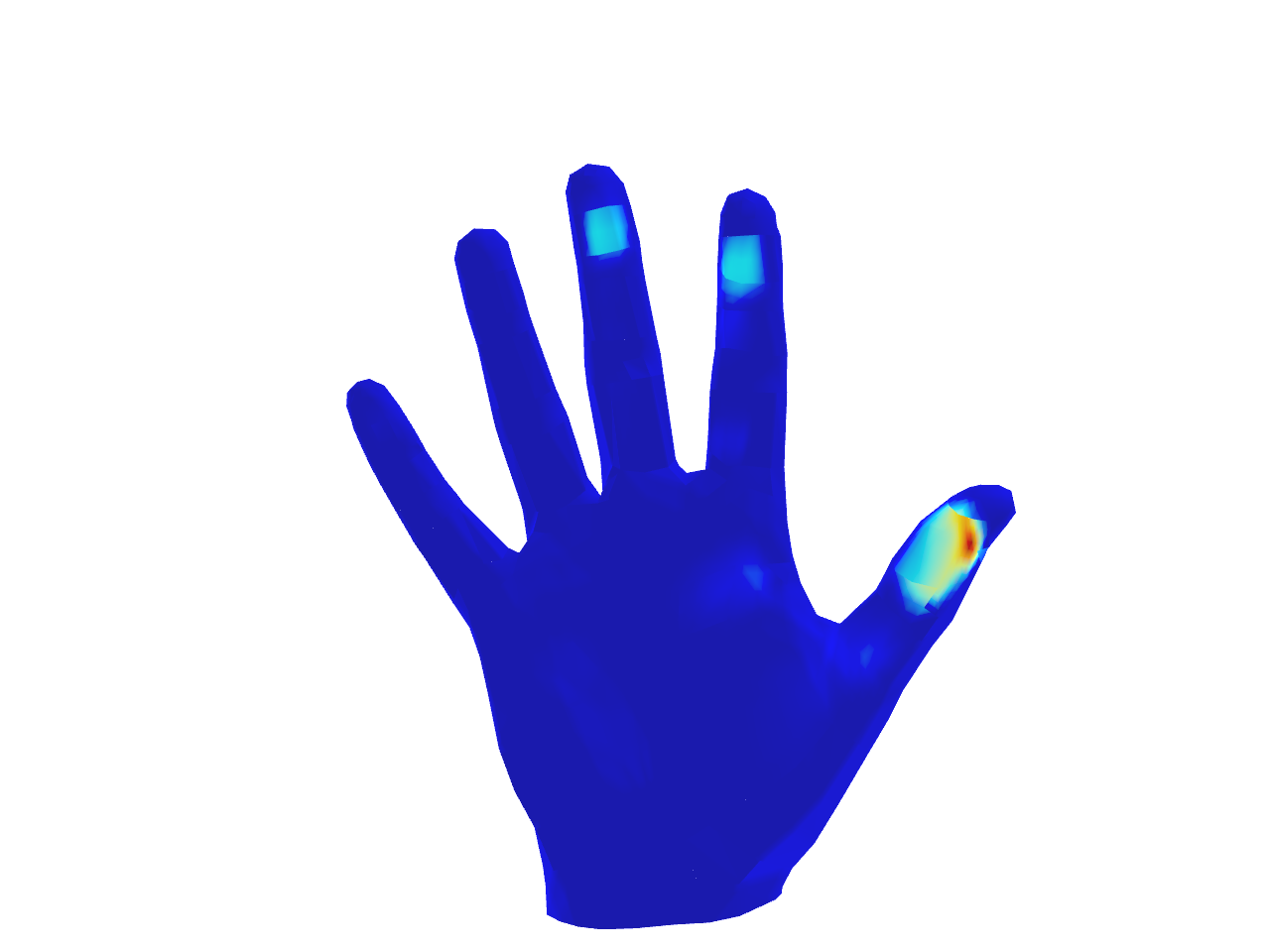} &
            \sqimgcenter{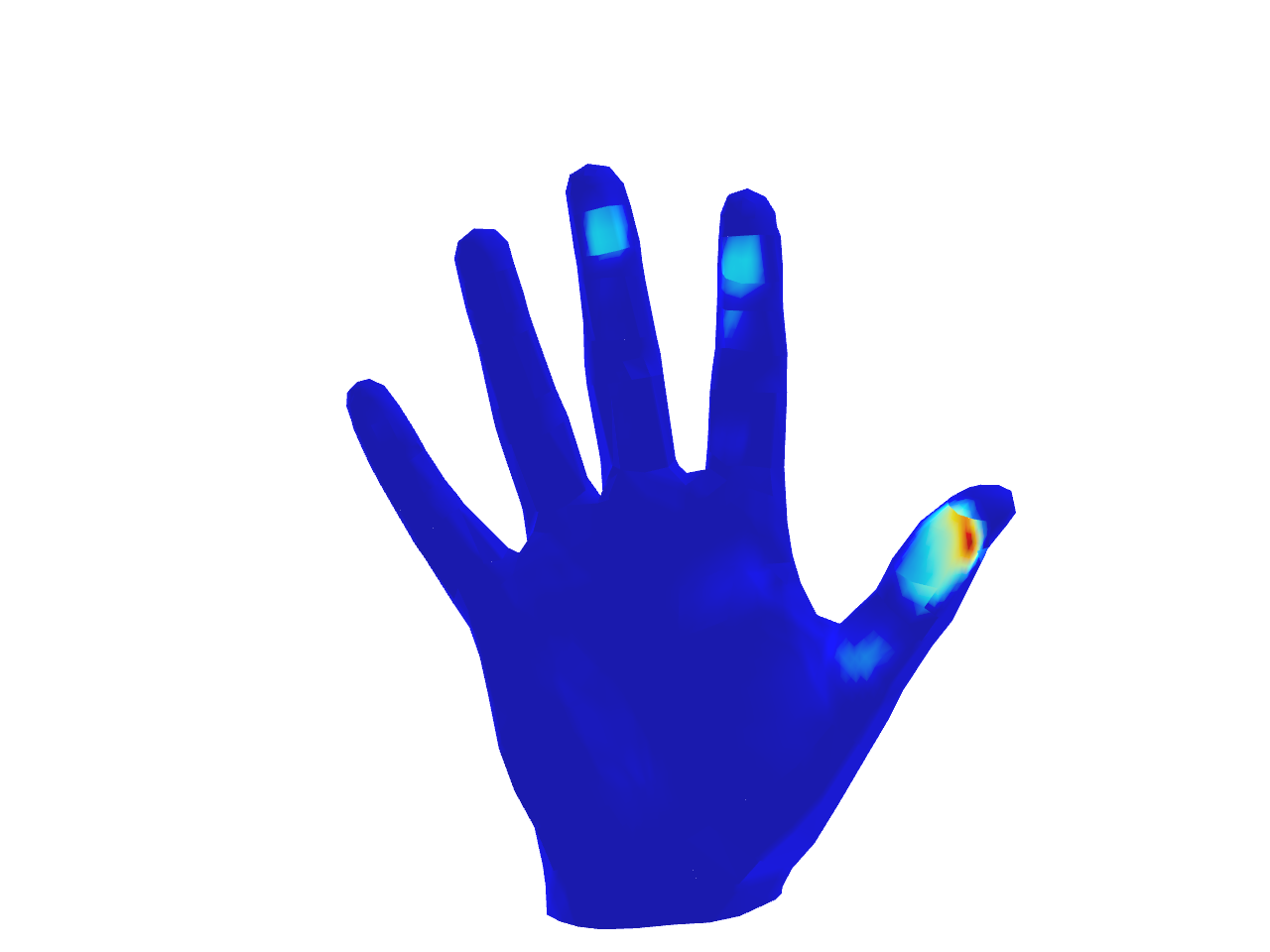} &
            \sqimgcenter{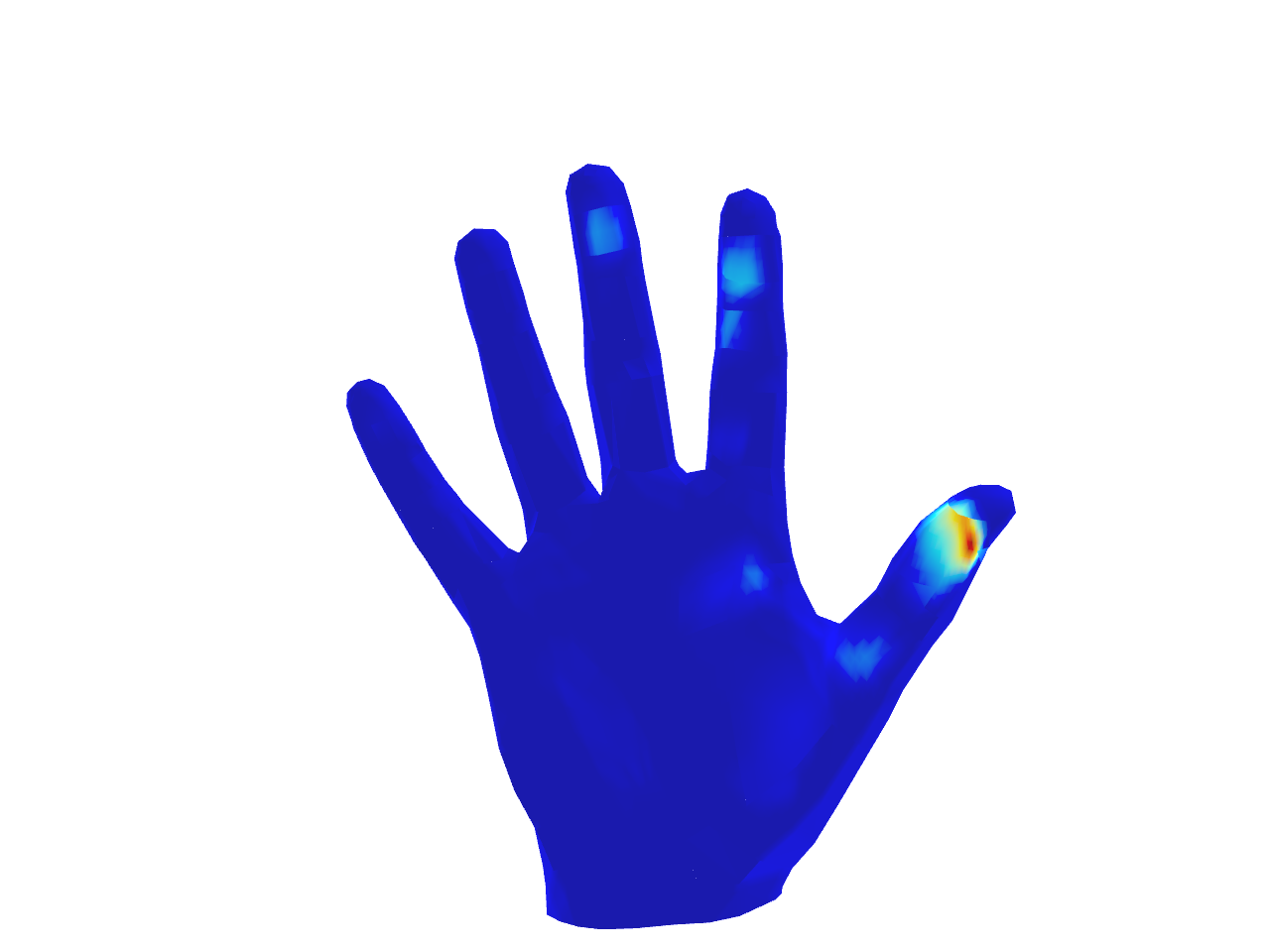} &
            \sqimgcenter{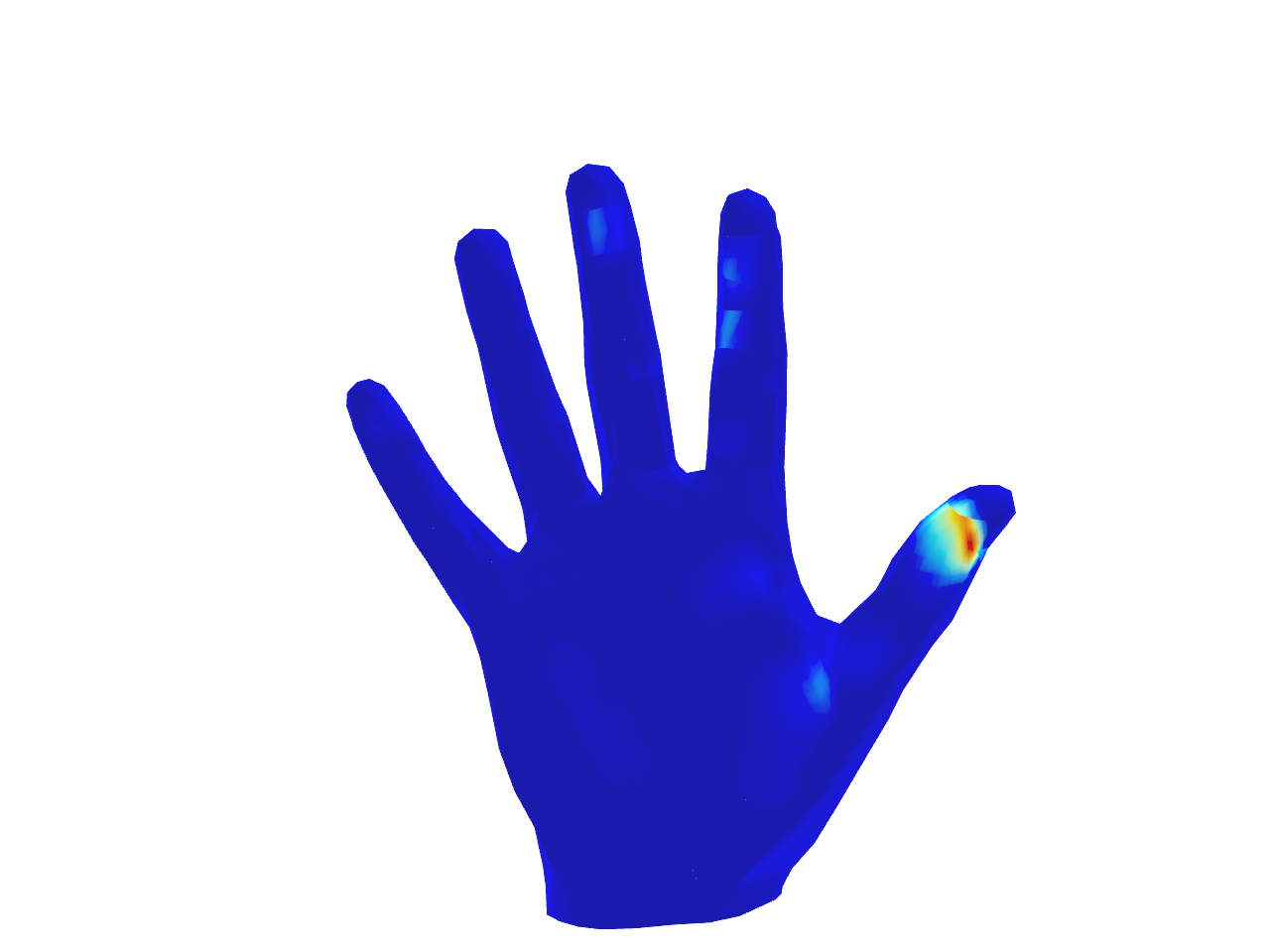} \\

        \rowlabel{gt} &
            \sqimgcenter{result_figs/data_v2t/grocery_target_p4/demo_040/tactile_heatmaps/right_pressure/demo_040_00060.png} &
            \sqimgcenter{result_figs/data_v2t/grocery_target_p4/demo_040/tactile_heatmaps/right_pressure/demo_040_00064.png} &
            \sqimgcenter{result_figs/data_v2t/grocery_target_p4/demo_040/tactile_heatmaps/right_pressure/demo_040_00068.png} &
            \sqimgcenter{result_figs/data_v2t/grocery_target_p4/demo_040/tactile_heatmaps/right_pressure/demo_040_00072.png} &
            \sqimgcenter{result_figs/data_v2t/grocery_target_p4/demo_040/tactile_heatmaps/right_pressure/demo_040_00076.png} &
            \sqimgright{result_figs/data_v2t/grocery_target_p4/demo_040/rgb_frames/demo_040_00060.jpg} &
            \sqimgright{result_figs/data_v2t/grocery_target_p4/demo_040/rgb_frames/demo_040_00064.jpg} &
            \sqimgright{result_figs/data_v2t/grocery_target_p4/demo_040/rgb_frames/demo_040_00068.jpg} &
            \sqimgright{result_figs/data_v2t/grocery_target_p4/demo_040/rgb_frames/demo_040_00072.jpg} &
            \sqimgright{result_figs/data_v2t/grocery_target_p4/demo_040/rgb_frames/demo_040_00076.jpg} \\

        \rowlabel{result} &
            \sqimgcenter{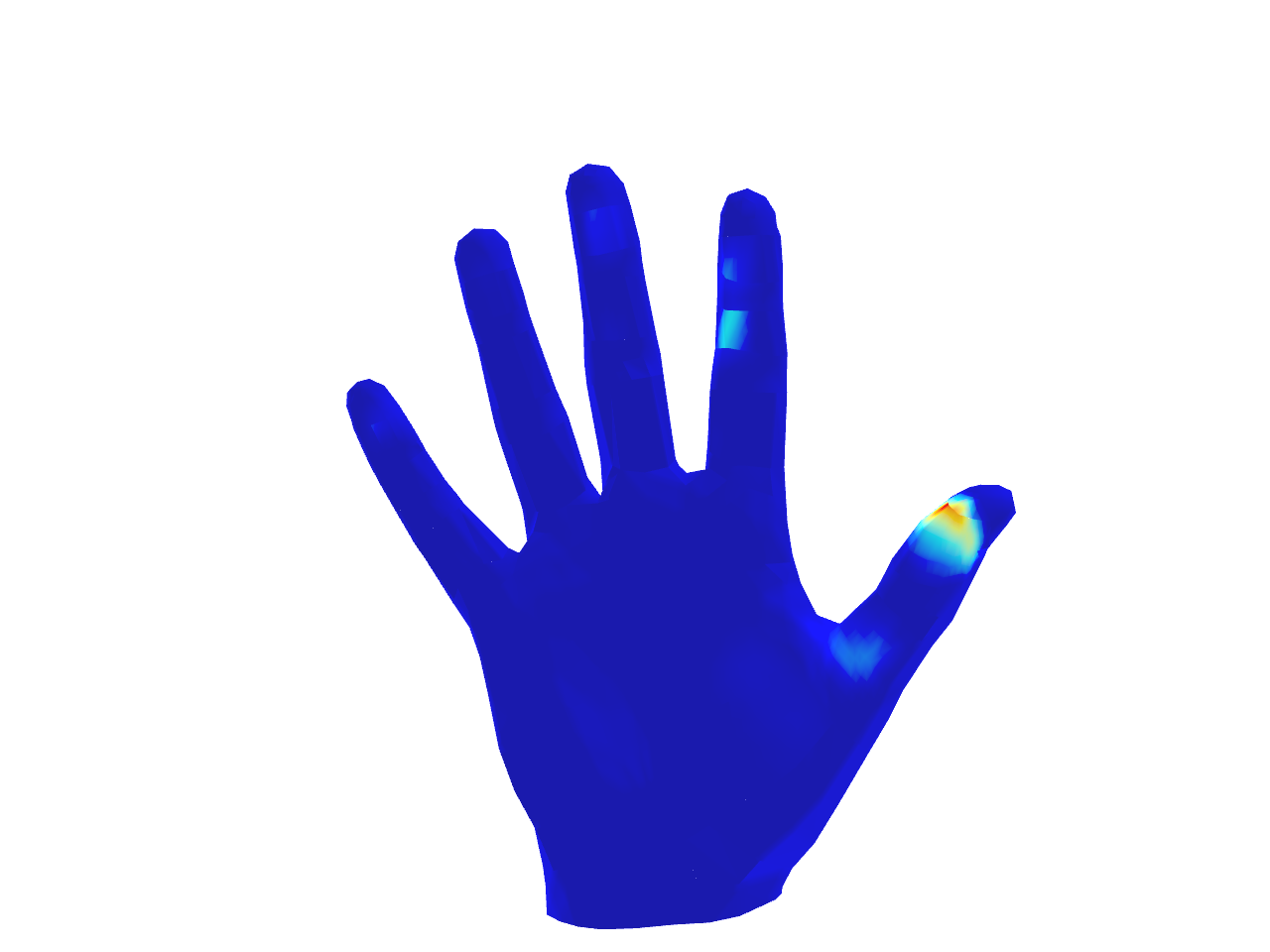} &
            \sqimgcenter{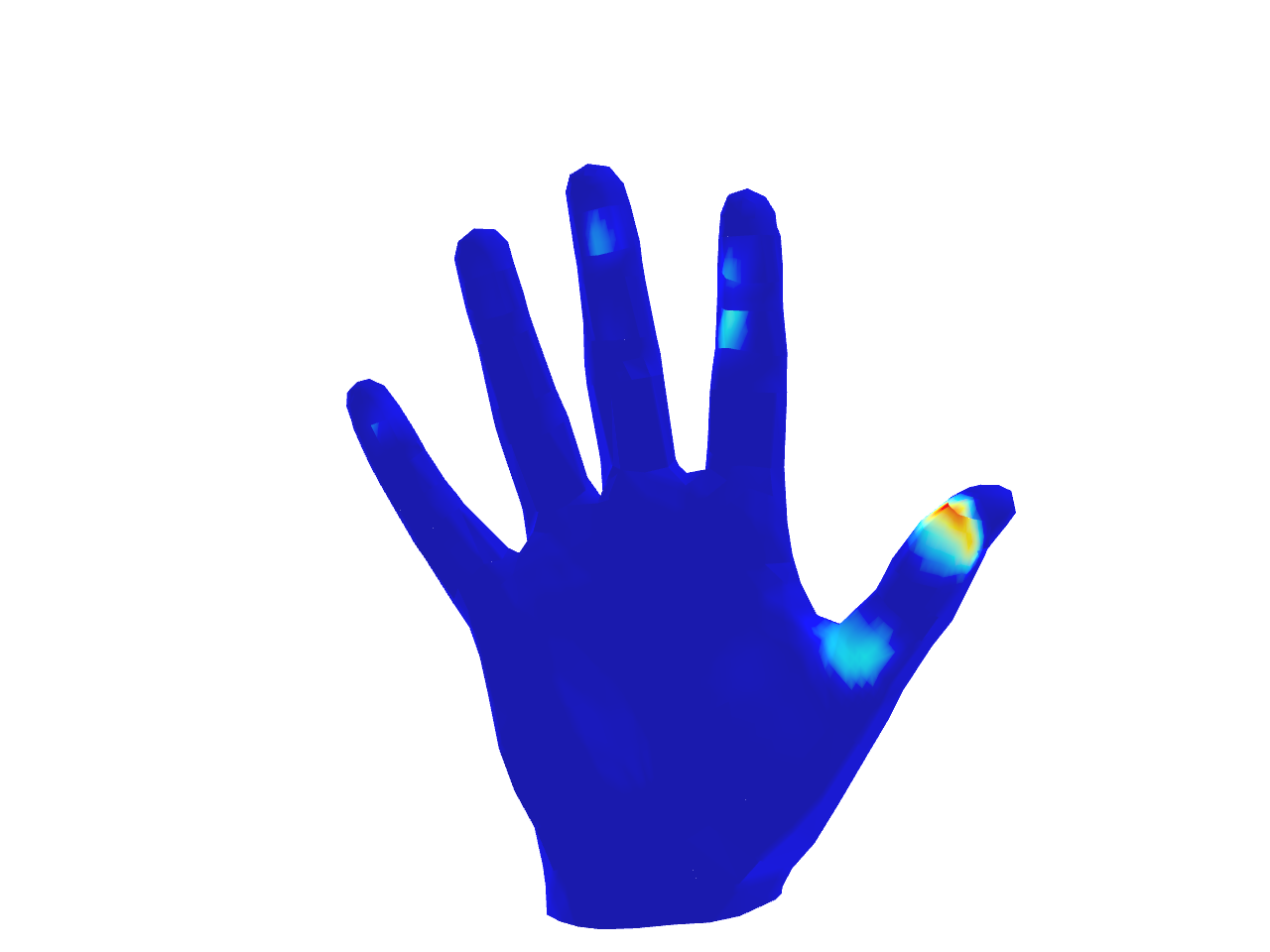} &
            \sqimgcenter{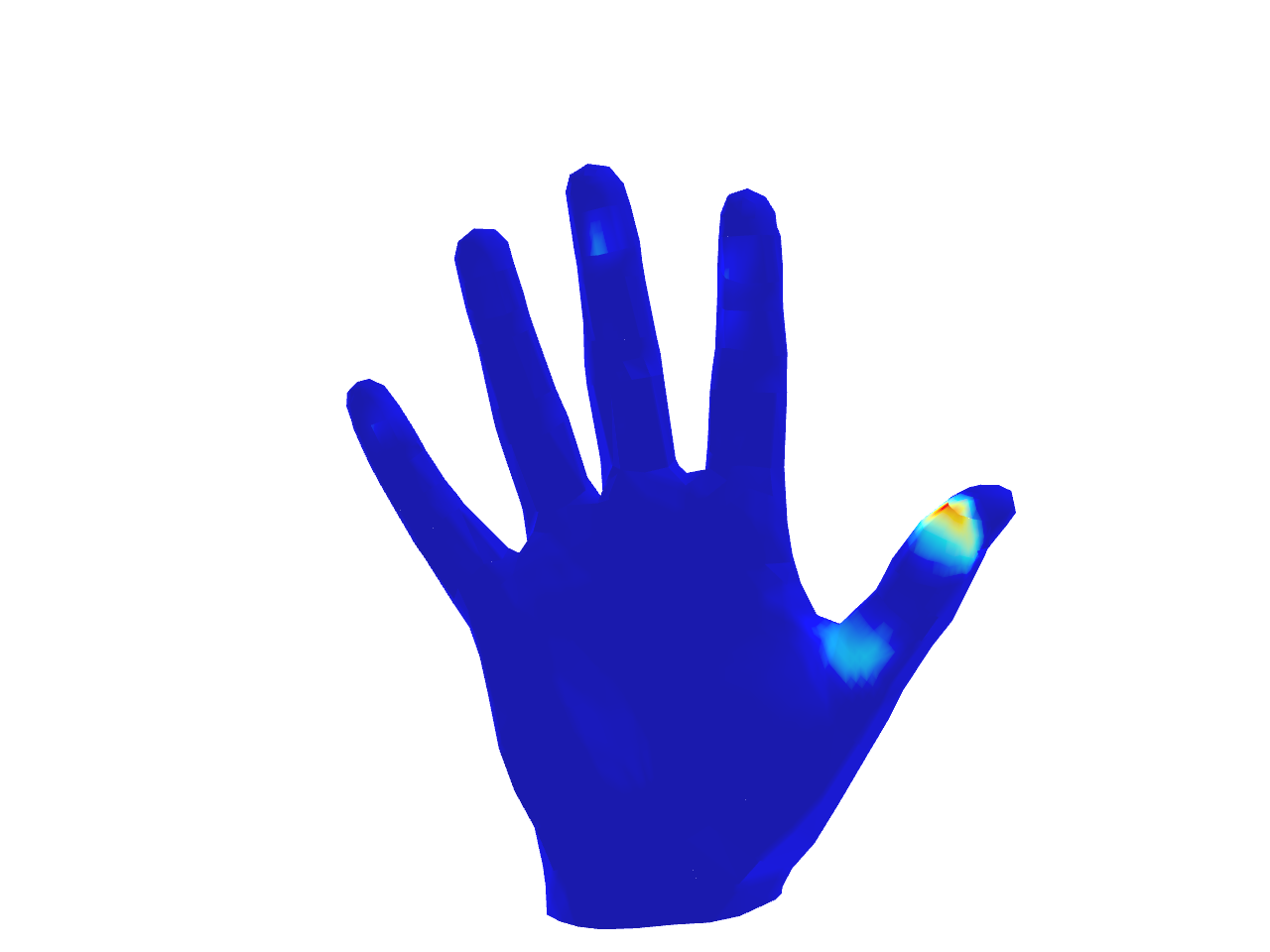} &
            \sqimgcenter{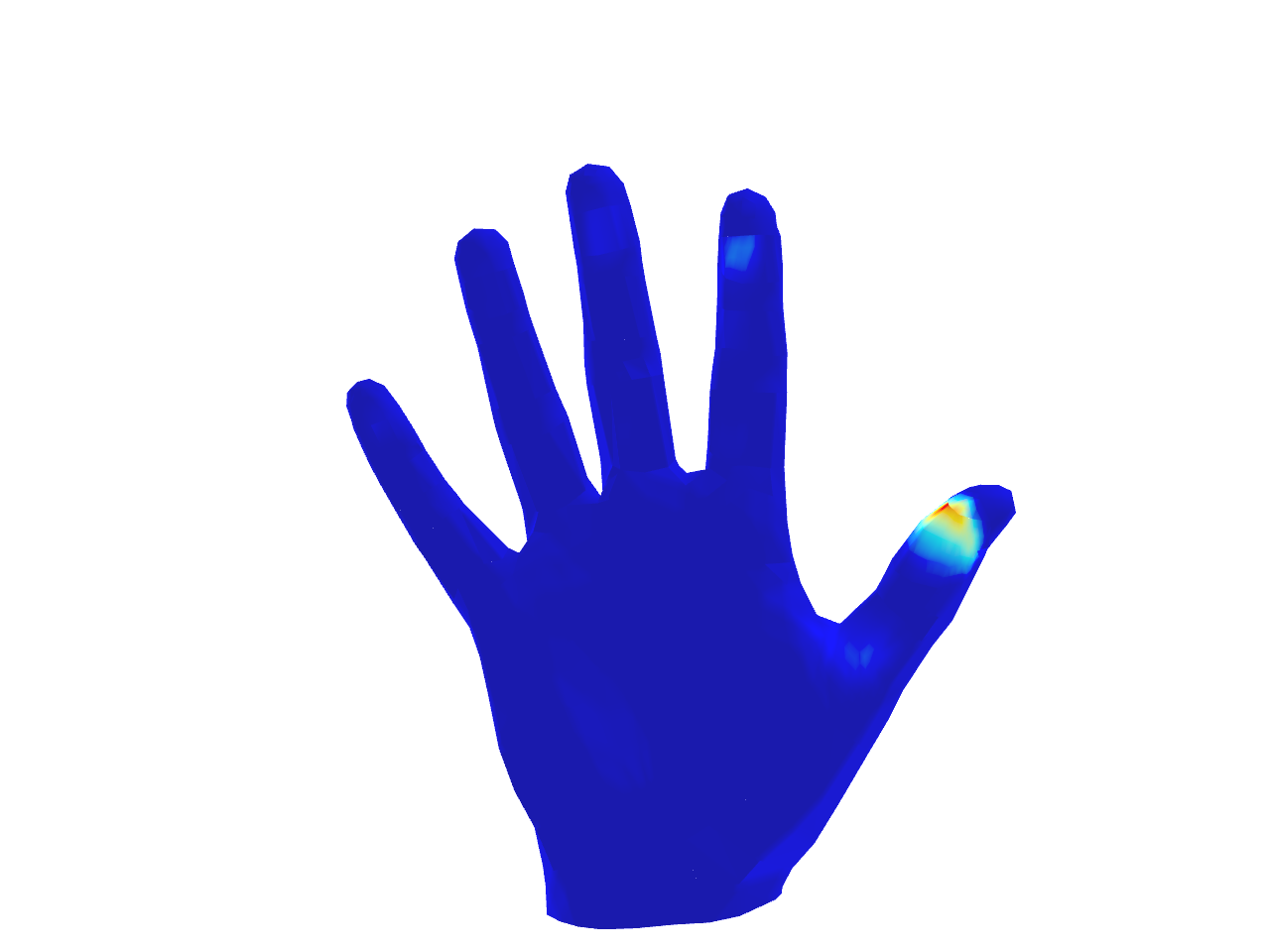} &
            \sqimgcenter{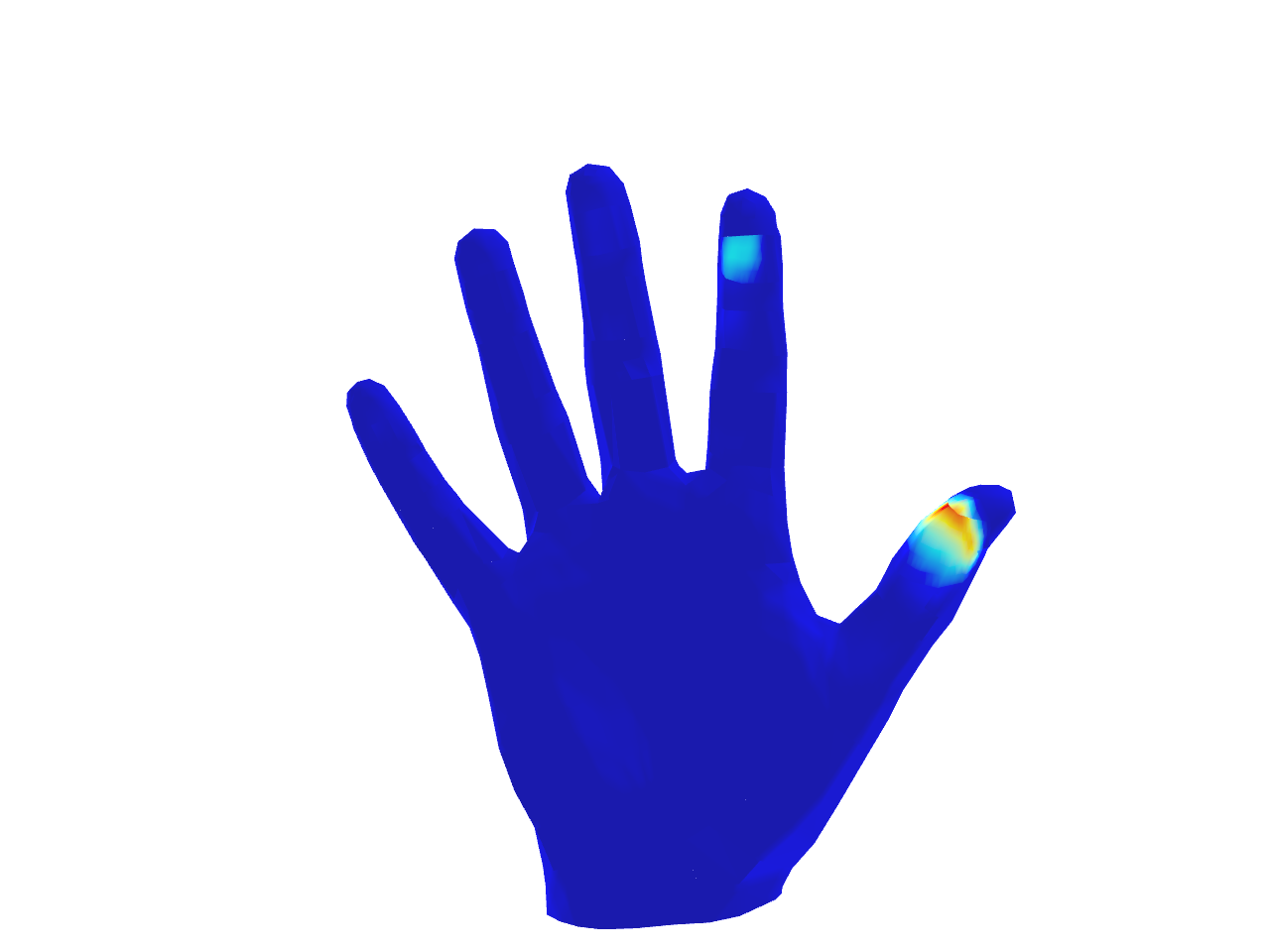} &
            \sqimgright{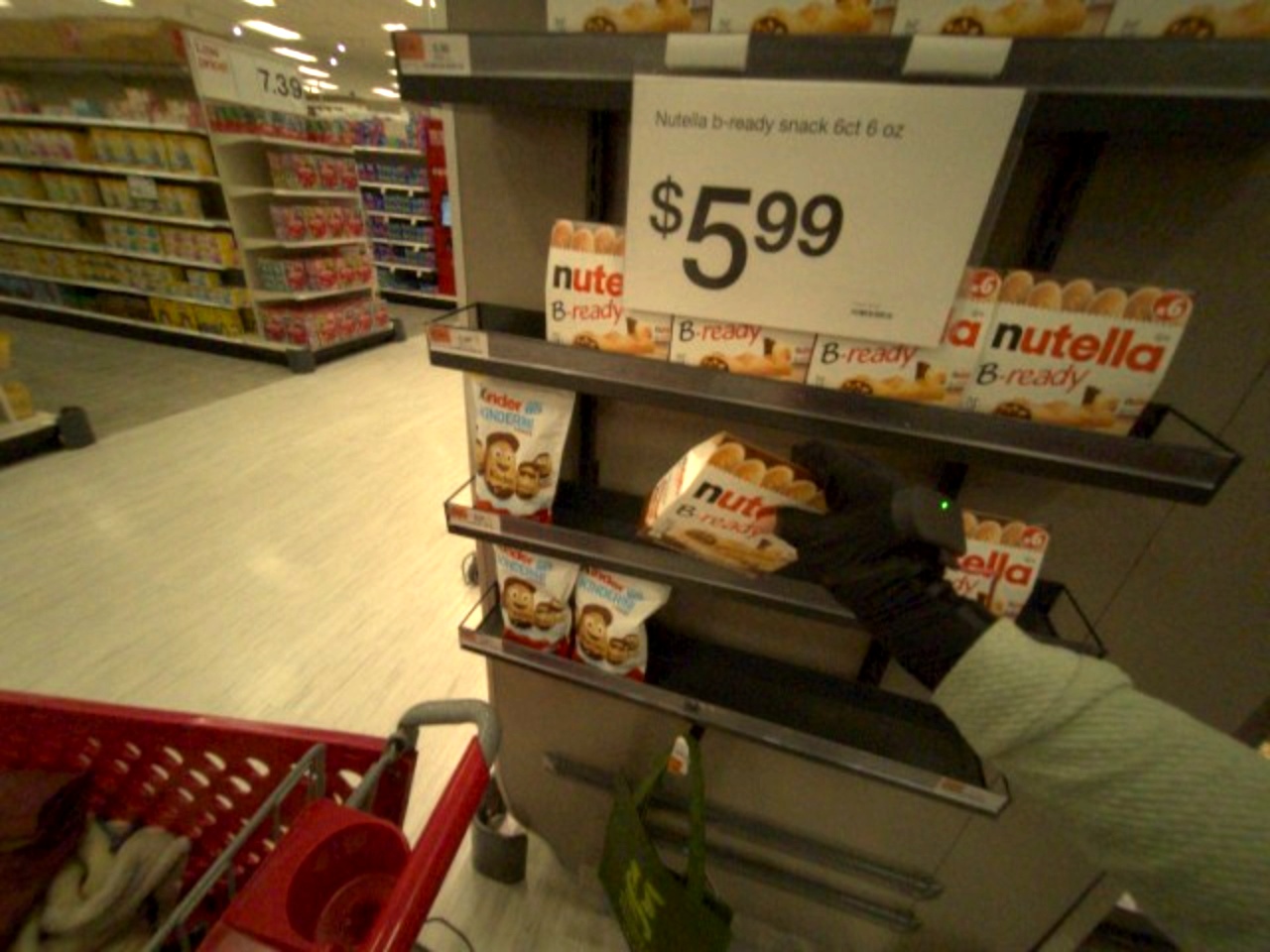} &
            \sqimgright{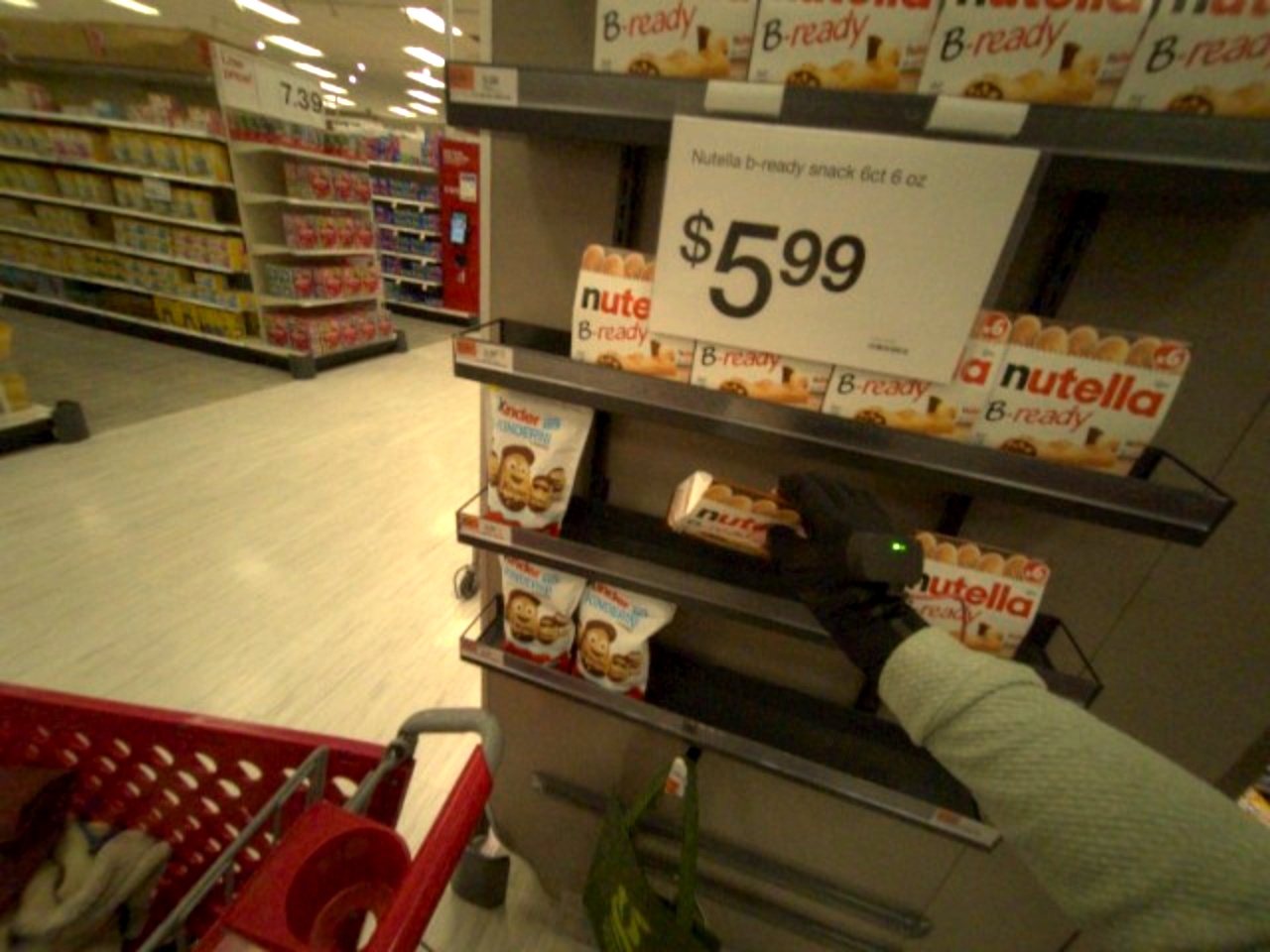} &
            \sqimgright{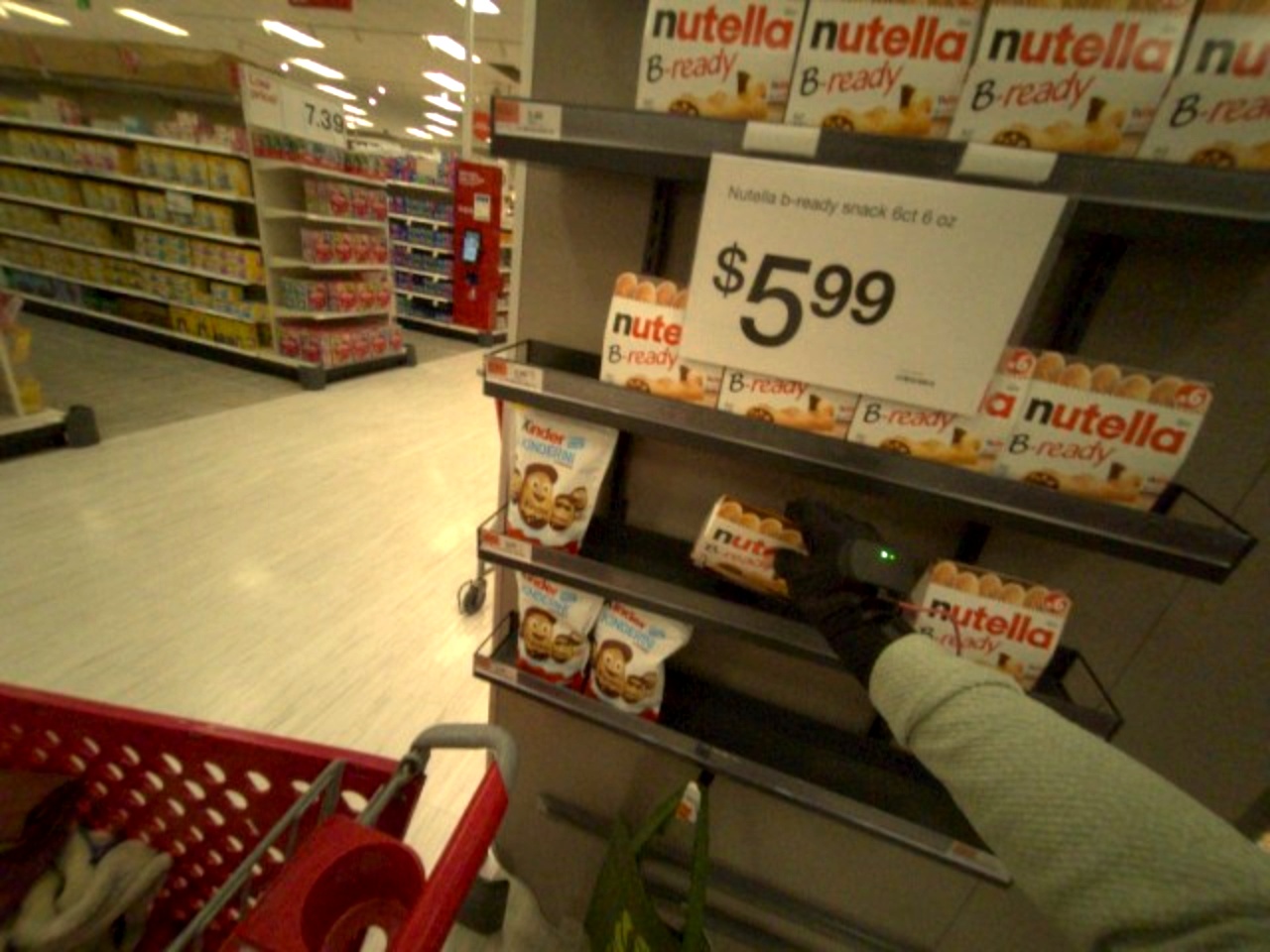} &
            \sqimgright{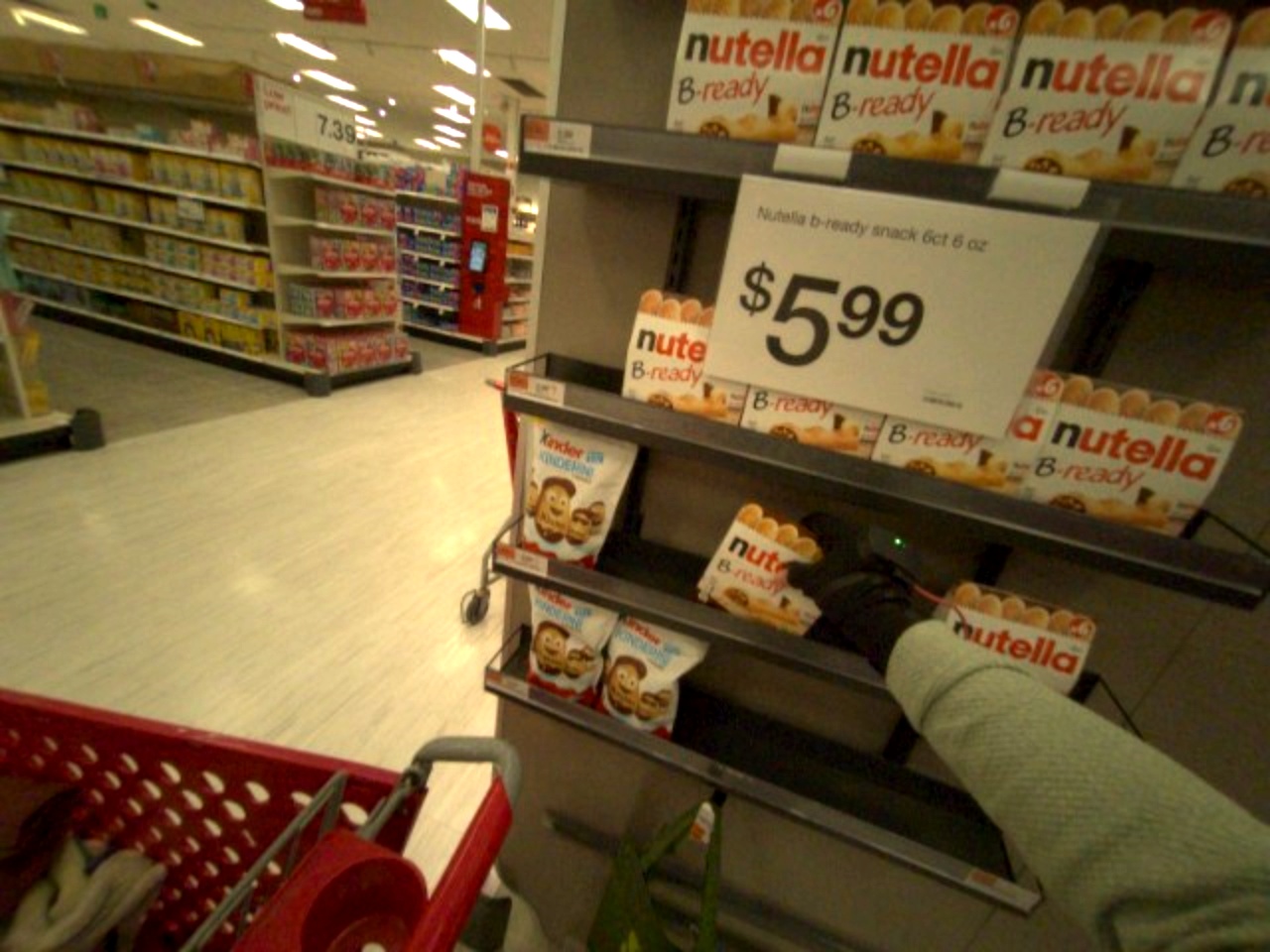} &
            \sqimgright{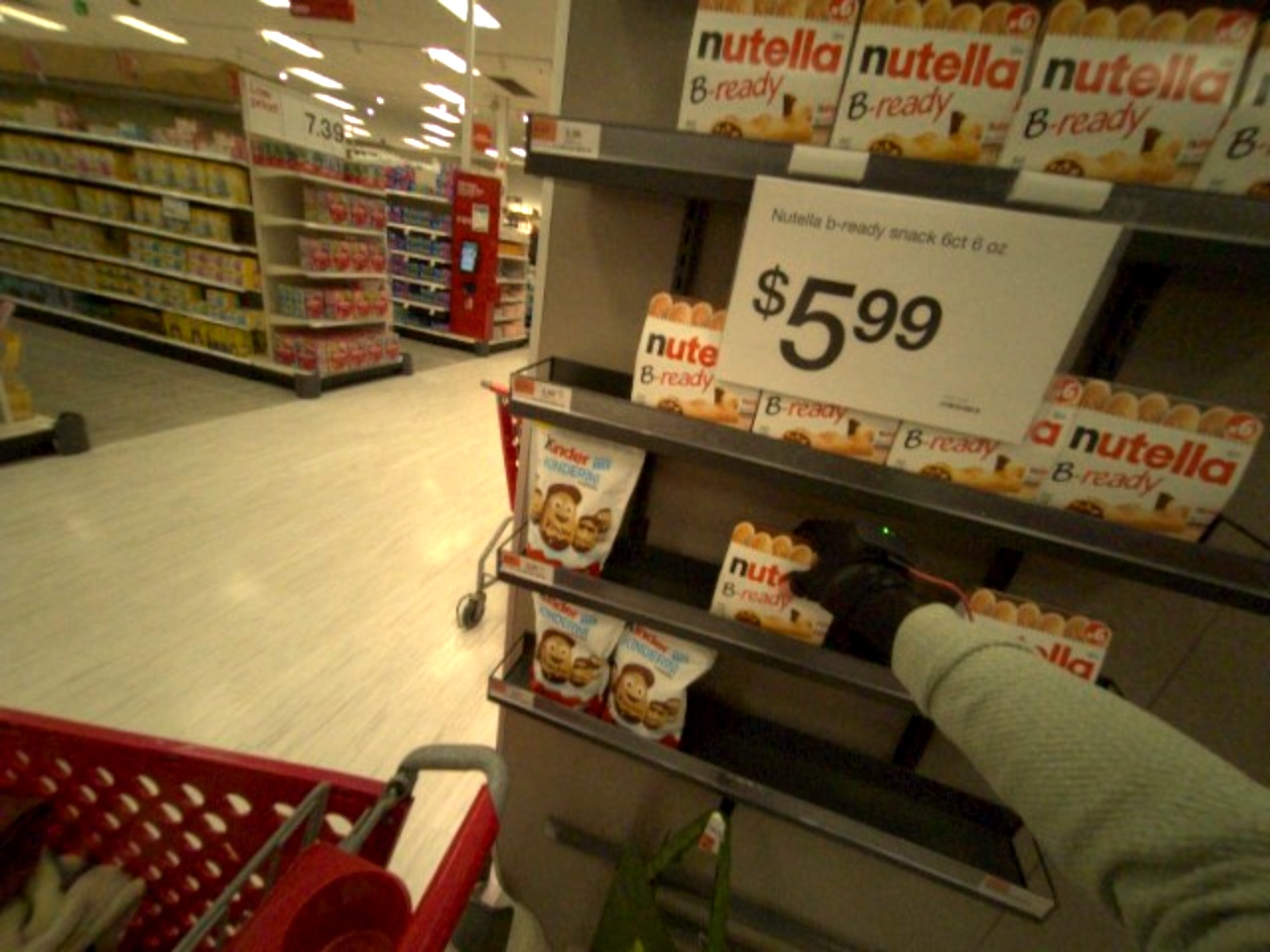} \\

          \\[0.8ex] 
          & \multicolumn{5}{c}{\textsl{(a) Video$\to$Tactile}} &
            \multicolumn{5}{c}{\textsl{(b) Tactile$\to$Video}} \\
    \end{tabular}

    \caption{
    \textbf{Qualitative retrieval results.} Each block shows the query, the ground-truth, and the retrieved target using five frames selected from the video sequence for visualization only. \textbf{Left:} Video-to-tactile retrieval. Top result exhibits a highly similar pressure distribution. \textbf{Right:} Tactile-to-video retrieval. Videos depicting similar interactions are retrieved from tactile input. The top block corresponds to grasping and flipping a round object, while the bottom block shows placing an object down.}
    \label{fig:qual_class}
\end{figure*}

\vspace{1mm}
\noindent\textbf{Metrics and baselines.}
We follow the ObjectFolder protocol~\cite{gao2023objectfolder} and compare linear and deep retrieval methods.
Linear baselines use Canonical Correlation Analysis (CCA)~\cite{10.1145/1873951.1873987} and Partial Least Squares Correlation Analysis (PLSCA)~\cite{jong2001canonical}; our deep baseline is a CLIP-style~\cite{radford2021learning} contrastive framework that aligns visual, tactile, and pose modalities in a shared embedding space, similar to Touch and Go~\cite{yang2022touch} and UniTouch~\cite{yang2024binding}.
We report Recall@1/5/10~\cite{girdhar2023imagebind} and mean Average Precision (mAP)~\cite{gao2023objectfolder, yang2024binding}.

\subsubsection{Tactile Pattern Classification}
\label{tactileclass}

We design two multimodal classification tasks that uses egocentric video, tactile signals, and hand pose to recognize (1) actions and (b) grasp types.
This probes how touch disambiguates actions and grasps, and supports applications such as manipulation recognition, tactile grasp taxonomies, and robot policy learning.

\vspace{1mm}
\noindent\textbf{Task definition.}
We design two independent classification tasks:
(1) hand action recognition and (2) grip type recognition.
Each task is trained and evaluated separately using the same data splits.
The hand action describes the overall motion or intent of the hand–object interaction,
whereas the grip type characterizes the hand’s contact configuration with the object.
For example, the action \textit{pointing} corresponds to an \textit{Index-Finger Extension} grip, while \textit{picking up} is typically associated with a \textit{Prismatic Two-Finger} grip.

\vspace{1mm}
\noindent\textbf{Metrics and baselines.}
We consider two tactile encoding strategies: (1) a lightweight CNN encoder for tactile feature extraction, which is similar to the encoders used for piezoresistive tactile sensors in previous work~\cite{luo2024tactile, luo2021learning}; and (2) a pretrained ResNet18~\cite{he2016deep} encoder applied to tactile images, a common approach used in prior work on tactile image representation~\cite{gao2023objectfolder, yang2022touch}. For both tasks we report standard classification accuracy (Acc.), which measures the percentage of correctly predicted labels in the test set.


\begin{table*}[t]
  \centering
  \caption{\textbf{Bi- and tri-modal cross-sensory retrieval results.} Combining modalities consistently improves retrieval performance, indicating that video, pose, and tactile cues provide complementary information.}
  \label{tab:cross_modal_all}

  \begin{subtable}{\textwidth}
    \centering
    \footnotesize
    \setlength{\tabcolsep}{0.48pt}
    \caption{Bi-modal cross-sensory retrieval.}
    \label{tab:bimodal_combined}

    \begin{tabular}{lcccc|cccc|cccc|cccc|cccc|cccc}
      \toprule
       & \multicolumn{4}{c}{video $\rightarrow$ tactile} 
       & \multicolumn{4}{c}{tactile $\rightarrow$ video}
       & \multicolumn{4}{c}{tactile $\rightarrow$ pose}
       & \multicolumn{4}{c}{pose $\rightarrow$ tactile} 
       & \multicolumn{4}{c}{video $\rightarrow$ pose} 
       & \multicolumn{4}{c}{pose $\rightarrow$ video} \\
      \cmidrule(lr){2-5}
      \cmidrule(lr){6-9}
      \cmidrule(lr){10-13}
      \cmidrule(lr){14-17}
      \cmidrule(lr){18-21}
      \cmidrule(lr){22-25}
      \textbf{Method} 
          & R@1 & R@5 & R@10 & mAP
          & R@1 & R@5 & R@10 & mAP
          & R@1 & R@5 & R@10 & mAP
          & R@1 & R@5 & R@10 & mAP
          & R@1 & R@5 & R@10 & mAP
          & R@1 & R@5 & R@10 & mAP \\
      \midrule
      Chance 
          & 0.07 & 0.07 & 0.07 & 0.07
          & 0.07 & 0.07 & 0.07 & 0.07
          & 0.07 & 0.07 & 0.07 & 0.07
          & 0.07 & 0.07 & 0.07 & 0.07
          & 0.07 & 0.07 & 0.07 & 0.07
          & 0.07 & 0.07 & 0.07 & 0.07 \\
      CCA ~\cite{10.1145/1873951.1873987}
          & 0.50 & 5.36 & 9.15 & 3.59
          & 0.71 & 4.43 & 7.79 & 3.39
          & 0.57 & 2.29 & 4.57 & 2.30
          & 0.64 & 2.64 & 5.08 & 2.43
          & 1.50 & 6.00 & 9.51 & 4.46
          & 1.57 & 5.93 & 10.86 & 4.79 \\
      PLSCA~\cite{jong2001canonical}
          & 0.21 & 2.72  & 5.58 & 2.26
          & 0.64 & 2.57  & 5.08 & 2.49
          & 0.14 & 0.86  & 1.72  & 1.06
          & 0.14 & 0.71  & 1.72  & 0.99 
          & 0.57 & 2.57  & 3.65  & 2.06 
          & 0.57 & 1.93  & 3.79  & 1.98 \\
      Ours
          & 7.15 & 26.73 & 39.74 & 15.47
          & 7.15 & 26.30 & 39.03 & 15.28
          & 7.15 & 21.87 & 30.88 & 13.43
          & 6.93 & 21.02 & 30.45 & 13.13 
          & 9.44 & 32.24 & 42.17 & 19.01
          & 11.72& 31.24 & 42.32 & 19.98 \\
      \bottomrule
    \end{tabular}
    \label{res:bi-modal}
  \end{subtable}

  \vspace{2mm}

  \begin{subtable}{\textwidth}
    \centering
    \small
    \setlength{\tabcolsep}{8pt}
    \caption{Tri-modal cross-sensory retrieval.}
    \label{tab:trimodal_combined}

    \begin{tabular}{lcccc|cccc|cccc}
      \toprule
       & \multicolumn{4}{c}{video + pose $\rightarrow$ tactile}
       & \multicolumn{4}{c}{tactile + pose $\rightarrow$ video}
       & \multicolumn{4}{c}{video + tactile $\rightarrow$ pose} \\
      \cmidrule(lr){2-5}
      \cmidrule(lr){6-9}
      \cmidrule(lr){10-13}
      \textbf{Method}
          & R@1 & R@5 & R@10 & mAP
          & R@1 & R@5 & R@10 & mAP
          & R@1 & R@5 & R@10 & mAP \\
      \midrule
      Chance
          & 0.07 & 0.07 & 0.07 & 0.07
          & 0.07 & 0.07 & 0.07 & 0.07
          & 0.07 & 0.07 & 0.07 & 0.07 \\
      CCA~\cite{10.1145/1873951.1873987}
          & 1.36 & 6.86 & 11.65 & 5.06
          & 2.29 & 8.51 & 13.51 & 6.45
          & 2.14 & 7.93 & 13.22 & 5.91 \\
      PLSCA~\cite{jong2001canonical}
          & 0.21 & 2.22  & 4.57 & 2.18
          & 0.50 & 4.43  & 8.08 & 3.29
          & 0.36 & 1.43  & 4.22  & 1.95 \\
      Ours
          & 14.08 & 42.96 & 62.26 & 26.86
          & 12.72 & 38.53 & 53.18 & 23.46
          & 15.44 & 43.39 & 57.61 & 26.86 \\
      \bottomrule
    \end{tabular}
    \label{res:tri-modal}
  \end{subtable}

\end{table*}

\vspace{-1em}
\section{Experiments}
\label{sec:experiments}
We evaluate baseline performance on:
\textbf{(1) cross-sensory retrieval}, and
\textbf{(2) tactile pattern classification}.
We report results for settings involving tactile in the main paper; ablations without tactile and additional applications are provided in the Supp. Mat.

\vspace{-0.5mm}
\subsection{Cross-Sensory Retrival Perfomance}
We evaluate retrieval across all modality pairs and observe consistent, large improvements over linear baselines (CCA, PLSCA),as  shown in Tab.~\ref{res:bi-modal} and Tab.~\ref{res:tri-modal}. In addition to the quantitative results, Fig.~\ref{fig:qual_class} provides qualitative retrieval examples that further illustrate the model's ability to recover semantically and physically consistent cross-modal matches, capturing similar contact patterns, dynamics and rasping behaviors across modalities. Below, we summarize the key trends for each group of tasks and highlight the insights that emerge from these results.
 
\vspace{1mm}
\noindent\textbf{Video $\leftrightarrow$ Tactile Retrieval.}
Across both retrieval directions, our contrastive baseline consistently outperforms others by a large margin.
In the video$\rightarrow$tactile setting, linear baselines achieve R@1 of 0.50\% and 0.21\%, respectively, whereas our contrastive model reaches 7.15\%.
A similar trend is observed in the tactile$\rightarrow$video direction.
Despite the substantial sensory gap between vision and touch, the learned embedding aligns visual cues with tactile signals. 
The symmetry across both directions further suggests that learned representation is genuinely multimodal rather than biased toward a single modality.

\vspace{1mm}
\noindent\textbf{Tactile $\leftrightarrow$ Pose Retrieval.}
Retrieval between tactile and pose exhibits substantial improvements over linear baselines. In the tactile$\rightarrow$pose setting, linear baselines achieve at most 0.57\% R@1, whereas our contrastive method reaches 7.15\%, with mAP improved to 13.43\%.
The inverse direction (pose$\rightarrow$tactile) shows a similarly substantial gains. These trends highlight that touch and pose share a strong geometric coupling which our model successfully captures, while linear baselines fails.

\vspace{1mm}
\noindent\textbf{Multimodal $\rightarrow$ Unimodal Retrieval.}
When retrieving tactile or pose from two input modalities, performance improves markedly.
In the video$+$pose$\rightarrow$tactile setting, our model achieves an mAP of 26.86\%, significantly higher than either modality alone, indicating that video provides global scene and object context while pose contributes fine-grained hand-motion structure, and their combination reduces ambiguity in tactile prediction.
A similar improvement is observed for tactile$+$pose$\rightarrow$video retrieval, where our method achieves an mAP of 23.46\%, substantially outperforming unimodal retrieval.
The video$+$tactile$\rightarrow$pose setting achieves an mAP of 26.86\%, suggesting that the combination of visual and tactile cues provides an effective supervisory signal for reconstructing hand articulation.
Across all fusion settings, multimodal inputs offer complementary information that reduces retrieval ambiguity, and the learned embedding effectively exploits shared cross-modal structure.

\begin{table}[t]
\centering
\caption{Action and grasp accuracy (\%). Tactile is most informative for grasp; tactile+vision gives best overall.}
\label{tab:classification_results}
\setlength{\tabcolsep}{6pt}

\begin{tabular}{lcccc}
\toprule
& \multicolumn{2}{c}{Action Acc.} & \multicolumn{2}{c}{Grasp Acc.} \\
\cmidrule(lr){2-3}\cmidrule(lr){4-5}
Modality & {RN18} & {Lite-CNN} & {RN18} & {Lite-CNN} \\
\midrule
V   & \multicolumn{2}{c}{\bfseries{40.26}} & \multicolumn{2}{c}{57.45} \\
P   & \multicolumn{2}{c}{33.22} & \multicolumn{2}{c}{46.32} \\
\addlinespace
T     & 29.95 & 31.59  & 60.23 & 57.12 \\
T+P   & 28.31 & 27.00  & 60.72 & 62.19 \\
T+V   & 30.11 & 32.73  & 51.72 & 65.47 \\
T+P+V &  35.02 & 37.32 & 55.65 & \bfseries 68.09 \\
\bottomrule
\end{tabular}

\vspace{1pt}
\footnotesize V=vision, P=proprioception, T=tactile. RN18=ResNet-18.
\vspace{-2em}
\end{table}

\subsection{Classification Performance.}
We evaluate our representation on hand action and grasp type prediction (Table~\ref{tab:classification_results}).
Using video alone performs reasonably well for both tasks (40.26\%, 57.45\%), highlighting the role of global visual context in manipulation. 
Tactile only is highly informative for grasp type (60.23\%, 57.12\%) but weaker for action recognition (29.95\%, 31.59\%), reflecting that grasp relies on local contact geometry while actions depend on higher-level global context. Adding pose brings only marginal gains, whereas fusing tactile with vision, especially with the lightweight CNN encoder, provides a more complete view of contact and object geometry and leads to consistent performance gains. 
Overall, integrating vision, pose, and tactile cues yields the best performance for grasp type classification, reflecting their complementary roles in modeling hand-object interactions. 


\subsection{Ablation Studies}
We ablate temporal context, tactile encoder design, and discretization to study their impact on multimodal retrieval and classification; full results are reported in the Supp. Mat.

\vspace{1mm}
\noindent
\textbf{Temporal window size.}
We vary the sliding-window length for tactile and visual inputs and observe a consistent trend: retrieval performance improves as the window grows from 5 to 20 frames.
Short windows severely hurt video$\rightarrow$tactile and tactile$\rightarrow$video retrieval because they miss the temporal evolution of contact, while a 20-frame window yields the best results (e.g., video$\rightarrow$tactile Recall@1 from 5.77\% to 8.49\%, video+pose$\rightarrow$tactile from 6.57\% to 12.82\%).
The gains are larger for pose-related retrieval, where longer windows nearly double Recall@1, underscoring the importance of modeling longer temporal dynamics for cross-modal alignment.

\vspace{1mm}
\noindent
\textbf{Tactile encoder capacity.}
We compare our lightweight 16×16 tactile encoder to a ResNet-18 applied to 224×224 pressure maps and find that the deeper model performs substantially worse across all retrieval directions (e.g., tactile$\rightarrow$pose mAP drops from 16.76\% to 6.53\%).
This indicates that tactile signals, unlike natural images, are sparse and highly structured, and are better served by compact encoders that preserve local contact topology rather than high-resolution texture; over-parameterization introduces noise and harms cross-modal alignment.

\vspace{1mm}
\noindent
\textbf{Tactile discretization strategies.}
We evaluate logarithmic and linear discretization with 3, 5, and 7 levels, as well as raw continuous inputs.
Performance generally peaks with mid-range discretization (5–7 levels), although the exact best setting varies by modality pair (e.g., video$\rightarrow$tactile mAP is highest with raw inputs, while tactile$\rightarrow$pose is best with linear 5-level discretization).
Overall, discretization acts as a useful regularizer, reducing sensor noise without severely degrading performance, while continuous tactile maps remain a strong default choice.

\begin{figure}[t]
    \centering

    \setlength{\tabcolsep}{0pt}
    \renewcommand{\arraystretch}{0.0}

    \newcommand{\rowlabel}[1]{\rotatebox[origin=c]{90}{#1}}

    \newcommand{\rgbitwim}[1]{%
        \includegraphics[height=1.6cm,trim=0 0 0 0,clip]{#1}%
    }
    \newcommand{\rgbim}[1]{%
        \includegraphics[height=1.6cm,trim=320 0 0 0,clip]{#1}%
    }
    \newcommand{\tacim}[1]{%
        \includegraphics[height=1.6cm,trim=240 0 240 160,clip]{#1}%
    }

    \begin{tabular}{
        >{\centering\arraybackslash}m{0.45cm}  
        *{5}{>{\centering\arraybackslash}m{1.6cm}} 
    }
        & \rule{0pt}{1.8ex}frame \#1 & frame \#2 & frame \#3 & frame \#4 & frame \#5 \\[0.8ex]

        \rowlabel{query} &
            \rgbitwim{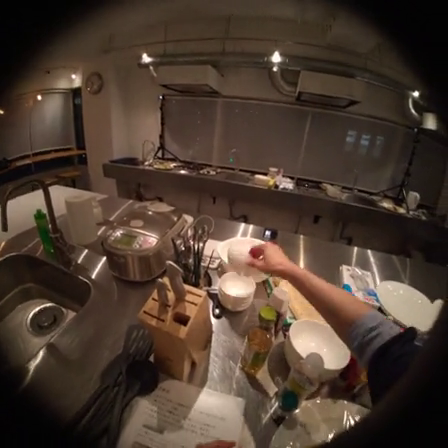} &\rgbitwim{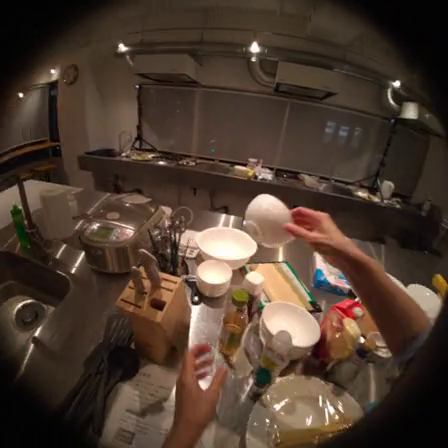} &\rgbitwim{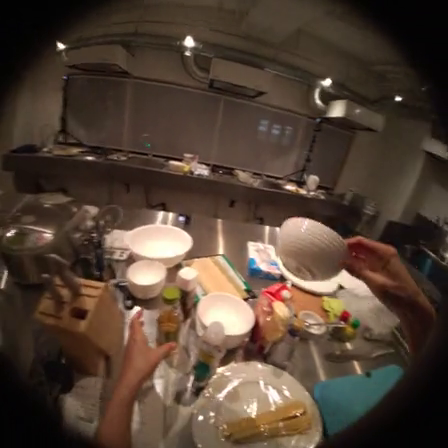} &\rgbitwim{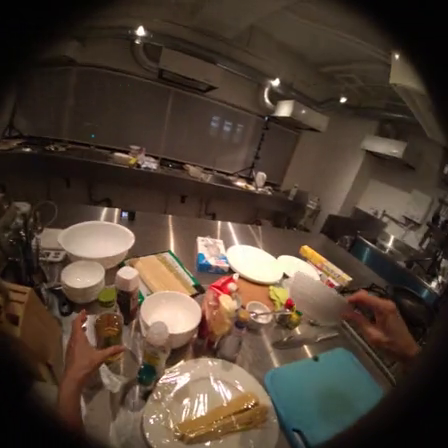} &\rgbitwim{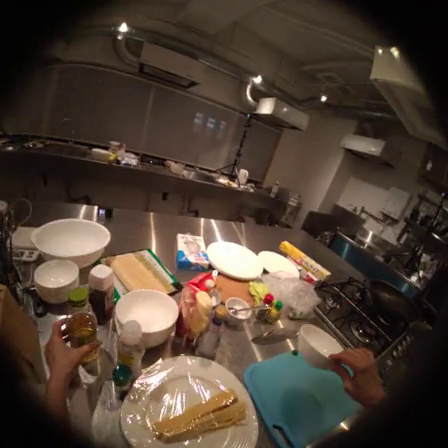} \\

        \rowlabel{t@1} &
            \tacim{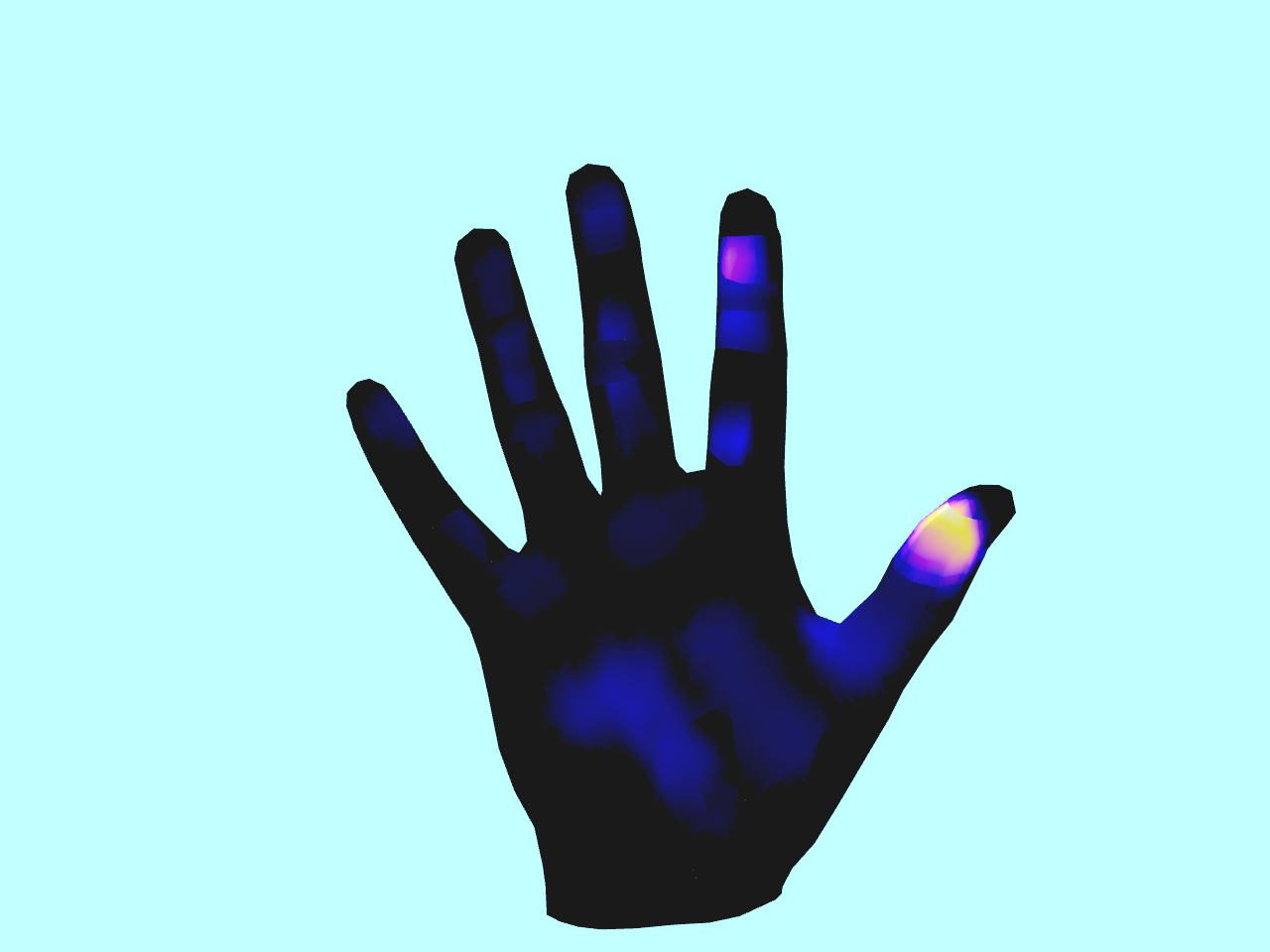} &\tacim{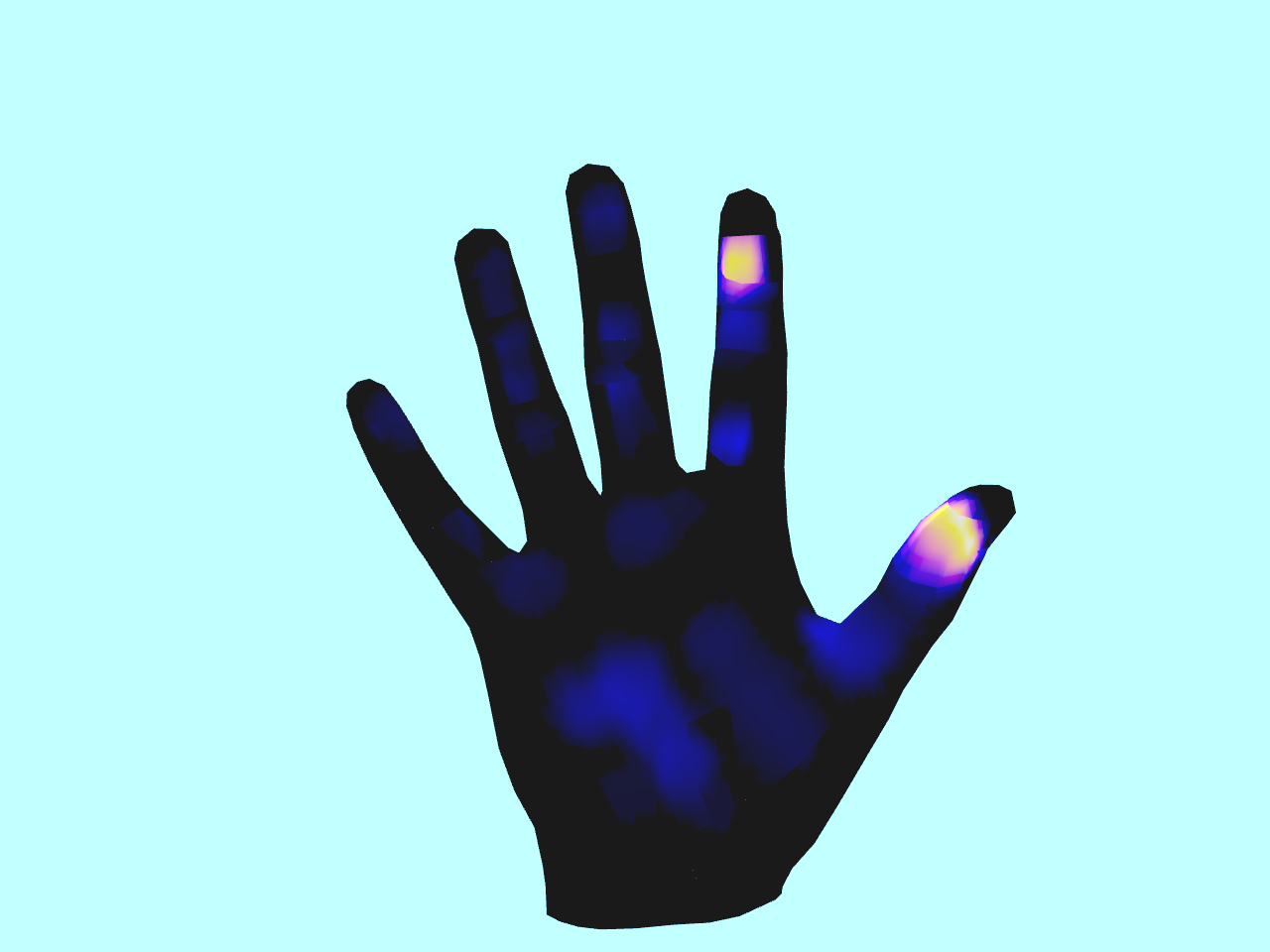} &\tacim{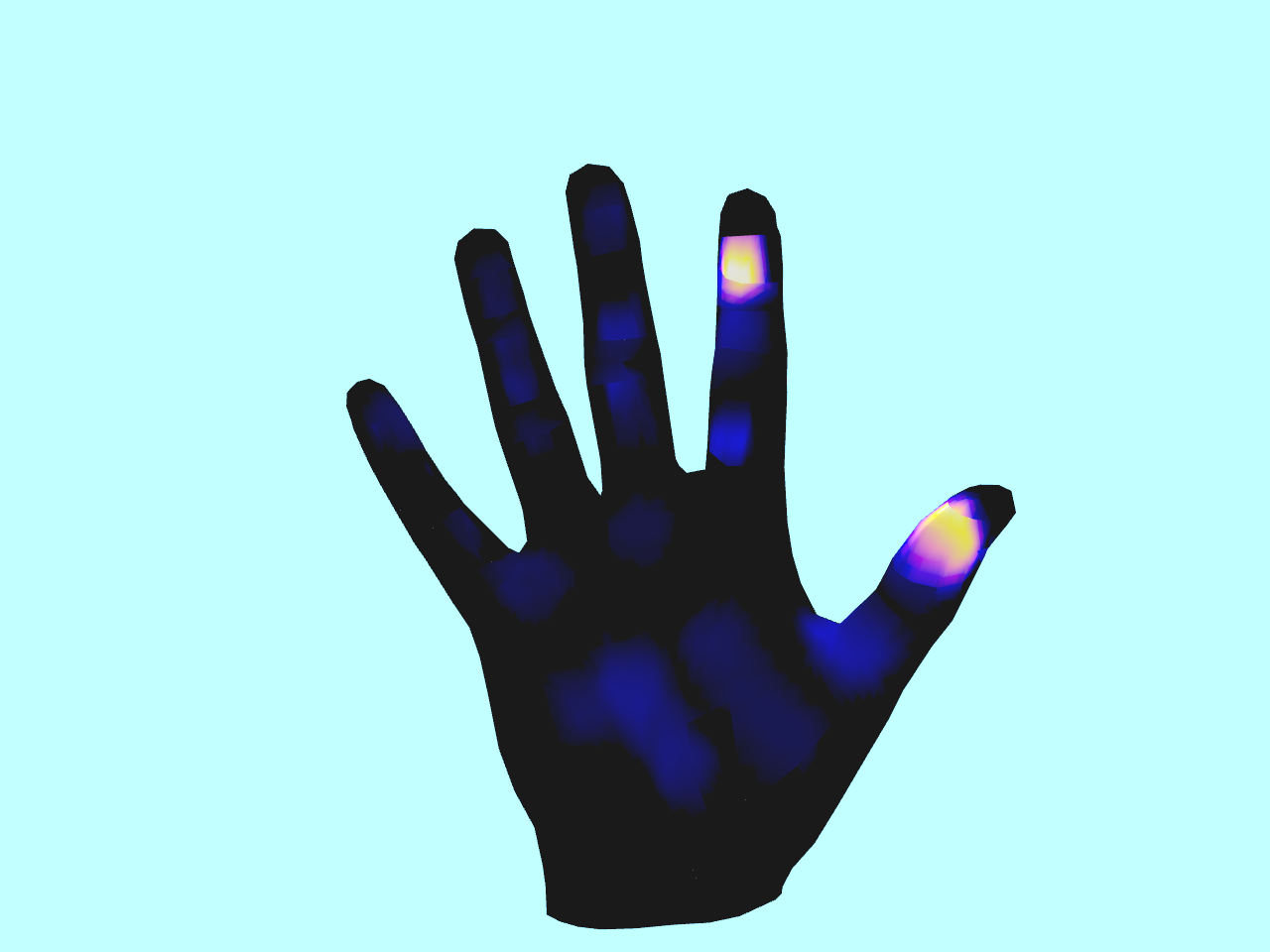} &\tacim{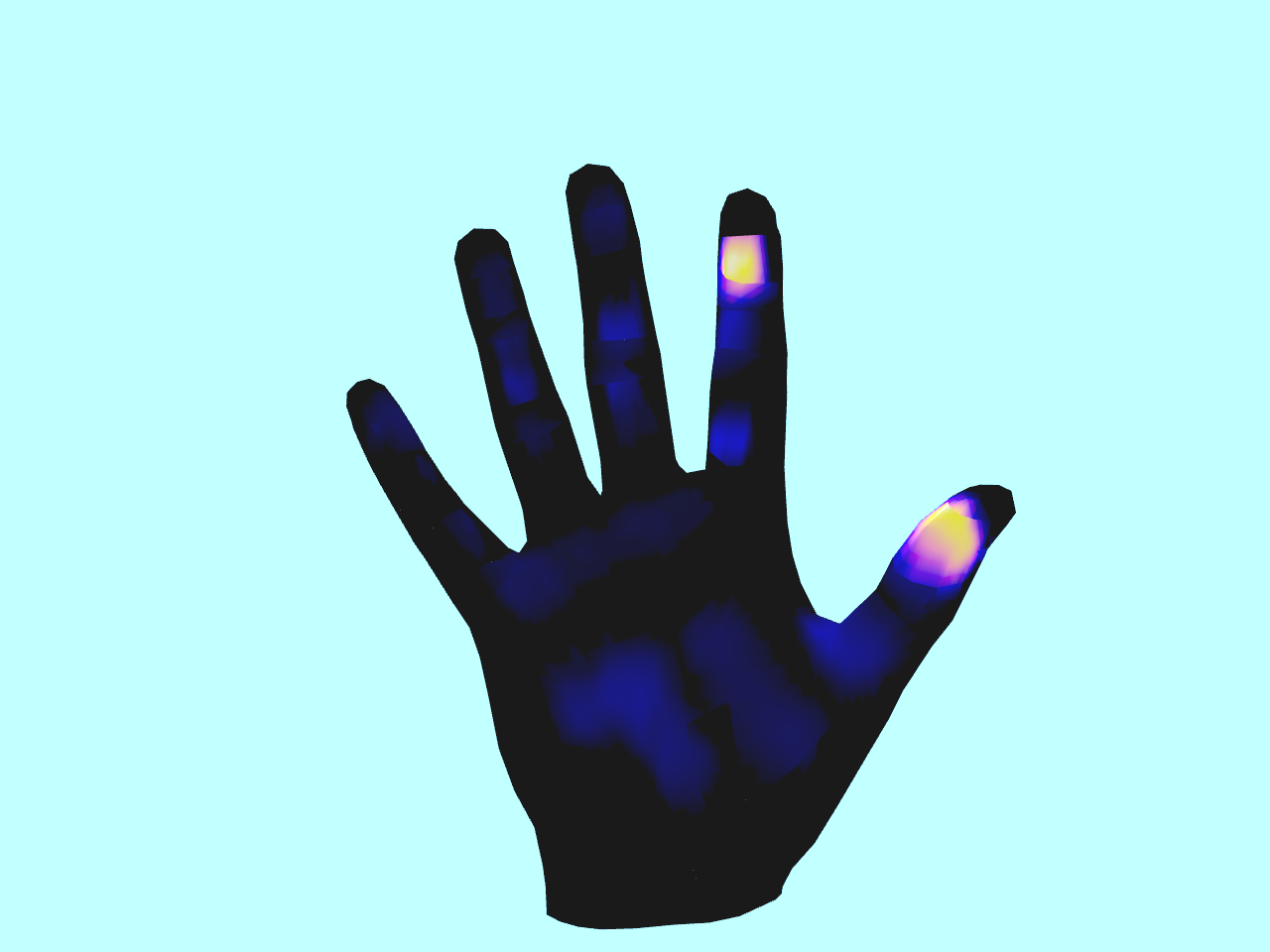} &\tacim{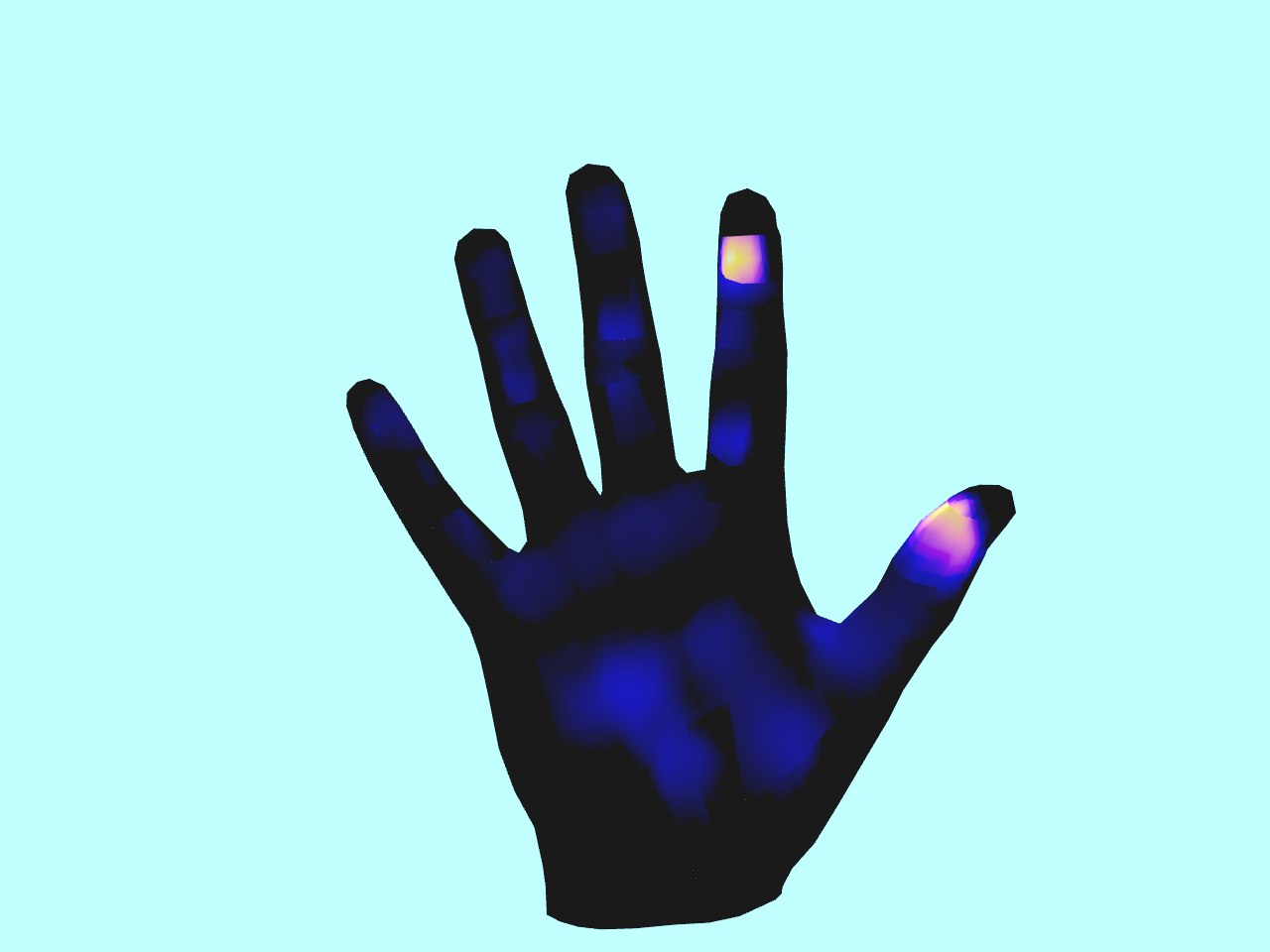} \\

        \rowlabel{v@1} &
            \rgbim{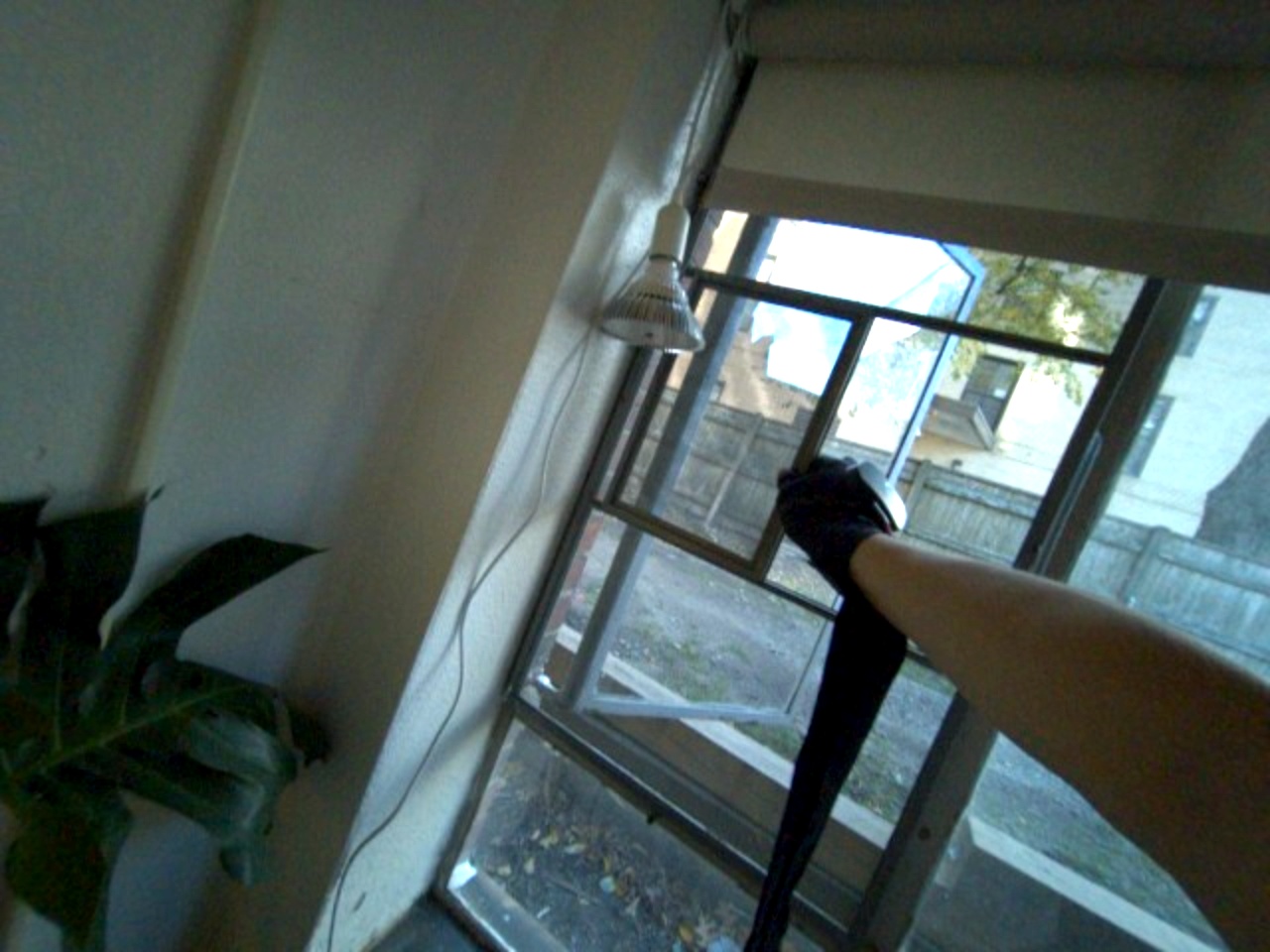} &\rgbim{result_figs/data_itw/rgb1/demo_031_00040.jpg} &\rgbim{result_figs/data_itw/rgb1/demo_031_00040.jpg} &\rgbim{result_figs/data_itw/rgb1/demo_031_00040.jpg} &\rgbim{result_figs/data_itw/rgb1/demo_031_00040.jpg} \\[0.6ex]

        \rowlabel{t@2} &
            \tacim{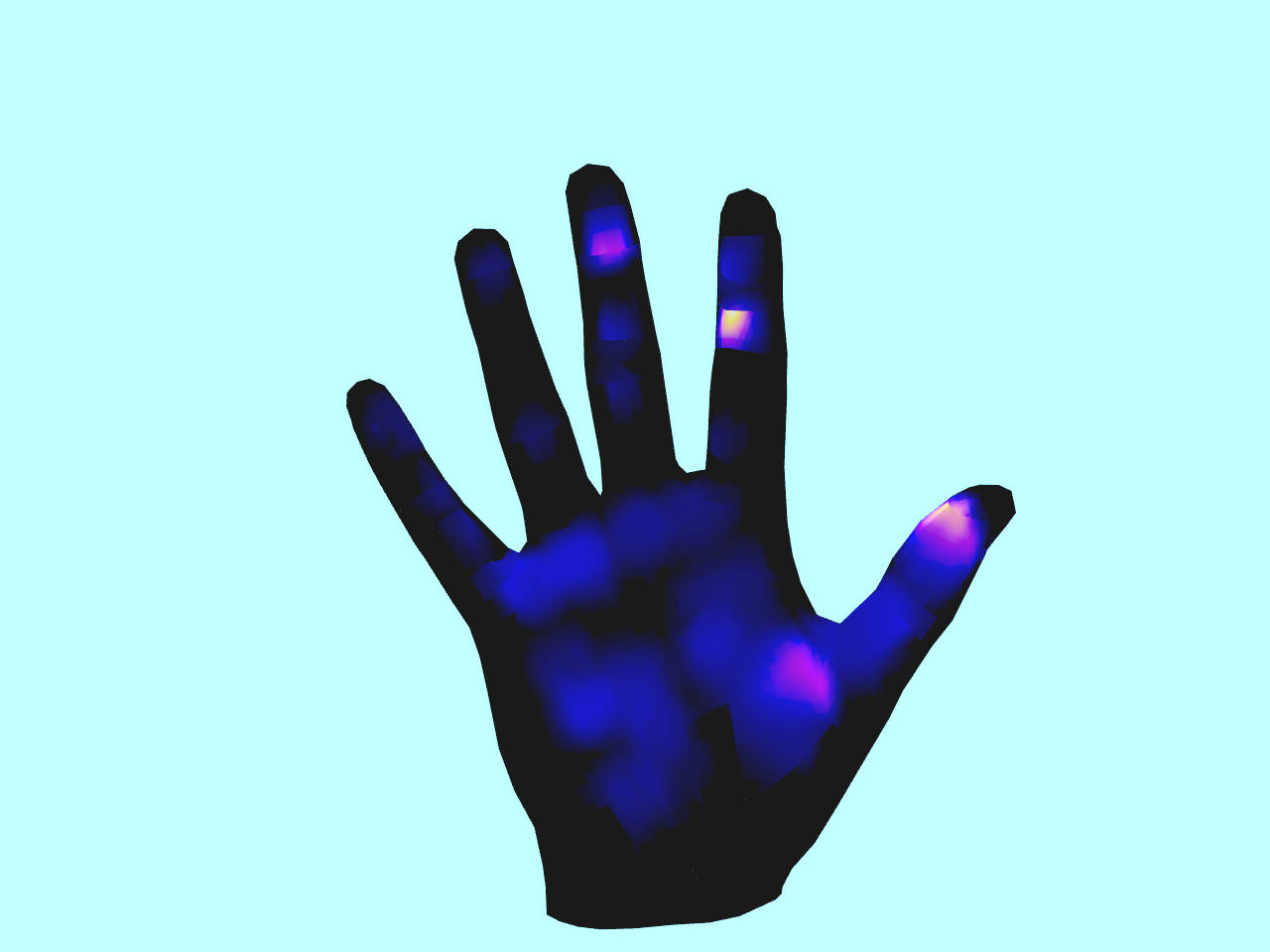} &\tacim{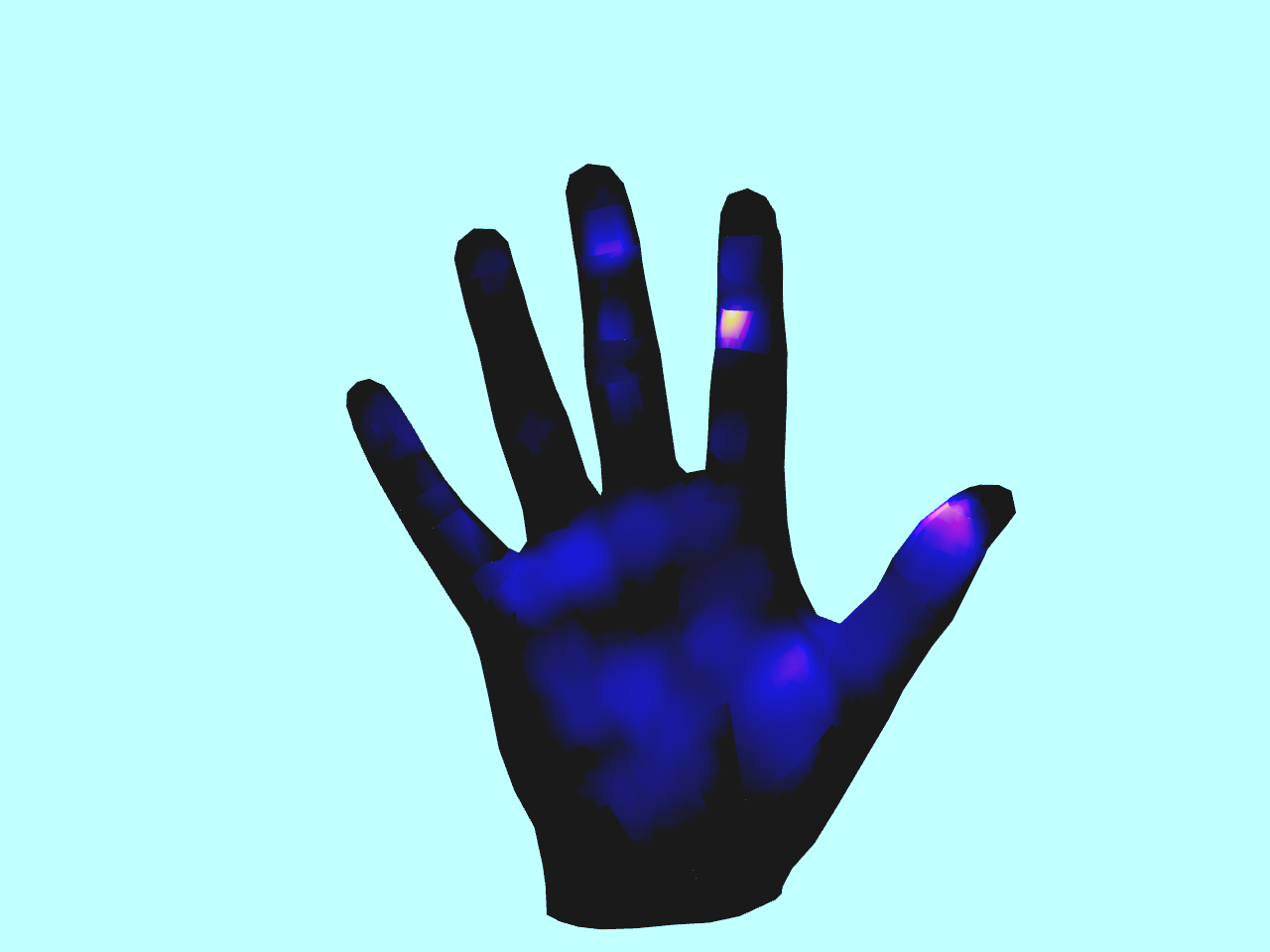} &\tacim{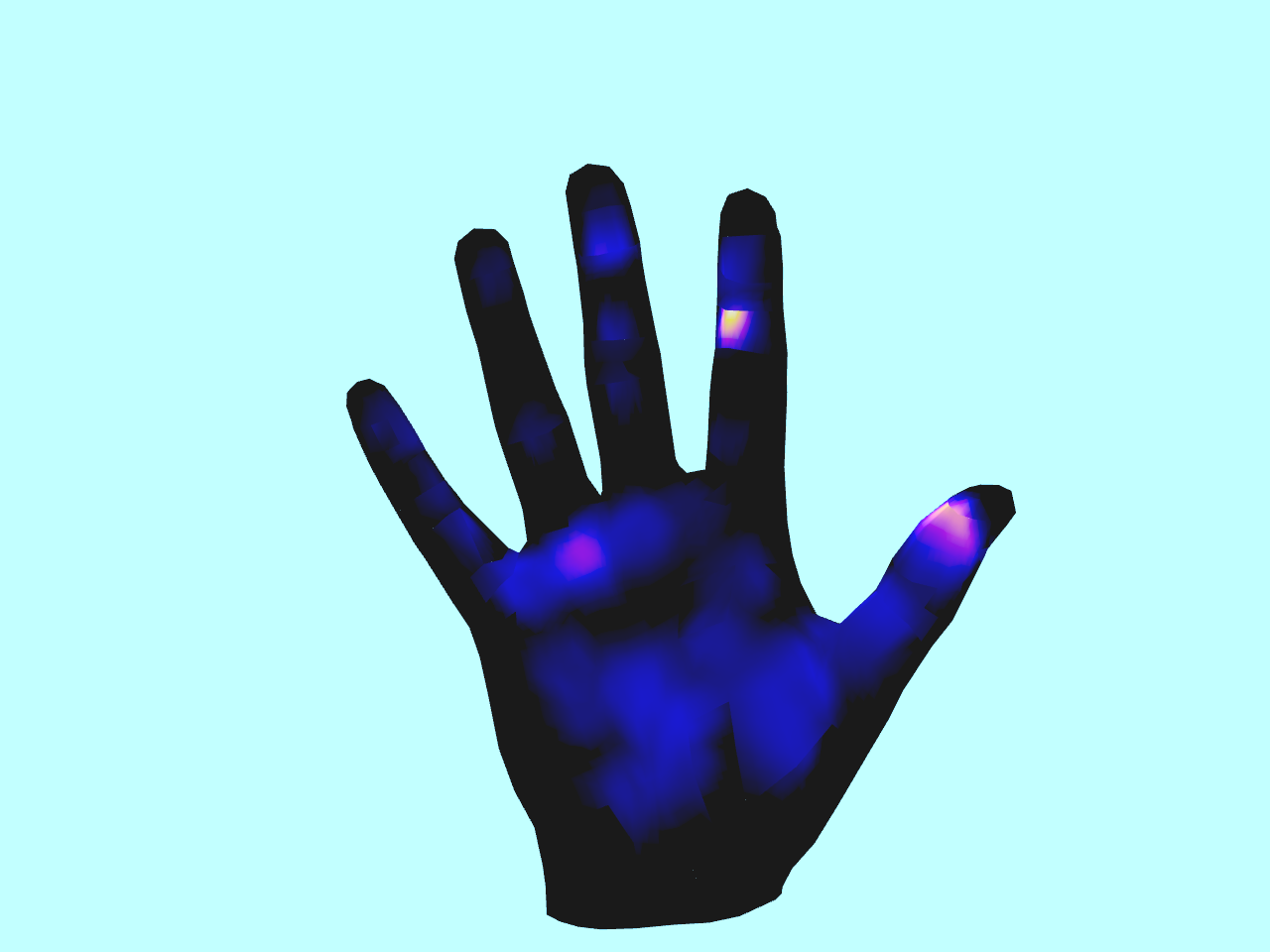} &\tacim{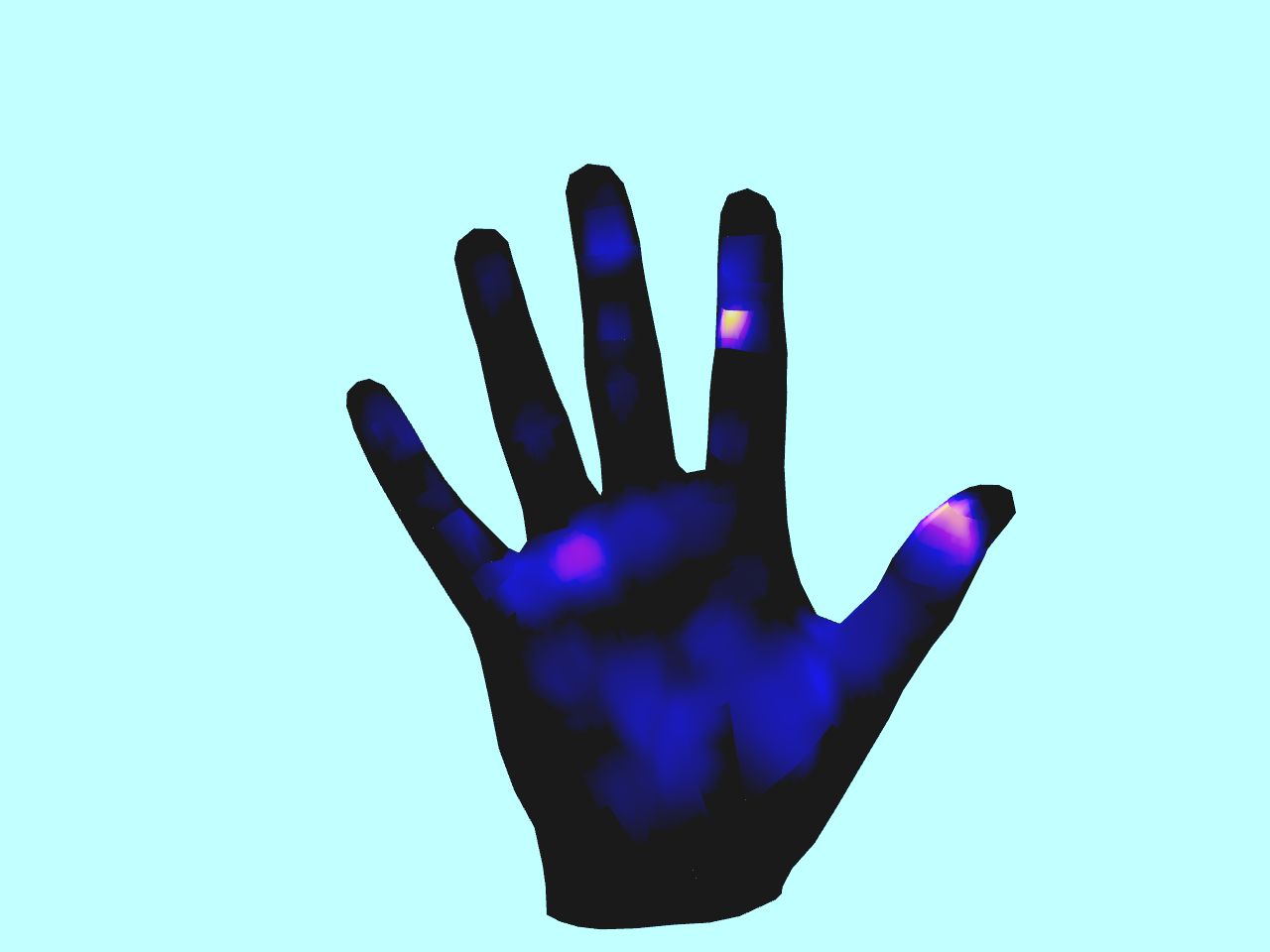} &\tacim{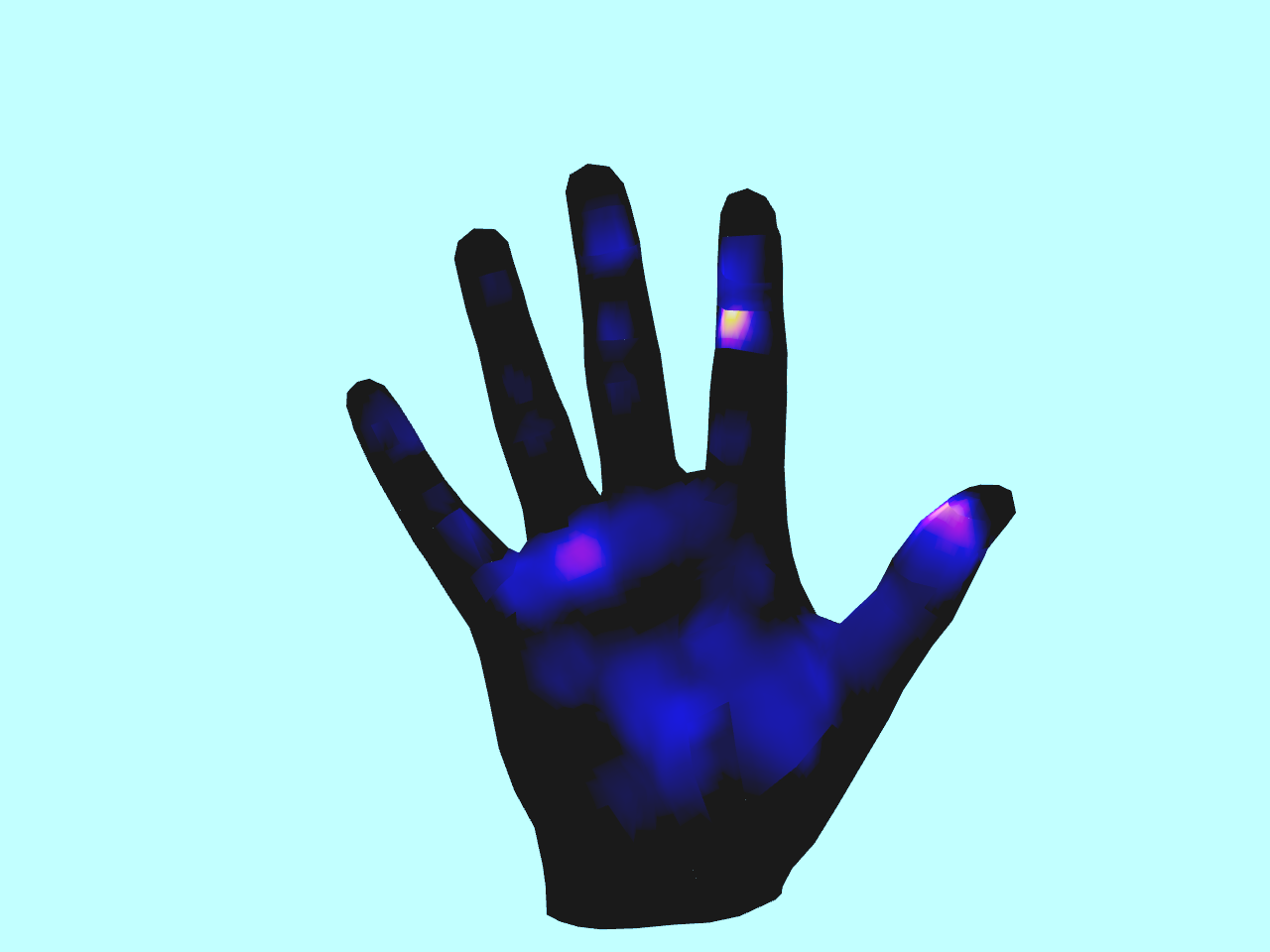} \\

        \rowlabel{v@2} &
            \rgbim{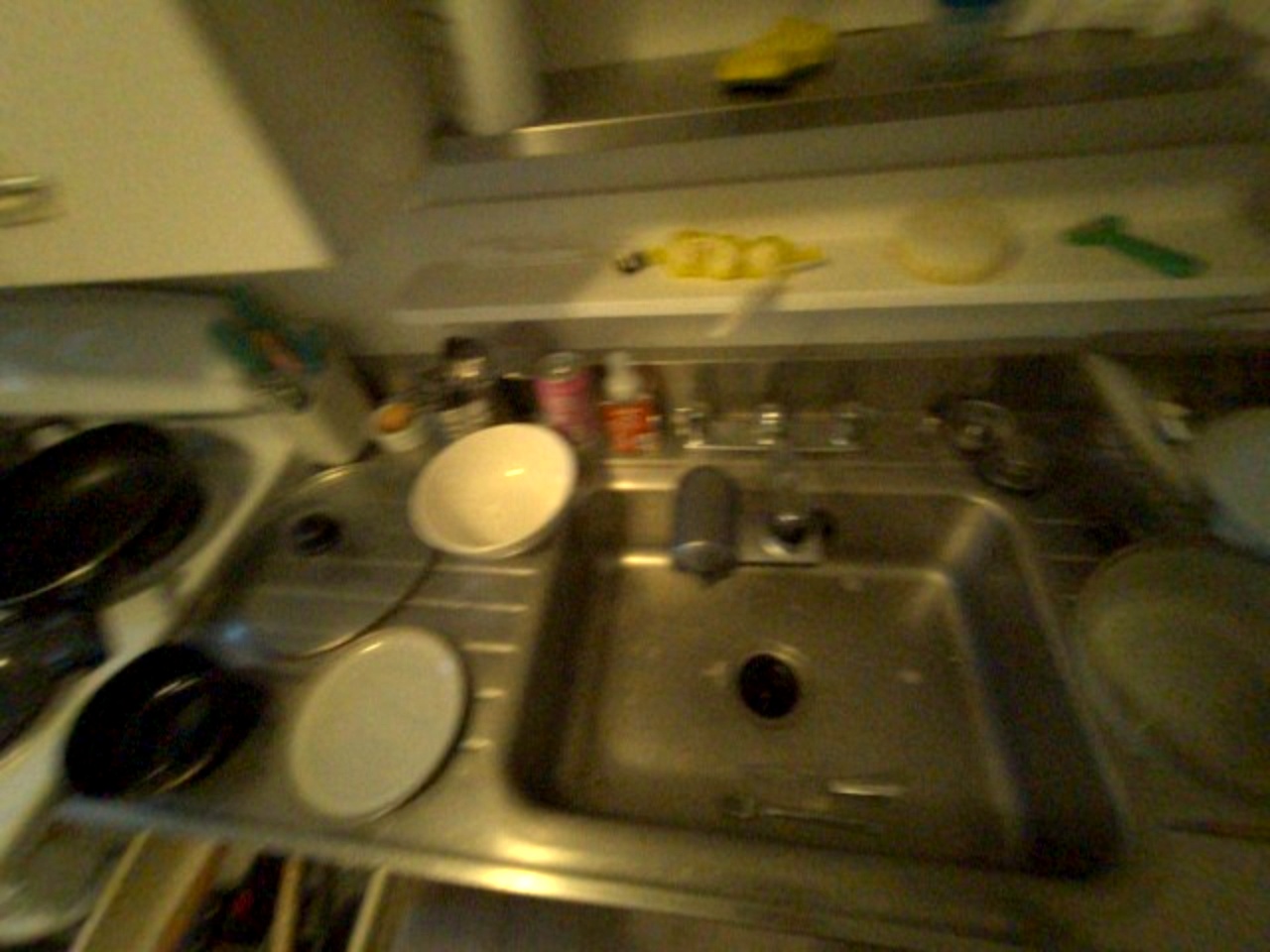} &\rgbim{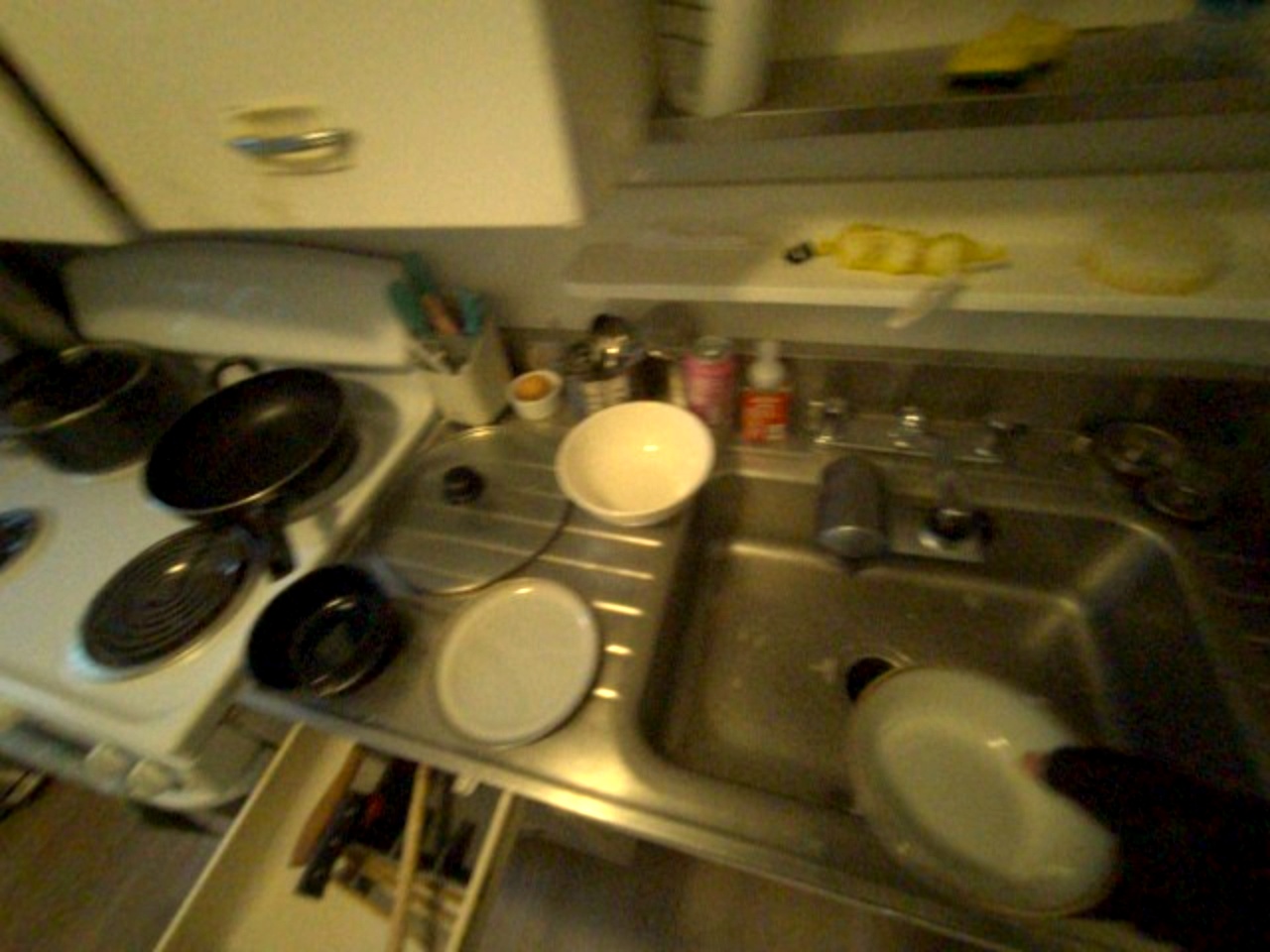} &\rgbim{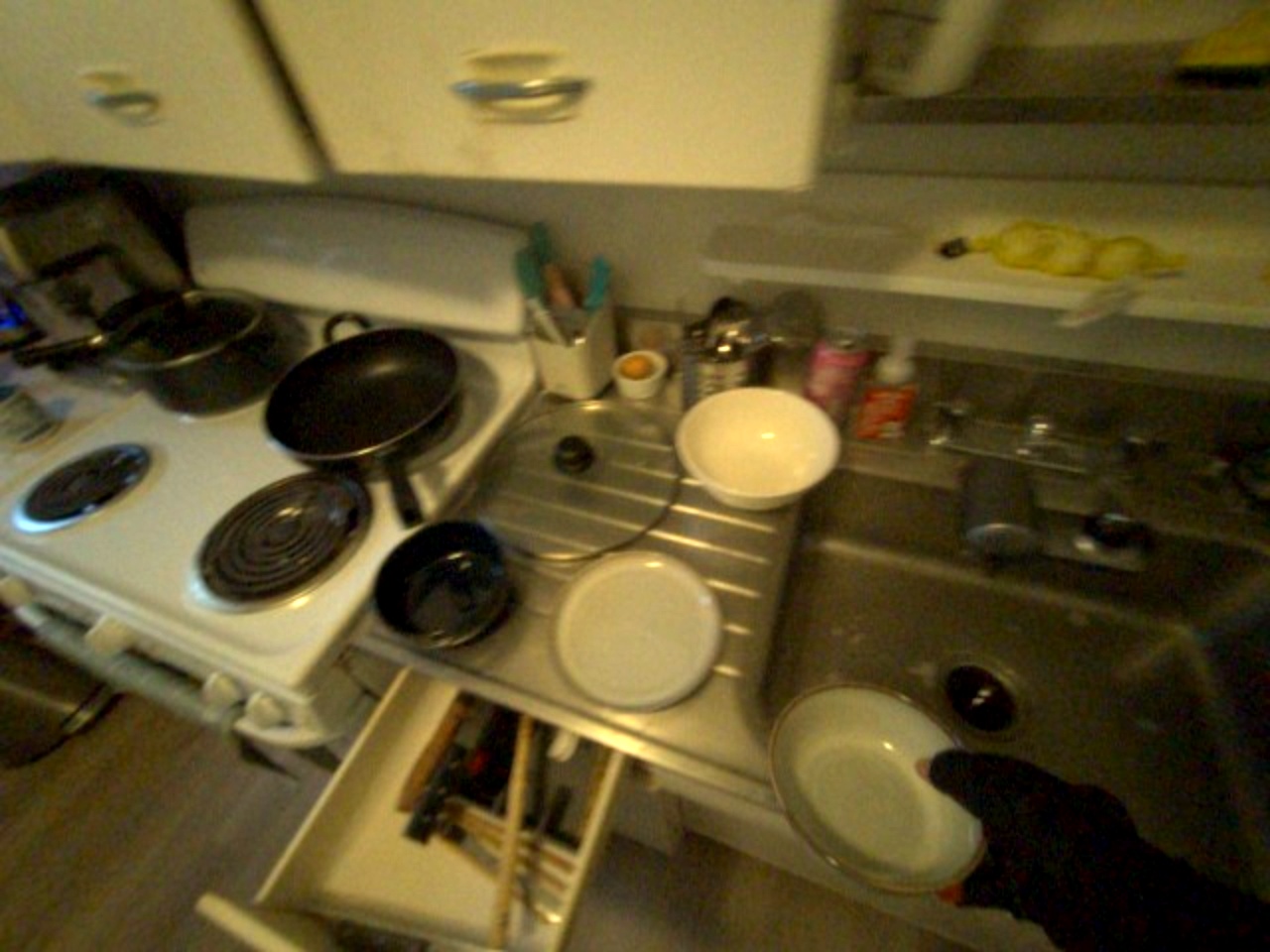} &\rgbim{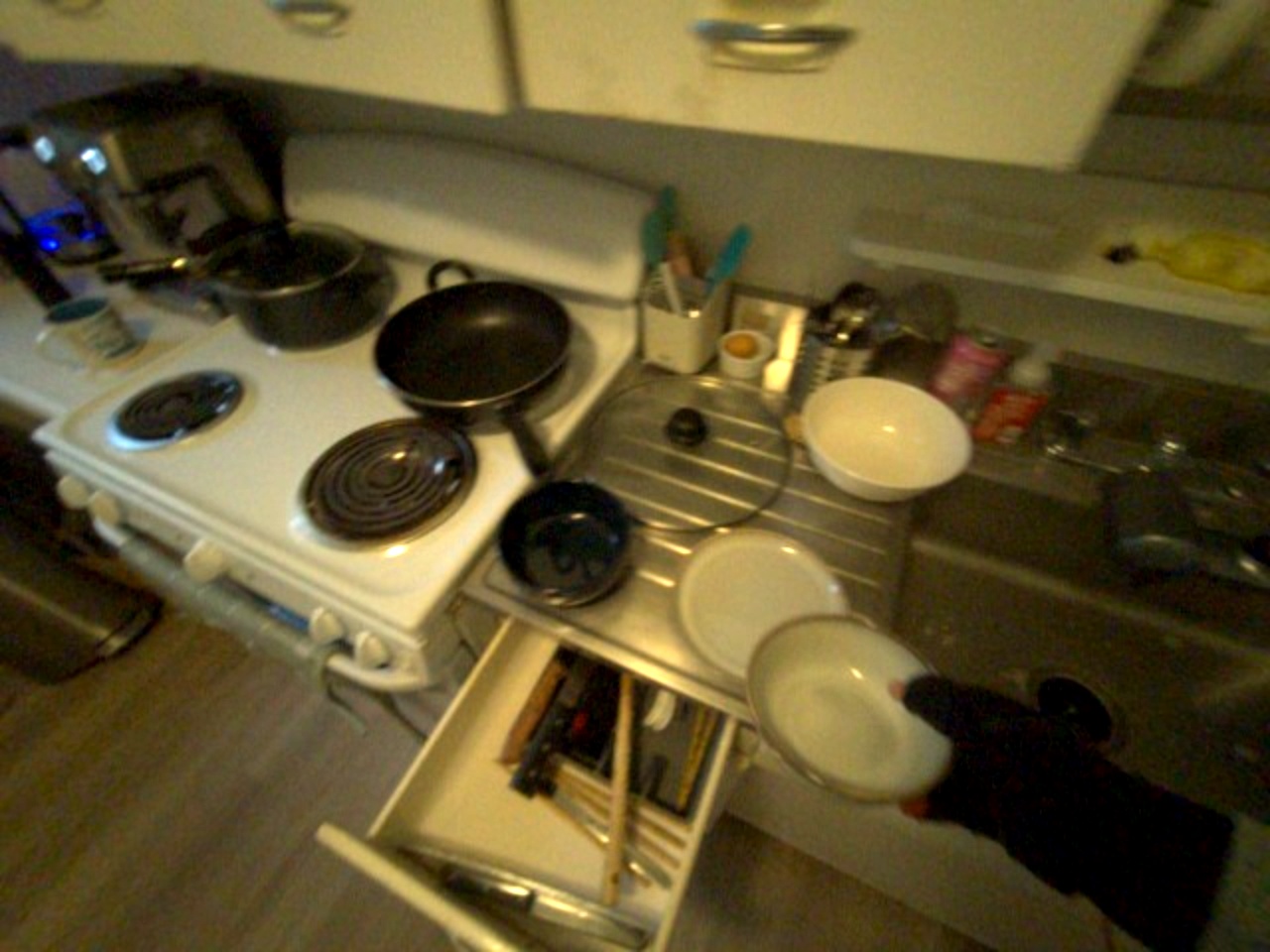} &\rgbim{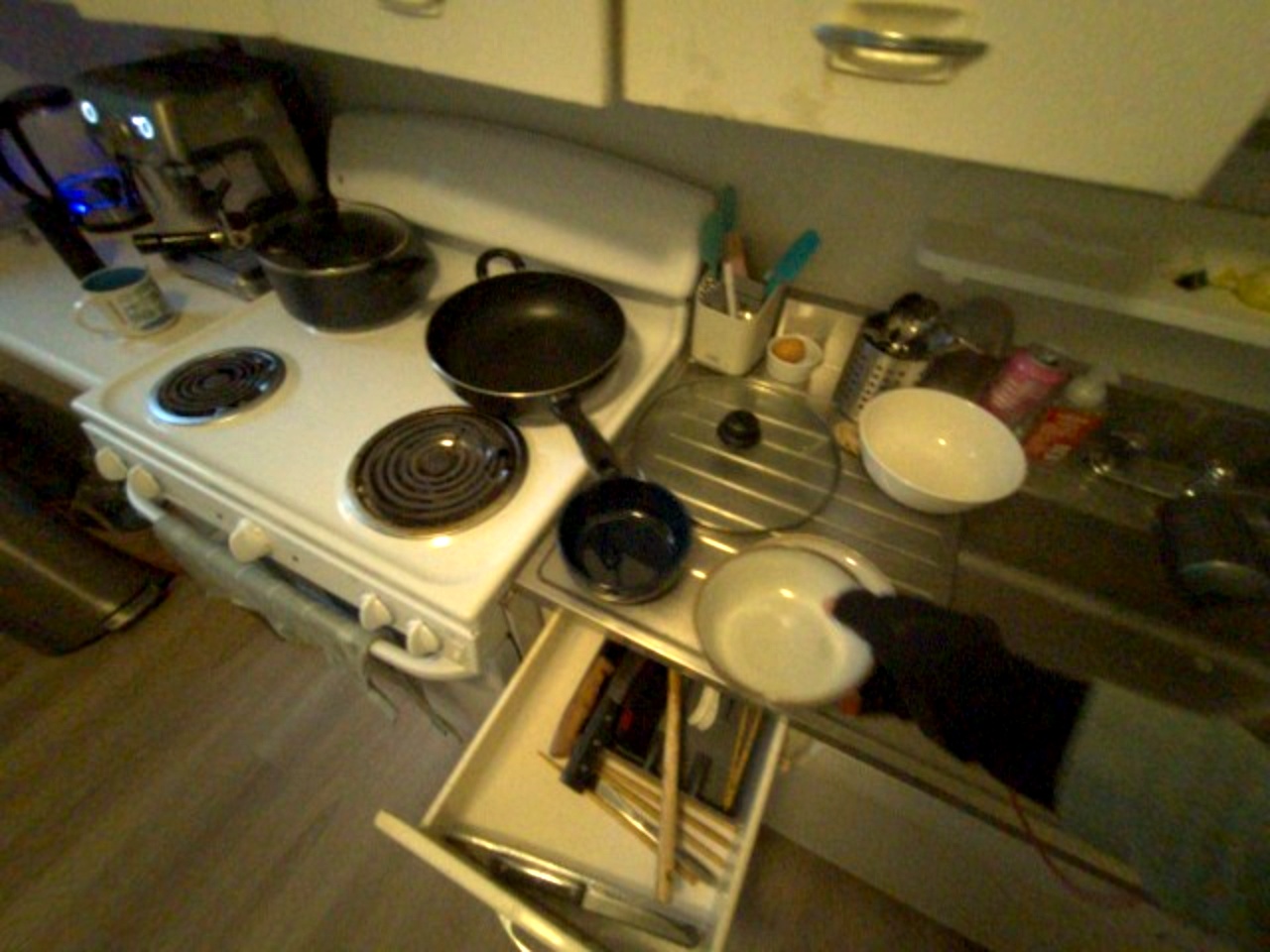} \\
    \end{tabular}

    \caption{\textbf{Qualitative zero-shot tactile retrieval results on Ego4D.} The top row shows the Ego4D query video (five frames). Rows 2-3 and 4-5 show the top-1 and top-2 retrievals from \data{} (tactile sequence and corresponding video). \data provides reliable retrieved tactile pattern, whose corresponding video shows similar hand motion and contact object geometry. See the supplementary video for the full result.}
    \label{fig:qual_itw}
\end{figure}

\subsection{Application}
We also evaluate the generalization capability of our model by performing cross-modal retrieval on the Ego4D dataset~\cite{ego4d}. As illustrated in Figure~\ref{fig:qual_itw}, given an input video query, our model retrieves the most semantically similar tactile signatures from the database. To validate these matches, we examine the source videos paired with the retrieved tactile data and observe that they exhibit human behaviors and manipulation primitives strikingly similar to the query. For extra qualitative results, please refer to our supplementary video.

\section{Conclusion}
\label{sec:conclusion}


We introduced \data, the first egocentric in-the-wild dataset that jointly captures full-hand tactile sensing, synchronized RGB video, and 3D hand pose across diverse everyday environments, providing a concrete foundation for studying how vision, touch, and motion interact during natural manipulation. 
On top of \data, we proposed benchmarks for cross-sensory retrieval and tactile-based grasp classification, showing that multimodal inputs consistently outperform unimodal ones, that tactile is a compact yet powerful signal for grasp understanding and pose prediction, and that temporal structure, lightweight tactile encoders, and moderate discretization significantly improve cross-modal alignment. Taken together, these results position \data\ as a foundation for community-driven progress in touch-centric perception and control, and we hope they bring together researchers in robotics, computer vision, neuroscience, and HCI communities.


\paragraph{Acknowledgment.}
We thank the MIT Office of Research Computing and Data (ORCD) for support through ORCD Seed Fund Grants, which provided access to 8 $\times$ H200 GPUs and additional funding support. We also thank the NVIDIA Academic Grant Program for GPU support, Murata, and Analog Devices for supporting this work through the MIT Gen AI Impact Consortium. Any opinion, findings, and conclusions or recommendations expressed in this material are those of the authors and do not necessarily reflect the views of NVIDIA, Murata, and Analog Devices.

{
    \small
    \bibliographystyle{ieeenat_fullname}
    \bibliography{main}
}

\newpage
\onecolumn
\setcounter{page}{1}
\begin{center}
    \large{\textbf{Supplementary Material}}
\end{center}
\thispagestyle{empty}
\appendix
\tableofcontents
\section{Applications}
\begin{figure}[!h]
    \centering
    \includegraphics[width=0.7\textwidth]{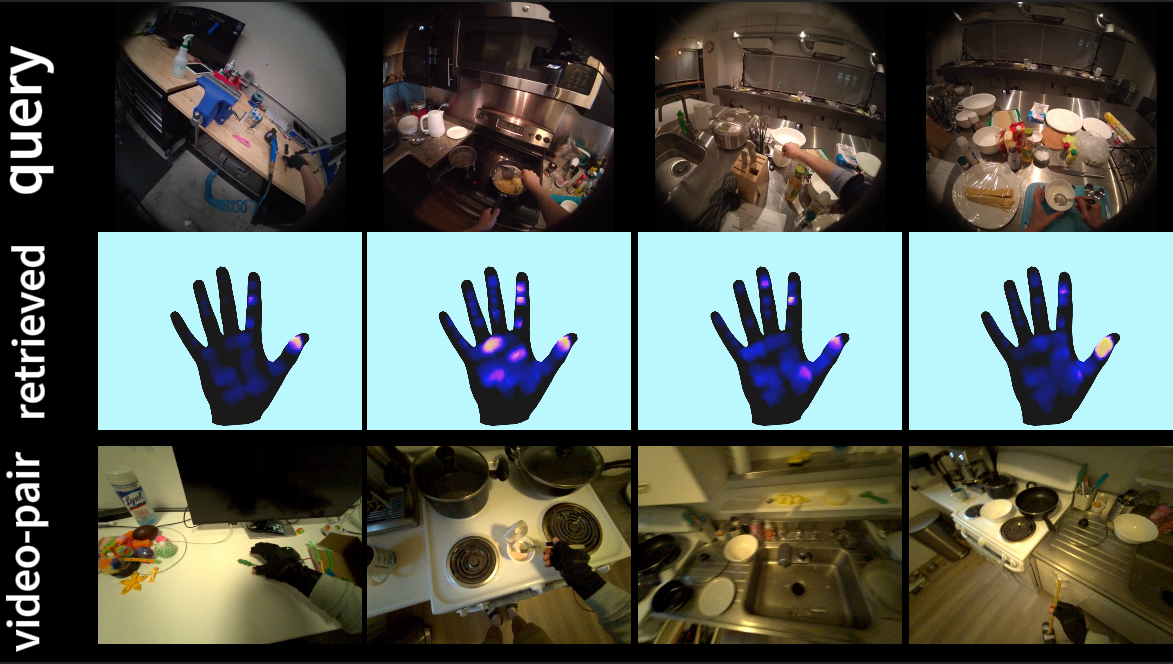}
    \caption{Qualitative zero-shot retrieval results on Ego4D dataset. Please see the attached video for more details.
    }
    \label{fig:ego4d}
\end{figure}

We evaluate the generalization capability of our model by performing cross-modal retrieval on the Ego4D dataset. As illustrated in Figure ~\ref{fig:ego4d}, given an input video query, our model retrieves the most semantically similar tactile signatures from the database. To validate these matches, we examine the source videos paired with the retrieved tactile data and observe that they exhibit human behaviors and manipulation primitives strikingly similar to the query. For extra qualitative results, please refer to our supplementary video (starting at 02:06).

\section{Supplementary Video}
We provide a video visualizing the data points and details, please see the attached video in the zip file.

\begin{figure}[!h]
    \centering
    \includegraphics[width=0.7\textwidth]{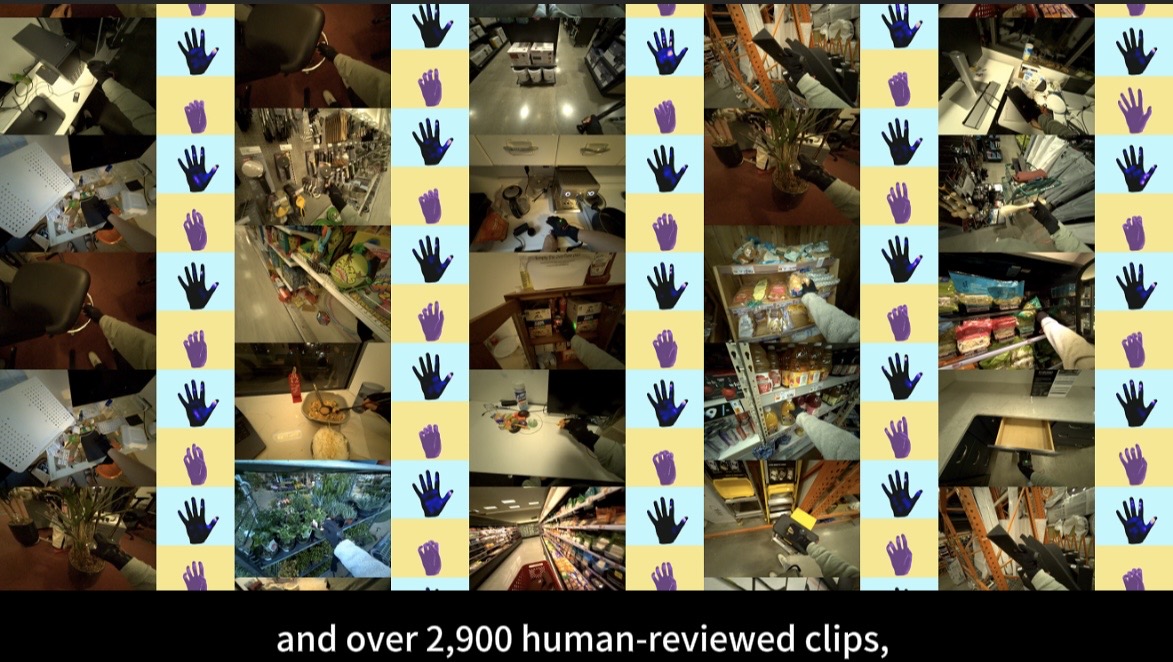}
    \caption{A screenshot from the supplementary video
    }
    \label{fig:supp}
\end{figure}
\newpage
\section{\data~ Visualization}
\subsection{Full Grasp Taxonomy Tactile Map Visualization}
We provide accumulated tactile visualization of all 29 grasp types in OpenTouch. The tactile pattern shows strong correlation with the grasp type.

\begin{figure}[!h]
    \centering
    \includegraphics[width=0.98\textwidth]{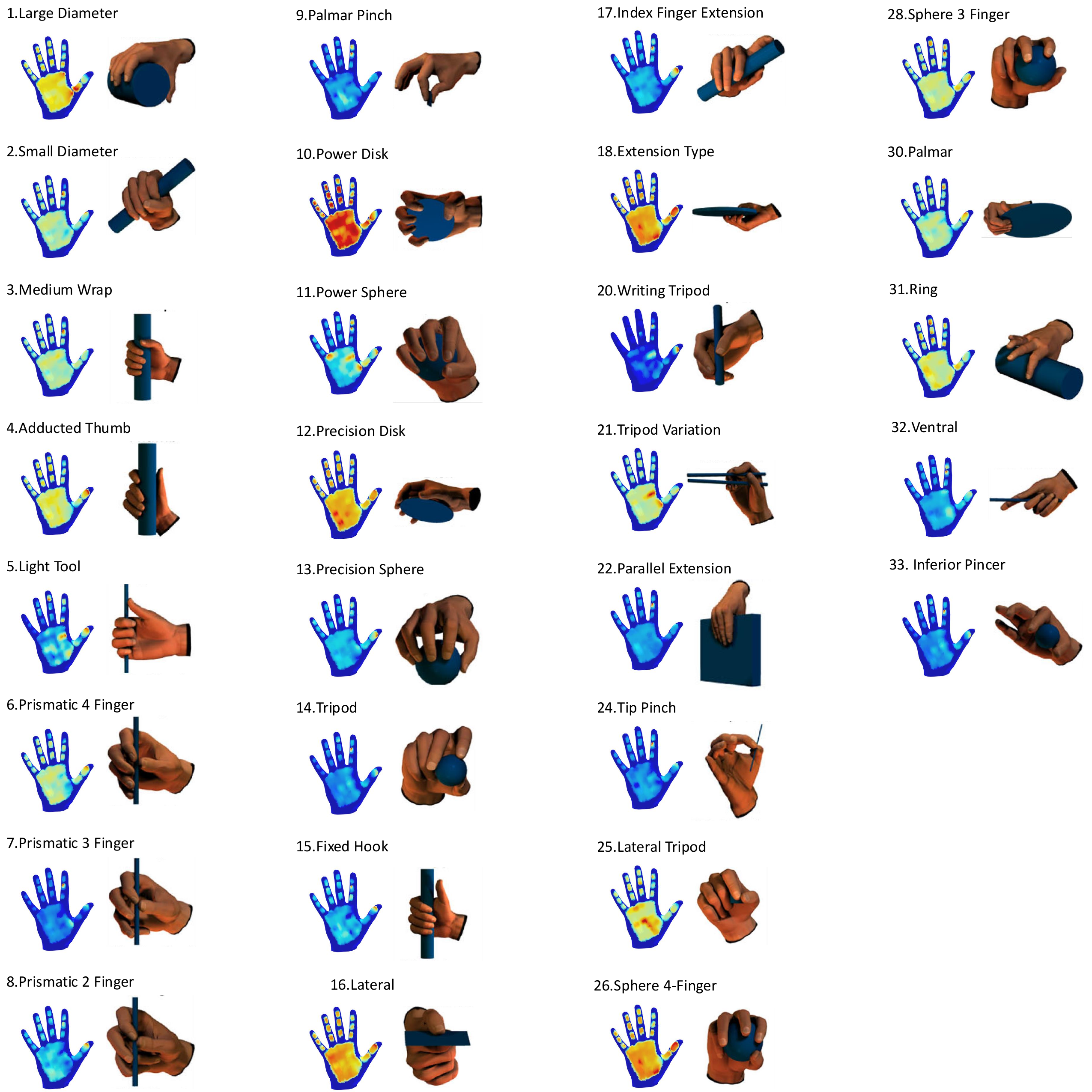}
    \caption{
        \data\ covers 29 grasp types based on the GRASP Taxonomy~\cite{feix2015grasp}.
    }
    \label{fig:full_grasp}
\end{figure}

\subsection{Data Statistic Visualization}

\begin{figure}[!h]
    \centering
    \includegraphics[width=0.98\textwidth]{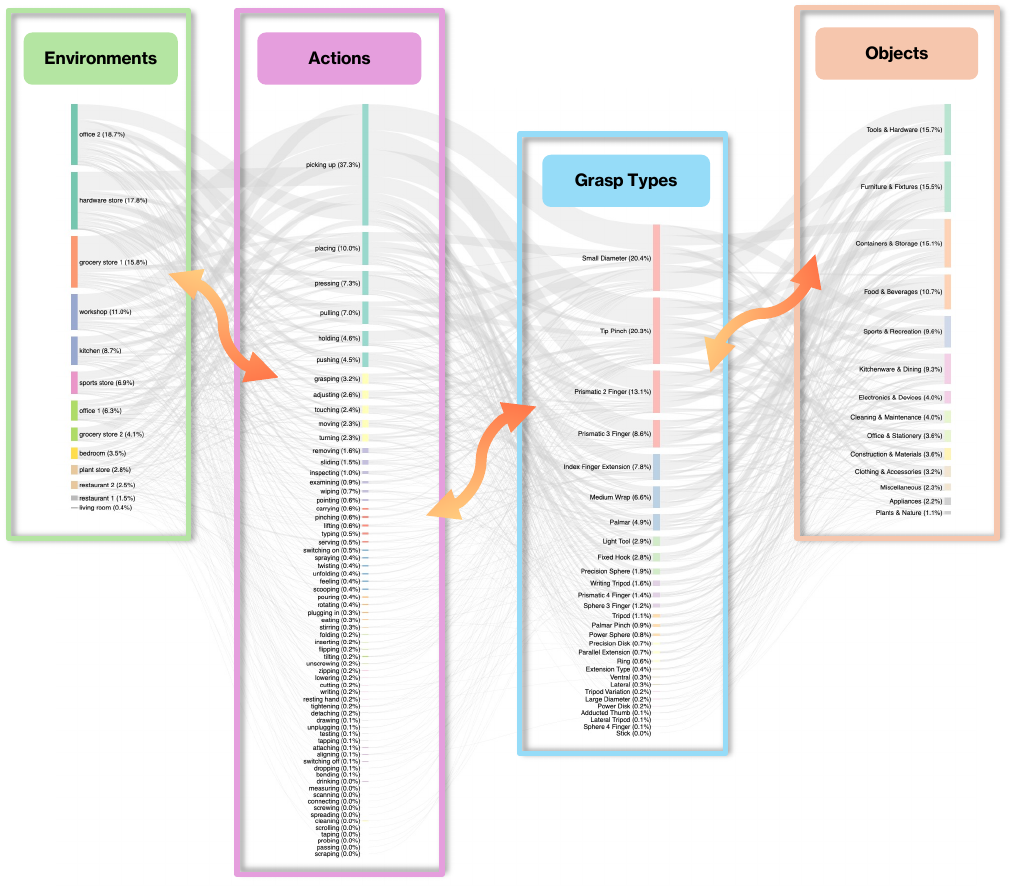}
    \caption{
        \data\ full dataset distribution.
    }
    \label{fig:full}
\end{figure}

We provide the full distribution of our dataset visualization in Figure \ref{fig:full}, a complete diagram showing the hierarchical flow from Environments $\leftrightarrow$ Actions $\leftrightarrow$ Grasp Types $\leftrightarrow$ Object General Categories across full dataset. Note that percentages differ from the top-10 filtered version in our full paper because the top-10 diagram excludes 26.7\% of 
data containing valid but less common action-grip combinations deflating all node percentages. 

\newpage
\section{System Setup}


Important to the design of \data\ was high quality sensing data across our target modalities: touch, pose, and vision. In this section, we describe our hardware setup for tactile data acquisition, hand-tracking, and egocentric RGB camera capture, as well as details surrounding time synchronization and calibration for these sensors.

\subsection{Customized Tactile Sensing Glove.} Existing tactile gloves often face trade-offs between resolution, coverage, and usability. Some designs use low-cost conductive textiles that sacrifice reproducibility requiring specialized machines \cite{BUSCHER2015glove, luo2024tactile}. These constraints make it difficult to capture dense tactile data in unconstrained, everyday environments. We fabricated a hand-shape tactile sensor that is thin, low-cost, rapidly manufactured, and fully open-source, enabling large-scale multimodal data collection \textit{in the wild}. Our design leverages FPCB technology to automatically route 16×16 electrodes around a commercial piezoresistive film, forming 169 sensors that uniformly cover the fingers and palmar surface of the hand. This approach combines the fabrication precision and reliability of electronics manufacturing with the mechanical compliance required for wearable sensing, achieving stable, high-resolution pressure mapping without bulky cabling or manual wiring \cite{murphy2025flexglove}.  

\begin{figure}[!h]
    \centering
    \includegraphics[width=0.98\textwidth]{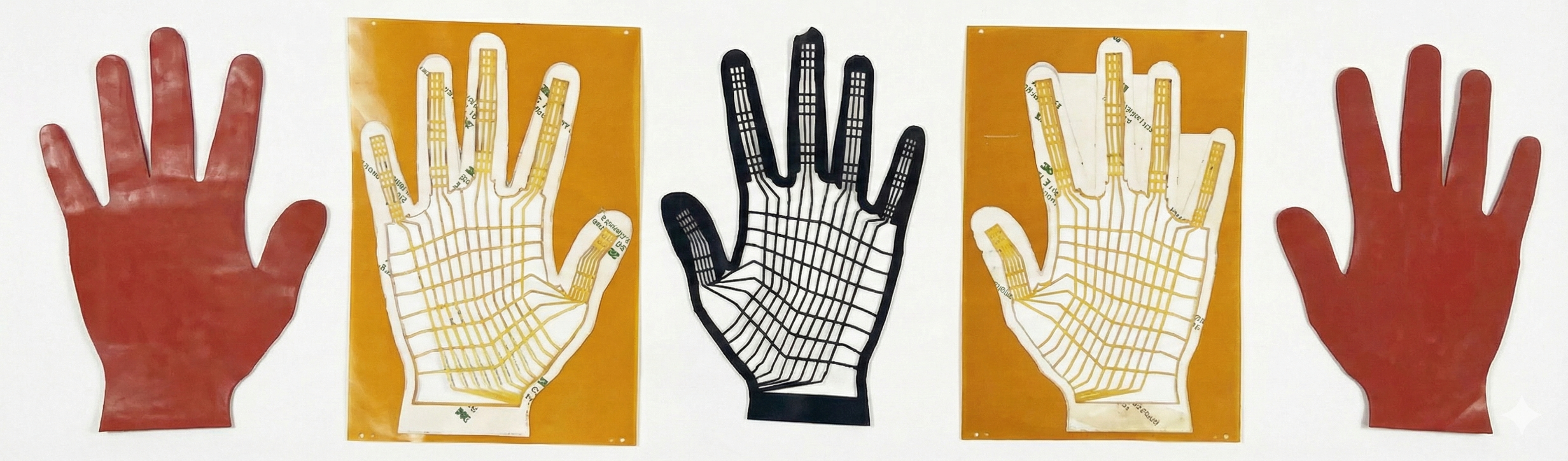}
    \caption{Our sensor has 5-layered piezoresistive structure: top silicone encapsulation -- top fpc -- piezoresistive file -- bottom fpc -- bottom silicone encapsulation \cite{murphy2025flexglove}.
    }
    \label{fig:grasp}
\end{figure}

\subsection{Hand-tracking Glove.} Hand pose data was captured using Rokoko Smartglove \cite{rokoko}, a professional grade motion capture system that utilizes an IMU and EMF sensor fusion approach. This system was selected due to its high-fidelity, ease of use, and integration capabilities for creating a multimodal dataset. Each glove has seven 6 DOF sensors to capture full rotational data of the hand and fingers. Data is streamed via a portable Wi-Fi at an update rate of 30Hz. The glove's rotational dynamic accuracy is reported to be $\pm1$ degree, while the displacement accuracy is a function of both the rotational accuracy and kinematics of the body. Prior to capturing data, the gloves require a system-defined calibration pose in which the subject must stand upright with their arms bent at the elbow and kept at a 90 degree angle to their body. This helps define a consistent zero-point that can then be used for accurate data transformation. Data is streamed from gloves to the Rokoko Studio software, where it can then be synchronized with egocentric vision and tactile data.

\subsection{Egocentric Video Capture.}
We capture egocentric video using Meta’s Project Aria glasses \cite{project_aria_research_kit}, which are equipped with two mono SLAM cameras, an RGB point-of-view camera with a $110^{\circ}$ field of view, eye-tracking cameras, IMUs, microphones, and more (Profile 28). Using the Project Aria companion app, we configure a recording profile that includes the RGB camera (1408×1408 pixels at 30 Hz), IMU data, and SLAM camera streams. For our benchmarks, we focus on the RGB video data. Recordings are stored locally on the glasses in the VRS format and later transferred to a desktop host for time synchronization with the tactile and pose data.

\begin{table}[t]
\centering
\scriptsize
\setlength{\tabcolsep}{8pt}
\caption{\textbf{We used profile 28 from Project Aria recording profiles.}}
\begin{tabular}{lcccccc}
\toprule
\textbf{Sensor} &
\textbf{Channels} &
\textbf{Sample Rate (kHz)} &
\textbf{Resolution} &
\textbf{FPS} &
\textbf{Auto Exp.} &
\textbf{Format} \\
\midrule
Microphones & 7 & 28 & -- & -- & -- & -- \\
EyeTracking Cameras  & -- & -- & 320$\times$240 & 60 & -- & JPEG \\
RGB Cameras & -- & -- & 1408$\times$1408 & 30 & ON & JPEG \\
SLAM Cameras & -- & -- & 640$\times$480 & 30 & ON & JPEG \\
IMU 1        & -- & 1000  & -- & -- & -- & -- \\
IMU 2        & -- & 800   & -- & -- & -- & -- \\
Magnetometer & -- & 10    & -- & -- & -- & -- \\
Barometer    & -- & 50    & -- & -- & -- & -- \\
\bottomrule
\end{tabular}
\end{table}

\subsection{Time Sync and Calibration.}
We synchronize vision, tactile, and hand‑pose data using a visually recorded start cue captured by the glasses’ RGB camera. When tactile and hand‑pose recording begins, a distinctive command line cue is shown and recorded by the camera; in post‑processing we locate the RGB frame containing this cue and take its device‑clock timestamp as the reference. We then find the nearest tactile and hand‑pose samples in their native timestamps, treat those samples as the shared origin, and convert both streams into the device clock domain by subtracting their matched sample time and adding the reference timestamp, discarding any samples that precede the cue. This yields a consistent timeline across modalities.

\subsection{3D Tactile Map Visualization}
To visualize full-hand contact in 3D, we first align the 2D taxel grid to a high-resolution hand mesh with a finest edge resolution of 0.004.
For each taxel, we find nearby mesh faces within a distance threshold of 0.005 on the fingertips and 0.007 on the palm region, and assign to each face the weighted average force of its associated taxels.
The resulting per-face pressures are then smoothed with a spatial Gaussian filter, producing a continuous 3D tactile field over the hand surface.

\subsection{Limitations.}
Touch involves multiple cues: normal pressure, shear, micro-vibration, and temperature, but our piezoresistive array measures only normal pressure; as a result, slip events, texture-induced vibrations, and thermal cues are not directly captured. Coverage is limited to the palmar side with a 16 $\times$ 16 grid, which constrains fingertip spatial resolution and the localization of small contact patches; absolute forces outside the calibrated 0.02–50kPa \cite{murphy2025flexglove} range may saturate. Durability is another concern: the FPCB experiences repeated bending and local folding around finger joints; despite bench tests to $\sim$10k cycles, real-world use (donning/doffing friction, sweat ingress at edges, incidental stretching/abrasion) can fracture copper traces or accelerate delamination, reducing sensor lifetime. Wireless streaming (ESP-NOW) and host-side timestamping introduce exposure to packet loss, RF interference, and clock drift, which can cause occasional dropped frames or sub-frame misalignment with video/pose streams. Fit and ergonomics may alter natural manipulation: a single glove size can change contact geometry and friction, the protective stack adds compliance and thickness. 


\newpage
\section{Implementation Details, Extra Experiments and Discussions}
In this section, we provide detailed specifications regarding our data annotation prompts and experimental setup. We first describe the network architectures for the visual, tactile, and pose branches, followed by the specific training configurations and loss functions employed in our model. Finally, we present supplementary experiments that offer further insights into the behavior of the retrieval and classification tasks.

\subsection{Annotation Prompt}

\paragraph{Prompt for GPT-5 annotation.}

\vspace{2mm}
\begin{center}
\begin{tcolorbox}[
    title={GPT Annotation Prompt},
    colback=gray!5,
    colframe=gray!40,
    width=0.95\textwidth  
]
\footnotesize

You are an expert multimodal annotator specializing in human--object interaction, haptics, and hand biomechanics.
You will analyze a sequence of images showing a hand--object interaction.
The images consist of three time-ordered pairs sampled by pressure phase:
\begin{itemize}[leftmargin=*]
    \item Frame 1 = \textbf{Onset}: lowest right-hand contact pressure before the peak (approach/initial touch)
    \item Frame 2 = \textbf{Peak}: highest right-hand contact pressure (max manipulation force)
    \item Frame 3 = \textbf{Post-peak}: lowest right-hand contact pressure after the peak (release/withdrawal)
\end{itemize}

Each pair contains:
\begin{itemize}[leftmargin=*]
    \item An RGB camera frame (from wearable glasses, egocentric view)
    \item A tactile heatmap showing right-hand contact
\end{itemize}

Use these predefined label sets for classification:
\begin{itemize}[leftmargin=*]
    \item \textbf{Object Categories:} \texttt{\{sorted(list(object\_categories))\}}
    \item \textbf{Environments:} \texttt{\{sorted(list(environments))\}}
    \item \textbf{Actions:} \texttt{\{sorted(list(actions))\}}
    \item \textbf{Grip Types:} \texttt{\{sorted(list(grasp\_types))\}}
\end{itemize}

Your task is to analyze the visual and tactile information and output a structured JSON annotation with these fields:
\begin{enumerate}[leftmargin=*]
    \item \textbf{object\_name} --- The specific object being manipulated by the right hand (free text, e.g., ``Instant Pot lid'', ``coffee grinder'').
    \item \textbf{object\_category} --- General class of the object (choose from the provided set).
    \item \textbf{environment} --- General setting (choose from the provided set).
    \item \textbf{action} --- The primary manipulation action (choose from the provided set).
    \item \textbf{grip\_type} --- The hand grasp type (choose from the provided set).
    \item \textbf{description} --- A fluent 2--4 sentence narrative that:
    \begin{itemize}
        \item Identifies the specific object being manipulated by the right hand, with details about appearance, size, material, color.
        \item Situates the environment, including setting, surface, background, lighting.
        \item Summarizes the manipulation action across all three frames.
    \end{itemize}
\end{enumerate}

\noindent\textbf{Important:} Focus specifically on what the right hand is doing and touching.
The right hand wears a black glove, but do not mention the glove in your description.
If the right hand's interaction is unclear, do not default to describing what the left hand is doing.

Before providing your final JSON output, work through your analysis systematically in \texttt{\textless thinking\textgreater} tags:
\begin{quote}
\texttt{\textless thinking\textgreater}\\
\textbf{Frame-by-frame examination}: Describe what you observe in each frame sequentially, focusing on the right hand's interaction.\\
\textbf{Object identification}: Based on your observations, identify the specific object the right hand is manipulating, including physical characteristics.\\
\textbf{Environment assessment}: Describe the setting, surface, background, lighting, and other visible objects.\\
\textbf{Action analysis}: Determine what manipulation action is being performed by the right hand.\\
\textbf{Grasp classification}: Identify the hand grasp type, considering multiple possibilities and explaining your reasoning.\\
\textbf{Category mapping}: For each classification field, consider 2--3 potential options from the predefined lists and explain which best fits.\\
\textbf{Description planning}: Draft key points for your natural language description.\\
\texttt{\textless /thinking\textgreater}
\end{quote}
...
\end{tcolorbox}
\end{center}

\begin{center}
\begin{tcolorbox}[
    title={GPT Annotation Prompt},
    colback=gray!5,
    colframe=gray!40,
    width=0.95\textwidth  
]
\footnotesize
...
Here are examples of expected outputs:

\noindent\textbf{Example 1:} Person pressing down an Instant Pot lid
\begin{lstlisting}[language=json,basicstyle=\ttfamily\footnotesize]
{
  "object_name": "Instant Pot lid",
  "object_category": "appliance",
  "environment": "kitchen",
  "action": "pressing",
  "grip_type": "Medium Wrap",
  "description": "A stainless lid is aligned over a 
  countertop pressure cooker in a kitchen. The right 
  hand settles on the rim with a firm wrap and 
  presses down to seat the lid. Contact shifts toward 
  the palm as the lid is pushed and locked into place."
}
\end{lstlisting}

\noindent\textbf{Example 2:} Person using a manual coffee grinder
\begin{lstlisting}[language=json,basicstyle=\ttfamily\footnotesize]
{
  "object_name": "manual coffee grinder",
  "object_category": "appliance",
  "environment": "kitchen",
  "action": "rotating",
  "grip_type": "Small Diameter",
  "description": "A compact cylindrical grinder is 
  held over a kitchen counter. The right hand wraps 
  around the small body while rotating the handle in 
  short, continuous motions. Tactile contact 
  concentrates on the fingertips and distal pads as 
  the device is turned."
}
\end{lstlisting}

\noindent\textbf{Example 3:} Person turning a metal handle to open a door
\begin{lstlisting}[language=json,basicstyle=\ttfamily\footnotesize]
{
  "object_name": "door handle",
  "object_category": "handle",
  "environment": "office",
  "action": "turning",
  "grip_type": "medium wrap",
  "description": "In a hallway scene, the right hand 
  encloses a metal lever handle. The grip firms and 
  rotates downward to actuate the latch, then the 
  handle is released as the door begins to open."
}
\end{lstlisting}

After completing your analysis in the \texttt{\textless thinking\textgreater} tags, provide your final output as a JSON object only.
Use only the allowed labels for all fields except \textbf{object\_name} and \textbf{description}.
If uncertain about a label, pick the closest option or ``unknown''.

\end{tcolorbox}
\end{center}

\vspace{2mm}
\subsection{Modality Encoding}
We use three modality-specific encoders, each processing sequences of $N$ frames sampled at 30 Hz. All three modalities are linearly projected into $d$-dimensional space for cross-modal alignment and downstream classification.
 
\vspace{1mm}
\noindent \textbf{Visual encoder.} We leverage the DINOv3 architecture~\cite{simeoni2025dinov3}, specifically the ViT-B/16 variant (\textit{dinov3-vitb16-pretrain-lvd1689m}), to extract semantic representations. The backbone parameters are frozen during training. Given a video sequence, the model processes each frame independently to extract features. These frame-level features are temporally aggregated via mean pooling to form a video-level representation, which is subsequently mapped to the shared embedding space using a linear projection layer.

\vspace{1mm}
\noindent \textbf{Tactile encoder.} The tactile encoder treats the sequence of $16\times16$ pressure maps as a single-channel video stream. Inspired by prior studies demonstrating that lightweight convolutional architectures can effectively model sequence tactile signals~\cite{luo2021learning, umi-touch, luo2024tactile}, we employ a three-layer Convolutional Neural Network (CNN)~\cite{lecun2002gradient} to extract spatial features. Each layer consists of 32 filters with $5\times5$ kernels, followed by ReLU activation~\cite{nair2010rectified} and $2\times2$ max-pooling. The resulting features are flattened and fed into a two-layer bidirectional Gated Recurrent Unit (GRU)~\cite{ravanelli2018light}  with a hidden dimension of 120 to capture temporal dependencies. The final representation is formed by concatenating the last hidden states of the forward and backward passes, followed by a ReLU activation and linear projection into the shared embedding space.

\vspace{1mm}
\noindent \textbf{Pose encoder.} The pose encoder processes sequences of 3D hand landmarks (21 keypoints)  to capture structural hand dynamics. Following~\cite{li2025multimodal}, we employ a four-layer MLP to extract frame-wise features. Prior to encoding, the raw coordinates undergo geometric normalization to ensure invariance to global translation and scale while preserving relative orientation. Finally, the extracted features are aggregated via adaptive temporal average pooling and linearly projected into the shared embedding space.

\subsection{Cross-modal Retrieval}

\vspace{1mm}
\noindent\textbf{Linear cross-modal retrieval.} We follow the implementation of ObjectFolder’s linear baselines~\cite{gao2023objectfolder}. Each modality (vision, touch, and pose) uses an independent backbone for feature extraction, and the resulting features are linearly projected into a shared 64-dimensional embedding space for cross-modal alignment. Canonical Correlation Analysis (CCA)~\cite{10.1145/1873951.1873987} maximizes the correlation between paired modalities, while Partial Least Squares Correlation Analysis (PLSCA)~\cite{jong2001canonical} extends CCA by incorporating partial least squares (PLS) to find projections that maximize the shared covariance between modalities. We empirically evaluate CCA/PLSCA with different component dimensions. SVD fails to converge at higher dimensions (d $\ge$ 24). We report results at the highest stable dimension (d$=$22).

\vspace{1mm}
\noindent\textbf{Deep cross-modal retrieval.} We train our model to learn a shared multimodal representation space where sequence of visual, tactile, and pose embeddings are aligned semantically. For each modality, we process a sequence of $N=20$ frames. Following~\cite{girdhar2023imagebind}, the outputs from each encoder $f_V$, $f_T$, and $f_P$ are linearly projected to a 64-dimensional space and L2-normalized, resulting in embeddings $z_v, z_t, z_p \in \mathbb{R}^{64}$. we adopt a symmetric InfoNCE objective, where matching pairs are treated as positives and all other samples in the batch serve as negatives.
For a batch of size $B$, the loss between two modalities 
$(a,b)\in\{(v \leftrightarrow t),(t \leftrightarrow p)\}$ is:
\begin{equation}
\mathcal{L}_{a \to b}
= -\frac{1}{B}\sum_{i=1}^B
\log
\frac{\exp\big(\langle z_a^{(i)}, z_b^{(i)} \rangle / \tau \big)}
{\sum_{j=1}^B \exp\big(\langle z_a^{(i)}, z_b^{(j)} \rangle / \tau \big)},
\label{eq:infonce}
\end{equation}
where $\tau=0.07$ is the temperature~\cite{wu2018unsupervised}. 
The total loss is:
\begin{equation}
\mathcal{L}
= \mathcal{L}_{a \to b} + \mathcal{L}_{b \to a}\big.
\label{eq:bidirectional}
\end{equation}

\noindent For multimodal queries, such as $(a,b \to c)\in\{( v + p \leftrightarrow t),(t + p \leftrightarrow v), (v + t \leftrightarrow p)\}$ , the embeddings are first fused through a lightweight linear head $\phi(\cdot)$:
\begin{equation}
z_{f} = \phi([z_a; z_b]) \in \mathbb{R}^{64},
\label{eq:fusion}
\end{equation}
and the same InfoNCE formulation (Equ.\ref{eq:infonce}) is applied between the fused query $z_{f}$ and the target embedding $z_c$.
The model is trained for 300 epochs using the Adam optimizer~\cite{kingma2014adam}
with a learning rate of $\times10^{-4}$ and a batch size of 256.
The learning rate follows a cosine annealing schedule with a 5-epoch warm-up phase.

\subsection{Tactile Pattern Classification}
\vspace{1mm}
\noindent\textbf{Tactile pattern classification.} For classification, we attach a lightweight prediction head on top of the fused embedding.
We train two independent single-task models using the same backbone: one for action recognition and the other for grasp type classification.
Each model takes a sequence of $N{=}10$ frames as input and outputs a single fused embedding, which is passed to a linear classification head consisting of a fully connected layer followed by a softmax over the corresponding number of classes. The classifier is trained using the cross-entropy loss~\cite{zhang2018generalized} and optimized with the Adam optimizer~\cite{kingma2014adam} using a learning rate of $1\times10^{-4}$ and a batch size of 64.



\vspace{2mm}
\noindent
All experiments are conducted on NVIDIA L40S and H200 GPU.

\section{Ablation Studies}
We conduct ablation studies to analyze relative performance trends and examine the impact of key design choices in our model, focusing on temporal modeling, tactile encoder capacity, and tactile signal representation. All ablations are performed on a reduced subset of the full dataset (approximately half of the data), following the same 80\% / 10\% / 10\% train–validation–test split as in the main experiments.

\subsection{Temporal Window Size.}
We investigate how the model reacts to different input sequence durations by varying the window size $N \in \{5, 10, 20\}$. The results are shown in Table~\ref{tab:ablation_time_window}. We find that longer temporal input generally improves retrieval performance for visual and tactile modalities. For example, in the Video+Pose$\to$Tactile setting, mAP increases from 19.89\% at $N=5$ to 24.46\% at $N=20$. This suggests that tactile contact patterns and visual motions unfold over time, and the model needs enough temporal history to identify these signals. 

\subsection{Tactile Encoder Capacity.}
To evaluate the efficiency and effectiveness of our tactile encoder, we compare Lite-CNN (which takes the native 16$\times$16 grayscale tactile map) with a standard ResNet-18 baseline (which expects 224$\times$224 inputs). For both models, we fix the input sequence length at $N=10$. Prior work often treats tactile pressure maps as images and applies vision encoders such as ResNet for feature extraction. Since our glove only provides a 16$\times$16 pressure map, and vision pretrained models generally require a larger input size, we upsample the original 16$\times$16 tensor to 224$\times$224 using bilinear interpolation to match ResNet-18’s input format and ensure a consistent experimental setup. Our results show that the lightweight Lite-CNN achieves better performance on this task. As shown in Table~\ref{tab:tactile_baseline_comparison}, our method improves mAP by 10.49\% over ResNet-18 in the video-to-tactile retrieval setting. Simply enlarging the tactile map and using a large vision encoder does not provide an advantage. Vision backbone models are fundamentally designed for natural images, and their assumptions do not match the low-resolution structure of tactile pressure maps. In contrast, a lightweight encoder that operates directly on the native resolution can better preserve the meaningful structure of tactile signals and produce more reliable representations.

\subsection{Tactile Discretization Strategies.}
We evaluate logarithmic and linear discretization with 3, 5, and 7 levels, as well as raw continuous inputs, using a fixed sequence length of $N=20$. Detailed results including Recall@1, Recall@5, Recall@10, and mAP are presented in Table~\ref{tab:tactile_ablation_dis}.

\begin{table*}[t]
\centering
\small
\setlength{\tabcolsep}{8pt}
\caption{\textbf{Ablation Study on Window Size.} We report Recall@1/5/10, and Mean Average Precision (mAP) for different context window sizes across various cross-modal retrieval tasks.}
\label{tab:ablation_time_window}
\resizebox{\textwidth}{!}{%
\begin{tabular}{l|cccc|cccc|cccc}
\toprule
\multirow{2}{*}{\textbf{Task}} & \multicolumn{4}{c|}{\textbf{Window = 5}} & \multicolumn{4}{c|}{\textbf{Window = 10}} & \multicolumn{4}{c}{\textbf{Window = 20}} \\
\cmidrule(lr){2-5} \cmidrule(lr){6-9} \cmidrule(lr){10-13}
 & R@1 & R@5 & R@10 & mAP & R@1 & R@5 & R@10 & mAP & R@1 & R@5 & R@10 & mAP \\
\midrule
Video $\to$ Tactile & 5.77 & 22.12 & 32.85 & 12.57 & 7.85 & 25.48 & 38.62 & 15.44 & \textbf{8.49} & \textbf{26.76} & \textbf{41.83} & \textbf{16.76} \\

Tactile $\to$ Video & 4.17 & 20.99 & 36.38 & 11.74 & 6.09 & 24.84 & \textbf{38.46} & 14.24 & \textbf{8.17} & \textbf{26.12} & 37.66 & \textbf{16.23} \\
Tactile $\to$ Pose & 6.89 & 21.63 & 32.05 & 13.33 & \textbf{7.69} & \textbf{24.20} & \textbf{34.62} & \textbf{14.82} & 7.21 & 23.56 & 33.97 & 14.33 \\
Pose $\to$ Tactile & 5.77 & 20.99 & 33.81 & 12.79 & \textbf{8.65} & \textbf{24.52} & \textbf{36.22} & \textbf{15.41} & 7.05 & 23.40 & 34.78 & 13.97 \\
\midrule
Video + Pose $\to$ Tactile & 9.62 & 32.53 & 50.32 & 19.89 & 6.57 & 28.04 & 40.54 & 15.28 & \textbf{12.82} & \textbf{39.42} & \textbf{56.73} & \textbf{24.46} \\
Tactile + Visual $\to$ Pose & 11.38 & 35.42 & 50.48 & 21.40 & \textbf{15.71} & 41.03 & \textbf{57.53} & \textbf{27.19} & 15.22 & \textbf{42.15} & 56.25 & 26.55 \\
Tactile + Pose $\to$ Visual & 7.53 & 26.28 & 40.87 & 15.99 & 8.81 & 33.49 & 50.48 & 19.44 & \textbf{15.87} & \textbf{39.74} & \textbf{55.77} & \textbf{25.96} \\
\bottomrule
\end{tabular}%
}
\end{table*}

\begin{table*}[t]
\centering
\small
\setlength{\tabcolsep}{8pt} 

\caption{\textbf{Ablation Study on Encoder Baseline.} We compare our method (Lite-CNN, 16$\times$16 resolution) with a ResNet-18~\cite{he2016deep} baseline (224$\times$224). Lite-CNN significantly outperforms the baseline across all cross-modal retrieval tasks.}
\label{tab:tactile_baseline_comparison}

\begin{tabular}{l|cc|cc|cc|cc}
\toprule
\multirow{2}{*}{\textbf{Task}} & \multicolumn{2}{c|}{\textbf{Recall@1}} & \multicolumn{2}{c|}{\textbf{Recall@5}} & \multicolumn{2}{c|}{\textbf{Recall@10}} & \multicolumn{2}{c}{\textbf{mAP}} \\
\cmidrule(lr){2-3} \cmidrule(lr){4-5} \cmidrule(lr){6-7} \cmidrule(lr){8-9}
 & RN & Lite-CNN & RN & Lite-CNN  & RN & Lite-CNN  & RN & Lite-CNN  \\
\midrule
Video $\to$ Tactile & 3.53 & \textbf{8.49} & 10.10 & \textbf{26.76} & 14.10 & \textbf{41.83} & 6.27 & \textbf{16.76} \\
Tactile $\to$ Visual & 2.56 & \textbf{8.17} & 8.81 & \textbf{26.12} & 12.34 & \textbf{37.66} & 5.23 & \textbf{16.23} \\
Tactile $\to$ Pose & 3.53 & \textbf{8.49} & 9.29 & \textbf{26.76} & 18.27 & \textbf{41.83} & 6.53 & \textbf{16.76} \\
Pose $\to$ Tactile & 3.85 & \textbf{8.17} & 13.62 & \textbf{26.12} & 20.19 & \textbf{37.66} & 8.37 & \textbf{16.23} \\
Video + Pose $\to$ Tactile & 5.77 & \textbf{6.57} & 21.47 & \textbf{28.04} & 33.81 & \textbf{40.54} & 12.64 & \textbf{15.28} \\
\bottomrule
\end{tabular}
\end{table*}

\begin{table*}[t]
    \centering

  \small
  \setlength{\tabcolsep}{8pt}
    \caption{\textbf{Ablation on Discretization Strategies.} We compare Logarithmic and Linear discretization at different levels (3, 5, 7) with the use of Raw Continuous signals. The Raw Continuous setting generally yields the highest mAP across most tasks. All values are reported in percentages (\%).}
    \label{tab:tactile_ablation_dis}
    \begin{tabular}{l l | cccc | ccc}
    \toprule
     & & \multicolumn{4}{c|}{\textbf{Bi-modal retrieval}} & \multicolumn{3}{c}{\textbf{Tri-modal retrieval}} \\
    \textbf{Method} & \textbf{Metric} & \textbf{V$\to$T} & \textbf{T$\to$V} & \textbf{T$\to$P} & \textbf{P$\to$T} & \textbf{VP$\to$T} & \textbf{TV$\to$P} & \textbf{TP$\to$V} \\
    \midrule
    
    \multirow{4}{*}{\textbf{Log 3-level}} 
      & Recall@1 & 4.49 & 4.17 & 3.21 & 3.21 & 5.61 & 13.14 & 9.78 \\
      & Recall@5 & 16.35 & 15.06 & 13.14 & 11.70 & 19.39 & 36.38 & 33.49 \\
      & Recall@10 & 23.72 & 22.12 & 19.39 & 18.59 & 30.45 & 46.96 & 46.47 \\
      & mAP & 9.41 & 8.69 & 7.29 & 7.08 & 11.71 & 22.62 & 20.21 \\
    \addlinespace
    
    \multirow{4}{*}{\textbf{Log 5-level}}
      & Recall@1 & 6.89 & 6.73 & 4.97 & 5.61 & 9.46 & 16.67 & 12.34 \\
      & Recall@5 & 22.12 & 20.67 & 16.99 & 17.31 & 29.97 & 41.35 & 35.58 \\
      & Recall@10 & 31.57 & 32.37 & 23.88 & 23.56 & 42.47 & 51.60 & 48.72 \\
      & mAP & 12.98 & 13.06 & 10.04 & 10.66 & 18.00 & 26.77 & 22.44 \\
    \addlinespace
    
    \multirow{4}{*}{\textbf{Log 7-level}}
      & Recall@1 & 7.53 & 7.53 & 5.13 & 5.45 & 11.06 & 14.42 & 12.50 \\
      & Recall@5 & 24.52 & 23.40 & 18.75 & 16.51 & 31.89 & 39.74 & 40.54 \\
      & Recall@10 & 38.78 & 35.42 & 27.24 & 26.76 & 47.60 & 53.53 & 54.81 \\
      & mAP & 15.06 & 14.69 & 10.72 & 10.58 & 20.16 & 25.36 & 24.13 \\
    \midrule
    
    \multirow{4}{*}{\textbf{Linear 3-level}}
      & Recall@1 & 6.41 & 7.05 & 4.65 & 5.29 & 8.17 & 15.54 & 12.18 \\
      & Recall@5 & 20.19 & 20.83 & 18.59 & 20.83 & 29.81 & 39.10 & 38.14 \\
      & Recall@10 & 30.13 & 30.29 & 29.17 & 29.33 & 44.23 & 52.40 & 51.28 \\
      & mAP & 12.34 & 13.05 & 10.71 & 11.49 & 17.56 & 25.77 & 23.12 \\
    \addlinespace
    
    \multirow{4}{*}{\textbf{Linear 5-level}}
      & Recall@1 & 9.29 & 6.73 & 6.41 & 6.89 & 11.22 & 14.90 & 13.94 \\
      & Recall@5 & 27.56 & 25.96 & 21.79 & 21.15 & 37.50 & 41.19 & 42.79 \\
      & Recall@10 & 37.82 & 38.30 & 32.69 & 29.97 & 53.37 & 54.65 & 55.61 \\
      & mAP & 16.55 & 15.12 & 12.98 & 13.01 & 22.50 & 26.12 & 25.73 \\
    \addlinespace
    
    \multirow{4}{*}{\textbf{Linear 7-level}}
      & Recall@1 & 8.17 & 7.53 & 6.57 & 6.57 & 13.62 & 16.67 & 13.94 \\
      & Recall@5 & 25.32 & 25.00 & 21.31 & 20.67 & 38.14 & 42.15 & 40.38 \\
      & Recall@10 & 39.74 & 40.54 & 30.77 & 30.29 & 55.93 & 55.45 & 56.09 \\
      & mAP & 16.05 & 15.17 & 12.91 & 12.65 & 24.14 & \textbf{27.14} & 25.60 \\
    \midrule
    
    \multirow{4}{*}{\textbf{Raw Continuous}}
      & Recall@1 & 8.49 & 8.17 & 7.21 & 7.05 & 12.82 & 15.22 & 15.87 \\
      & Recall@5 & 26.76 & 26.12 & 23.56 & 23.40 & 39.42 & 42.15 & 39.74 \\
      & Recall@10 & 41.83 & 37.66 & 33.97 & 34.78 & 56.73 & 56.25 & 55.77 \\
      & mAP & \textbf{16.76} & \textbf{16.23} & \textbf{14.33} & \textbf{13.97} & \textbf{24.46} & 26.55 & \textbf{25.96} \\
    \bottomrule
    \end{tabular}
\end{table*}

\end{document}